\documentclass[journal]{IEEEtran}
\usepackage{graphicx}
\usepackage{cite}
\usepackage{moreverb,url}
\usepackage{tabstackengine}

\usepackage[colorlinks,bookmarksopen,bookmarksnumbered,citecolor=red,urlcolor=red]{hyperref}
\usepackage{amsthm}

\newcommand\BibTeX{{\rmfamily B\kern-.05em \textsc{i\kern-.025em b}\kern-.08em
T\kern-.1667em\lower.7ex\hbox{E}\kern-.125emX}}

\usepackage{graphicx, amsmath, amssymb}
\ifCLASSOPTIONcompsoc
    \usepackage[caption=false, font=normalsize, labelfont=sf, textfont=sf]{subfig}
\else
\usepackage[caption=false, font=footnotesize]{subfig}
\fi

\newtheorem{prop}{Proposition}
\newtheorem{definition}{Definition}[section]

\begin{document}
\title{The Geometric Structure of Externally Actuated Planar Locomoting Systems in Ambient Media}
% \\\{blakeb, mtravers, choset\}@andrew.cmu.edu}
\author{
\IEEEauthorblockN{Blake Buchanan\IEEEauthorrefmark{1}, Tony Dear\IEEEauthorrefmark{2}, Scott David Kelly\IEEEauthorrefmark{3}, Matthew Travers\IEEEauthorrefmark{1}}, and Howie Choset\IEEEauthorrefmark{1}\\
    \thanks{\hspace*{-1.2em}\IEEEauthorblockA{\IEEEauthorrefmark{1}The Robotics Institute\\ Carnegie Mellon University, Pittsburgh, PA, USA}\\}
    \thanks{\hspace*{-1.1em}\IEEEauthorblockA{\IEEEauthorrefmark{2}Department of Computer Science \\ Columbia University, New York, NY, USA}\\}
    \thanks{\hspace*{-1.1em}\IEEEauthorblockA{\IEEEauthorrefmark{3}Mechanical Engineering and Engineering Science\\ University of North Carolina at Charlotte, Charlotte, NC, USA\\}}
    \thanks{\hspace*{-1.45em} \textbf{Corresponding Author}:\\ Blake Buchanan, The Robotics Institute, Carnegie Mellon University \\
    Email: \text{blakeb@andrew.cmu.edu}}
    }

\maketitle

\begin{abstract}
Robots often interact with the world via attached parts such as wheels, joints, or appendages. In many systems, these interactions, and the manner in which they lead to locomotion, can be understood using the machinery of geometric mechanics, explaining how inputs in the shape space of a robot affect motion in its configuration space and the configuration space of its environment. In this paper we consider an opposite type of locomotion, wherein robots are influenced actively by interactions with an externally forced ambient medium. We investigate two examples of externally actuated systems; one for which locomotion is governed by a principal connection, and is usually considered to possess no drift dynamics, and another for which no such connection exists, with drift inherent in its locomotion. For the driftless system, we develop geometric tools based on previously understood internally actuated versions of the system and demonstrate their use for motion planning under external actuation. For the system possessing drift, we employ nonholonomic reduction to obtain a reduced representation of the system dynamics, illustrate geometric features conducive to studying locomotion, and derive strategies for external actuation.
\end{abstract}


% \fi
\IEEEpeerreviewmaketitle

%%%%%%%%%%%%%%%%%%%%%%%%%%%%%%%%%%%%%%%%%%%%%%%%%%%%%%%%%%%%%%%%%%%%%%
\section{Introduction}
%Robot motion planning often turns to observations from biology for inspiration of gaits or locomotion modes. While the nuances of locomotion across different systems vary, the goal is generally the same---the agent exploits the underlying structure of the problem to effect movement in a desired way. But even if robots are able to mimic biological locomotion, the manner in which motions are produced may differ. In particular, biological agents may be able to cleverly exploit passive dynamics due to their body structure or environmental forces, while robots attempt to fully simulate these effects in their controllers.
% In our companion paper (\cite{dear2018a}), we explore motion planning for a multi-link wheeled robot like that shown in Fig.~\ref{fig:snakeRobot}, for which previous studies assume that both internal joints are fully controlled, in which case a simple geometric model connects internal joint trajectories to overall locomotion in the world.

\begin{figure}[t]
\begin{center}
\includegraphics[width=.48\textwidth]{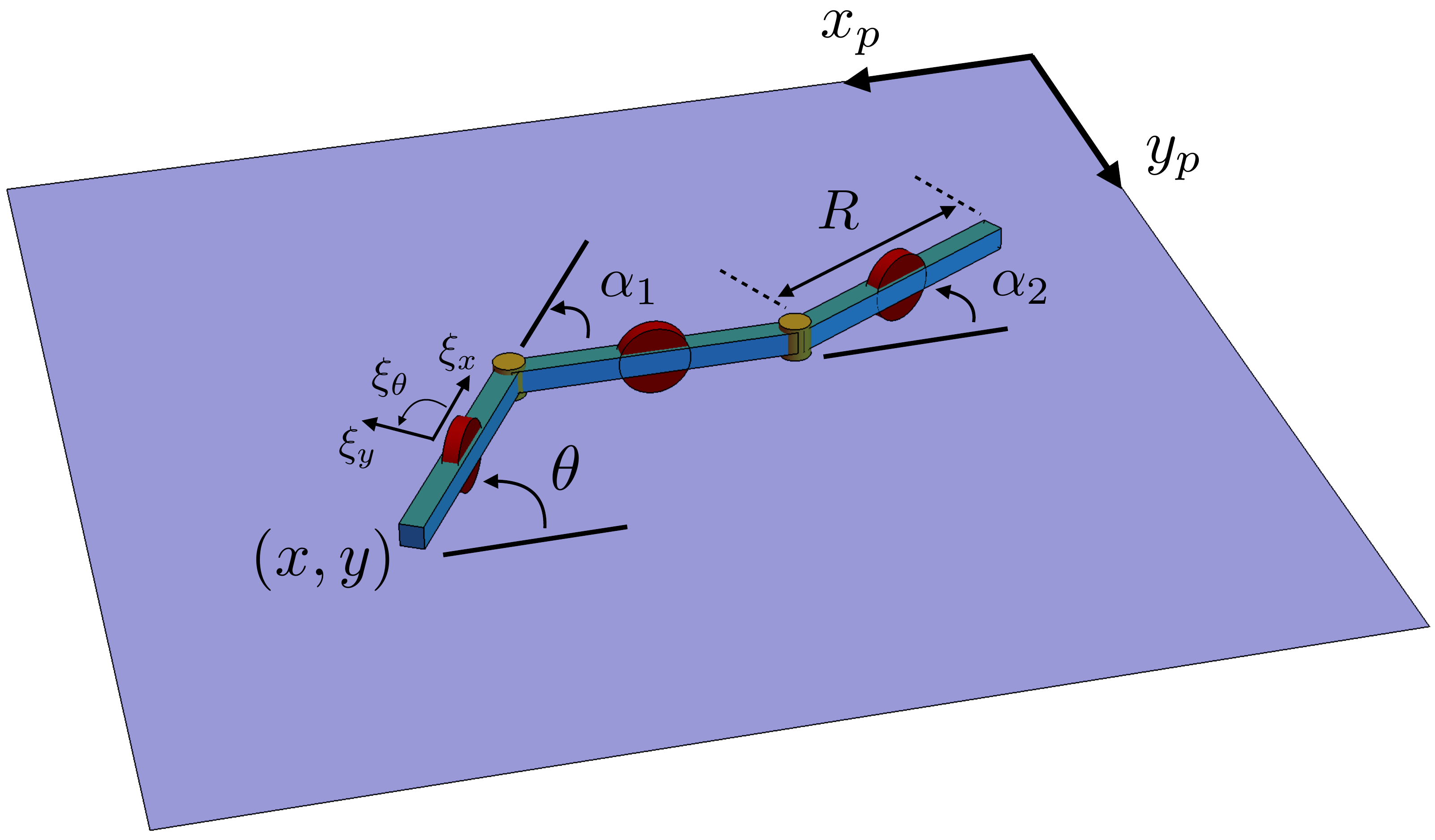}
\end{center}
\caption{A three-link nonholonomic robot on top of a movable platform. The coordinates $(x,y,\theta)$ denote the inertial configuration of the robot's proximal link, with body velocities $(\xi_x, \xi_y, \xi_\theta)$. The relative joint angles are $(\alpha_1, \alpha_2)$. The platform's inertial position is $(x_p,y_p)$.}
\label{fig:snakeRobot} 
\end{figure}

It is often the case that the locomotion of a robot can be understood in a simplified context by analyzing only its joint trajectories. For example, a simple geometric model has been developed for the multi-link wheeled robot shown in Fig.~\ref{fig:snakeRobot}, in which joint trajectories map to world trajectories through a principal connection. We have shown in \cite{dear2020} that this robot retains similar locomotive capabilities if one joint becomes passive, by combining the geometric model with the dynamics of the passive joint. In this work, we study an alternative problem where the joints of the robot are completely passive, with input energy being channeled from the surrounding environment. In particular, the presence of a controlled moving platform underneath the robot, as shown in Fig.~\ref{fig:snakeRobot}, provides this input.
%%%%%%%%%%%%%%%%%%%%%%%%%%%%%%%%%%%%%%%%%%%%%%%%%%%%%%%%%%%%%%%%%%%%%%
%%%%%%%%%%%%%%%% begin figure %%%%%%%%%%%%%%%%%%%
%%%%%%%%%%%%%%%% end figure %%%%%%%%%%%%%%%%%%% 
%%%%%%%%%%%%%%%%%%%%%%%%%%%%%%%%%%%%%%%%%%%%%%%%%%%%%%%%%%%%%%%%%%%%%%

The first objective of this work is to seek to understand how this system's geometric structure changes and to what extent it can still be exploited to understand locomotion when control authority is assumed over the environment. It is not always the case, however, that the locomotion of a robot can be understood by analyzing only its joint trajectories. There are many systems for which drift is present in the dynamics and a mapping between the joint trajectories of the robot and world trajectories does not exist, meaning one is tasked with invoking and / or developing additional tools to analyze locomotion or accomplish control.

\begin{figure}[t]
\begin{center}
\includegraphics[trim={0.25cm 0.5cm 0.5cm 0},clip,width=.35\textwidth]{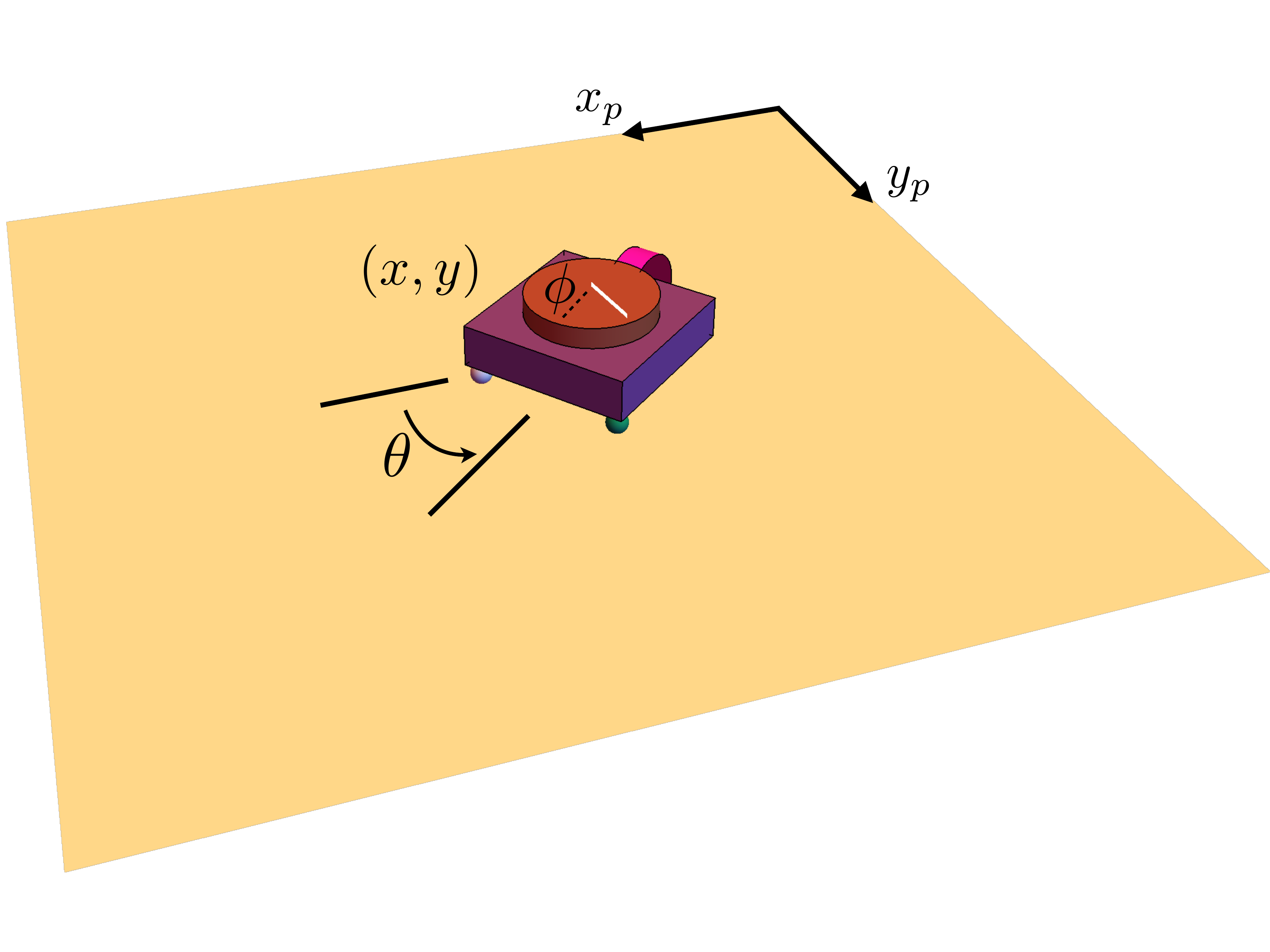}
\end{center}
\vspace{-0.5cm}
\caption{A Chaplygin beanie atop a movable platform. The vehicle's rotor angle relative to the heading is shown as $\phi$, its heading as $\theta$, its position relative to the platform as $(x,y)$, and the position of the platform in a world frame, $(x_p,y_p)$.}
\label{fig:beanieonaplatform}
\end{figure}

The canonical system with drift considered in this paper is the Chaplygin beanie on a movable platform, shown in Fig. \ref{fig:beanieonaplatform}. Any change in the rotor angle $\phi$ of the Chaplygin beanie will cause it, and the platform beneath it, to accrue nonzero momentum, and thus drift in its environment. The second objective of this work is to understand how the geometric structure of such a system can be exploited when its locomotion is inherently driftful and not described by a principal connection, and again when control authority is assumed over the environment.

While the tools with which we analyze the locomotion of these two systems differ geometrically, they share a common theme in that the robot is coupled to a platform that acts as the medium through which dynamics are imparted to the robot. More importantly, we use these examples to illustrate how the  problem structure changes due to how external inputs are applied. We make an explicit choice to consider only translational motion of the platform, engendering symmetries that we leverage in developing the tools we use to accomplish motion planning and control.

This paper juxtaposes results from a recent conference paper\cite{Buchanan_2020} with complementary new material to construct a more unified narrative concerning externally actuated nonholonomic planar locomoting systems in ambient media. We first present results that draw on familiar tools for the three-link kinematic snake robot on a platform for which locomotion is governed purely by a principal connection. We then discuss our proposed modifications to these tools which allow one to leverage connection-like mappings outside of the theory of principal connections to aid in motion planning when considering symmetry-breaking fiber variables. Finally, we present results concerning the control of the Chaplygin beanie on a platform, a system for which locomotion is not governed by a principal connection. The material in the present paper differs from that appearing in our recent conference paper (Section \ref{sec:chap}) in that our conference paper did not leverage or extend familiar tools arising from principal connections that we introduce in Sections \ref{sec:roleofext} and \ref{sec:snake}. These new tools provide a way to accomplish motion planning and control for externally actuated planar locomoting systems when an internally actuated version of the system can be analyzed using principal connections.

We have structured the material as follows. We review related work in geometric mechanics and locomotion in Section \ref{sec:priorwork}, provide some mathematical preliminaries in Section \ref{sec:math}, and present our contributions to the geometric theory in Section \ref{sec:roleofext}. Section \ref{sec:snake} considers the problem of a passive three-link wheeled robot on a movable platform, exemplifying the cases in which external inputs are applied with respect to a body frame as well as inertial frame dependent on system orientation. In Section \ref{sec:chap} we first discuss a completely passive system and provide a formal proof of stable trajectories, guiding our approach to the external actuation of the Chaplygin beanie on a movable platform. We illustrate the geometric features of the problem that are conducive to studying its locomotion and conduct a sampling-based analysis to derive motion primitives. We conclude with a review of our contributions and discuss novel problems arising from our analyses and possible approaches to solving them.
%%%%%%%%%%%%%%%%%%%%%%%%%%%%%%%%%%%%%%%%%%%%%%%%%%%%%%%%%%%%%%%%%%%%%%
%%%%%%%%%%%%%%%% begin figure %%%%%%%%%%%%%%%%%%%
%%%%%%%%%%%%%%%% end figure %%%%%%%%%%%%%%%%%%% 
%%%%%%%%%%%%%%%%%%%%%%%%%%%%%%%%%%%%%%%%%%%%%%%%%%%%%%%%%%%%%%%%%%%%%%

%%%%%%%%%%%%%%%%%%%%%%%%%%%%%%%%%%%%%%%%%%%%%%%%%%%%%%%%%%%%%%%%%%%%%%
\section{Prior Work}
\label{sec:priorwork}
In recent decades, techniques and methods from geometric mechanics have been a popular way to model and control mechanical systems. A key idea is that of \emph{symmetries} in a system's configuration space, which allows for the \emph{reduction} of the equations of motion to a simpler first-order form. This has been addressed for general mechanical systems by \cite{marsden1990reduction}, \cite{marsden2013introduction}, and \cite{marsden1993reduced}, as well as nonholonomic systems by \cite{bloch1996nonholonomic} and \cite{ostrowski1998reduced}. 

For locomoting robots, geometric reduction is often leveraged in tandem with a decomposition of the configuration variables into actuated shape variables, describing internal configurations of the robot, and position, describing the position and orientation of the body frame with respect to an inertial one. If such a decomposition is possible, then the configuration space often takes on a \emph{fiber bundle} structure, whereby a mapping called the \emph{connection} relates trajectories from the shape space to the position space. Analysis of the connection then gives us intuition into ways to perform motion planning and control on the system, engendering visualization and design tools, detailed by \cite{kelly1996geometry, ostrowski1998geometric, grover2018, kadam2017trajectory, melli2006motion, kadam2018geometry, Hatton01072011}.

Much of the progress in the geometric mechanics of locomotion is predicated on the assumption that the symmetries of a system coincide exactly with the position degrees of freedom. The three-link robot of Fig.~\ref{fig:snakeRobot} with actuated joints is one of the simplest examples of such a system, and as such, it has received considerable attention from researchers such as \cite{ostrowski1999computing} and \cite{Shammas01102007a} treating it as a \emph{kinematic} system, so named because its three constraints eliminate the need to consider second-order dynamics when modeling its locomotion. This allows for the treatment of the system's locomotion, and subsequent motion planning, as a result of geometric phase \cite{murray1993nonholonomic, mukherjee1993nonholonomic, kelly1995geometric, ostrowski2000optimal, bullo2001kinematic}.

While the notion of geometric phase lends itself to the study of certain locomoting robots like the three-link wheeled snake robot, a different set of tools are required for the analysis and computation of motion primitives for others. One can still consider the configuration space to be split into a base and a fiber space, but a connection relating trajectories in these spaces is not guaranteed. One such system is that of the Chaplygin beanie on a movable platform. Any actuation of its single shape variable will cause motion along its fibers, and in a partially uncontrollable way. The notion of nonholonomic momentum was used for the analysis of an internally actuated Chaplygin beanie in \cite{kelly2012proportional}. We will employ reduction using nonholonomic momentum in our analysis of a completely passive Chaplygin beanie on a platform system, which will then guide our approach to finding motion primitives for an externally actuated version of the system. Other internally actuated and passively compliant versions of this robot have been studied in \cite{fairchild2011single, fedonyuk2019dynamics} with limit cycle behaviors elucidated in \cite{pollard2019swimming}.

External actuation to achieve locomotion has been used for sorting and controlling the motion of mechanical parts via controlled vibrations in \cite{reznik2001c, vose2009friction, vose2012sliding}, and has been used in the famous ball-on-a-plate problem \cite{debono2015application, ghiasi2012optimal}. In this paper, we give the problem of external actuation a geometric flavor and invoke familiar tools from the geometry of locomotion. In prior work we posed the problem of actuating a three-link wheeled snake robot via an external platform as a specific deviation from the usual geometric assumptions \cite{dear2016variations}. We also investigated an externally actuated version of the Chaplygin beanie in \cite{Buchanan_2020}. These two example systems belong to different classes in that the external actuation of one can be studied using a principal connection formulation, while the other demands reduction by means of nonholonomic momenta. The framework within which each of the examples is analyzed differs geometrically, clarifying the set of tools necessary to understand external actuation in each setting.

%%%%%%%%%%%%%%%%%%%%%%%%%%%%%%%%%%%%%%%%%%%%%%%%%%%%%%%%%%%%%%%%%%%%%%
\section{Mathematical Preliminaries}
\label{sec:math}
In this section, provide an overview of the mathematical ideas we will invoke in the present paper and define terms we will use throughout. 
\subsection{Principal Fiber Bundles}
We now review the underlying geometric structure for locomoting systems like the three-link wheeled snake robot.
%%%%%%%%%%%%%%%%%%%%%%%%%%%%%%%%%%%%%%%%%%%%%%%%%%%%%%%%%%%%%%%%%%%%%%
%%%%%%%%%%%%%%%% begin figure %%%%%%%%%%%%%%%%%%%
\begin{figure}[t]
	\begin{center}
		\includegraphics[width=.35\textwidth]{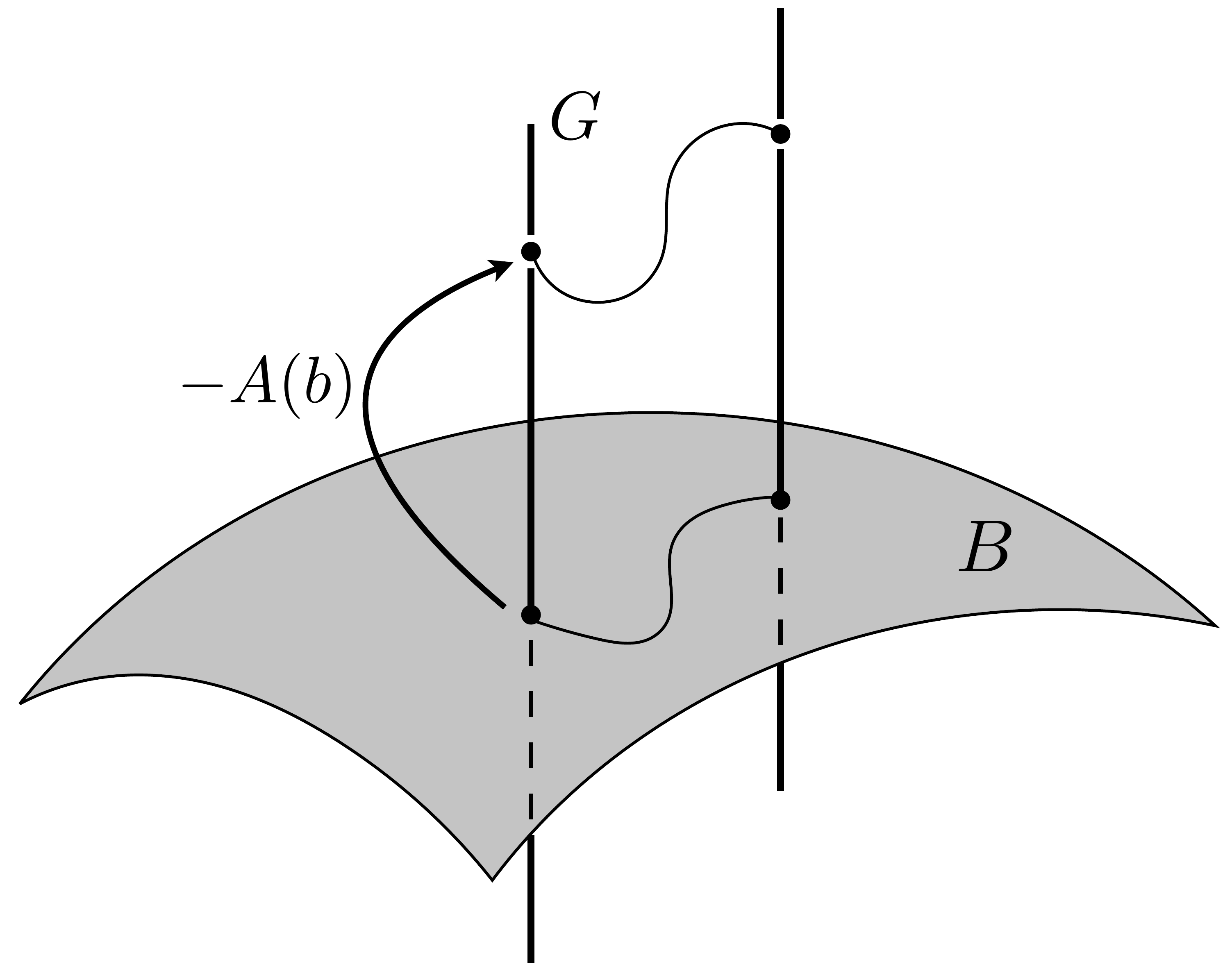}
	\end{center}
	\caption{A visualization of the principal fiber bundle. Base trajectories in $B$ are lifted via the connection $-A(b)$ to trajectories in the fiber space $G$.}
	\label{fig:fiber-bundle} 
\end{figure}
%%%%%%%%%%%%%%%% end figure %%%%%%%%%%%%%%%%%%% 
%%%%%%%%%%%%%%%%%%%%%%%%%%%%%%%%%%%%%%%%%%%%%%%%%%%%%%%%%%%%%%%%%%%%%% 
The usage of geometric mechanics in locomotion has traditionally assumed the existence of a principal fiber bundle structure in a system's configuration space. This structure decomposes the space into two distinct subspaces, often called a \textit{fiber bundle}, with a mapping called a \textit{connection} relating trajectories from one to trajectories in the other.
\begin{definition}[Fiber Bundle]
A \textbf{fiber bundle} is a trio of topological spaces $Q$, the total space, $B$, the base space, and $G$, the fiber space, together with a map $\pi:Q\rightarrow B$ called the bundle projection such that every $r \in B$ has an open neighborhood $U \subset B$, there exists a homeomorphism $\varphi:\pi^{-1}(U)\rightarrow U\times G$, and $Q\cong G\times B$, i.e., $Q$ is locally equivalent to $G\times B$, but not necessarily globally.
\end{definition}
In this work, we will consider the configuration spaces of a robot and its environment to comprise \textit{smooth} fiber bundles, particularly \textit{principal fiber bundles}.
\begin{definition}[Principal Fiber Bundle]
A \textbf{principal fiber bundle} is a smooth fiber bundle $(Q,B,F,\pi)$ together with an action of a Lie group $G$ on $Q$ such that
\begin{enumerate}
    \item the action of the Lie group preserves fibers ($gq \in Q$ for every $g \in G$)
    \item the action is free (for $g,h \in G$, if $q\in Q$, then $gq = hq$ implies $g = h$) and transitive (for pairs $q_i,q_j\in Q$, there exists a $g\in G$ such that $gq_i = q_j$) on each fiber.
\end{enumerate}
\end{definition}
Fig.~\ref{fig:fiber-bundle} shows a typical visualization of the fiber bundle as a direct product of a base space $B$ and a fiber space $G$. The space $B$ corresponds to the system's shape variables, which are traditionally assumed to be fully actuated. The space $G$ corresponds to the system's configuration variables, which typically correspond to \textit{symmetries} in the system, so that the dynamics of the system do not depend explicitly on the fiber variables. Since we will make use of symmetries in both of the examples presented, a system exhibits a \textit{continuous symmetry} if its dynamics are invariant under the free action of a Lie group on its configuration manifold \cite{marsden2013introduction}.
% \begin{definition}[Symmetry]
% Let $G$ be a Lie group with a group action $\Phi$ that acts on the left of a manifold $Q$. A mechanical system exhibits a continuous \textbf{symmetry} if either of the following is true.
% \begin{enumerate}
% \item if its dynamics are invariant under the free action of a Lie group on its configuration manifold.
%     \item The equations arising from kinematic constraints on the system are invariant under changes in position and orientation. Such equations are said to be \textit{symmetric} with respect to the Lie group $G$.
%     \item The tangent lifted action corresponding to the action $\Phi$ leaves the Lagrangian invariant. That is,
%     $L(q,\dot{q}) = L(\Phi(g,q),T_q\Phi(g,q)\dot{q})$.
% \end{enumerate}
% % Conserved quantities associated with continuous symmetries in this way are termed momentum maps.
% \end{definition}

In this paper, the symmetry exploited for the three-link wheeled snake robot on a platform is given by the invariance of the system's dynamics under the action of $\text{SE}(2)$ on its configuration manifold $Q = \text{SE}(2) \times \mathbb{S}^1 \times \mathbb{S}^1$. The Chaplygin beanie on a platform system exhibits a symmetry arising from the action of the Lie group $G = \text{SE}(2) \times \mathbb{R}^2$ on the configuration manifold $Q = \text{SE}^2 \times \mathbb{R}^2 \times \mathbb{S}^1 \times \mathbb{S}^1$.

The local form of the connection $A(b)$, which depends only on base variables $b$, governs how the system's fiber variables change due to changes in the base. The equation is typically of the form
\begin{equation} 
\dot g = -T_e L_g (A(b) \dot b),
\label{eq:fiber-reconstruction-eq}
\end{equation}
where $\dot g$ are the velocities of the fibers, and $\dot b$ are the velocities of the shape variables. The mapping $T_eL_g$ for $g \in \text{SE}(2)$ is given by
\begin{equation}
	T_e L_g = \begin{bmatrix}
		\cos \theta & -\sin\theta & 0 \\ \sin\theta & \cos \theta & 0 \\ 0 & 0 & 1
	\end{bmatrix},
	\label{eq:TeLg}
\end{equation}
with $\theta$ representing the orientation of the robot relative to an inertial reference frame in the plane. The mapping $T_e L_g$  transforms fiber velocities from a body-attached frame to a world frame; in practice, one can drop the $T_e L_g$ mapping by performing all analysis in the body frame. If we are able to express the fiber velocities $\dot g$ as body velocities $\xi$ via the transformation $\dot g = T_e L_g \xi$, then Eq.~\eqref{eq:fiber-reconstruction-eq} can be recast into the \emph{kinematic reconstruction equation}\cite{hatton2011geometric}, given by
\begin{equation} 
\xi = -A(b) \dot b.
\label{eq:reduced-reconstruction-eq}
\end{equation}

As an example, consider the wheeled robot of Fig.~\ref{fig:snakeRobot} in the absence of an underlying platform. Its configuration space can naturally be written as $Q_w = G_w \times B_w$, where $g_w = (x,y,\theta)^T \in G_w$ are the position and orientation fiber variables, and $b_w = (\alpha_1, \alpha_2)^T$ are the joint, or shape, variables.\footnote{We use the subscript $w$ for the configuration space of the \textit{wheeled} snake robot. Similarly, we will use the subscript $c$ for the Chaplygin beanie.} The fiber $G_w$ is the Lie group of planar translations and rotations SE$(2)$. The body velocities $\xi$ are related to the world velocities $\dot g_w$ through the mapping $T_e L_g$.
% In the second example of the Chaplygin beanie on a platform, the configuration space is deconstructed as $Q_c = G_c \times B_c$, where the fiber space is $G_c = SE(2)$ as before (since locomotion is again planar), while the shape space $B_s = (s_x, s_y)$ consists of the internal mass's position relative to the swimmer's body \footnote{As before, the subscript $s$ refers to the configuration of the \textit{swimming} robot.}.

The structure of the connection form in Eq.\ \eqref{eq:reduced-reconstruction-eq} can be visualized in order to understand the response of $\xi$ to input trajectories without regard to time, according to \cite{hatton2011geometric}. By integrating each row of Eq.\ \eqref{eq:reduced-reconstruction-eq} over a given joint trajectory, one can obtain a measure of displacement corresponding to the body frame directions. For $\text{SE}(2)$, this measure provides the exact rotational displacement, \textit{i.e.}, $\dot \theta = \xi_\theta$ for the third row, and an approximation of the translational component for the first two rows.

We can consider the time integral to be a line integral over the trajectory $\psi: [0,T] \rightarrow B$ in the shape space, since the kinematics depend only on the gait executed in shape space. This can then be viewed as an integral over $\beta$, the region of the joint space enclosed by $\psi$. 
Assuming that we have periodic trajectories, or gaits, the integral can be realized by Stokes' theorem as
\begin{equation}
\begin{aligned}
    \int_0^T \xi dt & = -\int_0^T A(b(\tau)) \dot b(\tau)\, d\tau = -\int_{\psi} A(b)\, db \\ & = -\int_{\beta} \text{d}A(b) + [A_1,A_2]db + \text{\parbox{2cm}{\centering higher-order terms}}.
\label{eq:stokes}
\end{aligned}
\end{equation}
The first term in the final integral is the exterior derivative of $A(b)$ and is computed as the curl of $A(b)$ in two dimensions, \textit{i.e.,}
\begin{equation}
\text{d}A_i(b) = \frac{\partial A_{i,2}}{\partial \alpha_1} - \frac{\partial A_{i,1}}{\partial \alpha_2},
\label{eqn:dA}
\end{equation}
where $A_{i,j}$ is the element corresponding to the $i$th row and $j$th column of $A$. The second term is the Lie bracket of the two vector fields comprising the local connection form $A(b)$. In this work, we consider only the first term in our approximation of the displacement. Prior works on similar locomoting systems have shown that this approximation suffices for the study of locomotion, even in the absence of Lie bracket and higher-order terms when considering an appropriate choice of coordinates \cite{dear2016variations, dear2020, hatton2013geometric}. Eq. \eqref{eqn:dA} can be extended to arbitrary dimensions \cite{bittner2018geometrically}, however, we define it for two-dimensional base spaces since the present paper focuses on planar systems.

%%%%%%%%%%%%%%%%%%%%%%%%%%%%%%%%%%%%%%%%%%%%%%%%%%%%%%%%%%%%%%%%%%%%%%
%%%%%%%%%%%%%%%% begin figure %%%%%%%%%%%%%%%%%%%
\begin{figure}[t]
\begin{center}
\includegraphics[width=.35\textwidth]{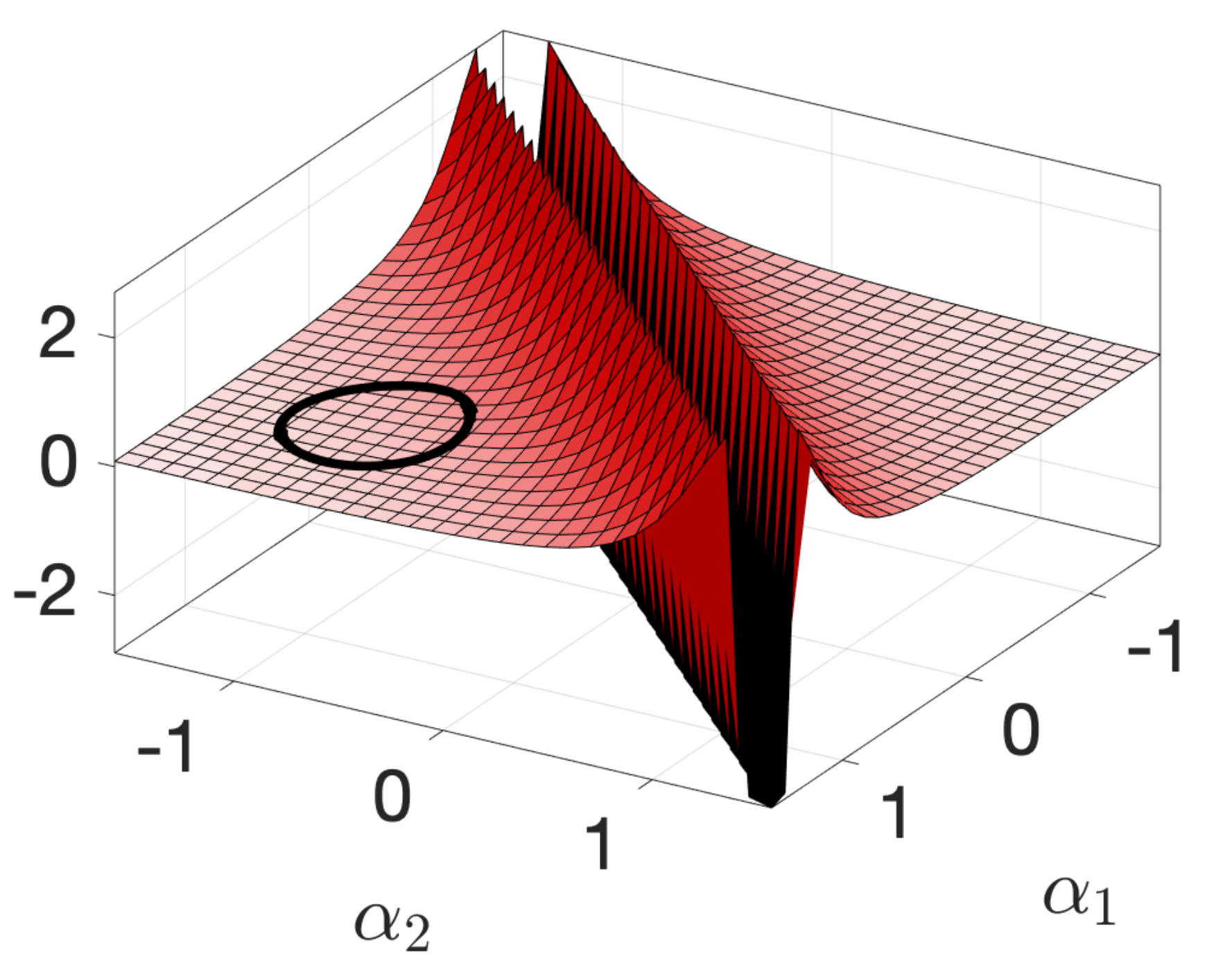}
\includegraphics[width=.35\textwidth]{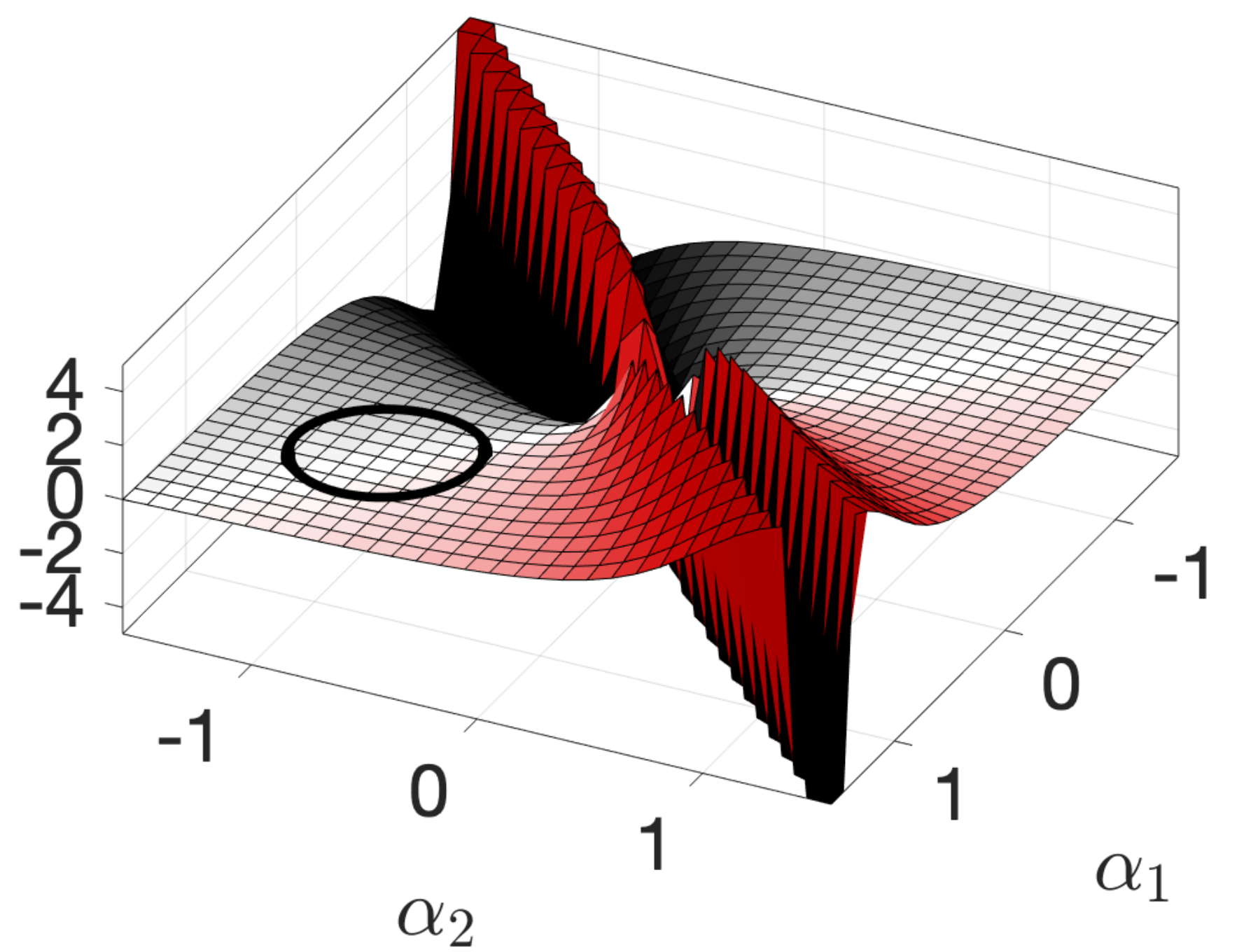}
\end{center}
\caption{The $x$ (top) and $\theta$ (bottom) components of the exterior derivative of the connection that relates the three-link robot's joint velocities to the robot's body velocities. Note that the $x$ plot is non-negative everywhere and that both plots are symmetric about the $\alpha_1=\alpha_2$ diagonal. An example of a gait (closed loop) is also shown on these plots.}
\label{fig:TL-ED} 
\end{figure}
%%%%%%%%%%%%%%%% end figure %%%%%%%%%%%%%%%%%%% 
%%%%%%%%%%%%%%%%%%%%%%%%%%%%%%%%%%%%%%%%%%%%%%%%%%%%%%%%%%%%%%%%%%%%%%

The magnitudes of the three-link robot's connection exterior derivative over the joint space are depicted in Fig.\ \ref{fig:TL-ED}, along with a gait trajectory shown as a closed curve on the surfaces. The area integral over the enclosed region is the geometric phase, a measure of the displacement in the body $x$ and $\theta$ directions (the body $y$ plot is not shown because it is zero everywhere). The $x$ plot is positive everywhere, meaning that any closed loop will lead to net displacement along the $\xi_x$ direction. The $\theta$ plot is anti-symmetric about $\alpha_1 = -\alpha_2$, meaning that gaits symmetric about this line will yield zero net reorientation while simultaneously moving the robot forward.

\subsection{Reduction by Nonholonomic Momentum} Identifying conserved quantities in mechanical systems permits the use of reduction tools we are ultimately interested in leveraging in our analysis of an externally actuated locomoting system with drift in its dynamics, e.g., the Chaplygin beanie on a movable platform. In particular, Noether's theorem \cite{Noether1971InvariantVP} states that continuous symmetries correspond to conserved quantities in a mechanical system. In this section, we introduce the terminology necessary to leverage symmetries in the reduction we employ for the Chaplygin beanie on a movable platform.

Let $v_q$ be a tangent vector at point $q$ in the tangent space $T_qQ$. The momentum map corresponding to a continuous symmetry is computed as
\begin{equation}
    \langle J(v_q),\xi\rangle = \langle \mathbb{F}L(v_q),\xi_Q(q)\rangle,
\end{equation}
where $\mathbb{F}L$ is the fiber derivative of the system Lagrangian, given by
\begin{equation}
    \begin{aligned}
        \mathbb{F}L & :TQ\rightarrow T^*Q \\ & :(q,\dot{q}^i\frac{\partial}{\partial q^i}) \mapsto (q,\frac{\partial L}{\partial \dot{q}^i}dq^i)
    \end{aligned}
\end{equation} Define the $i$th component of the vector field $\xi_Q$ as $(\xi_Q)^i$. We compute the momentum map with respect to coordinates $q^i$ as
\begin{equation}
    J = \frac{\partial L}{\partial \dot{q}^i} (\xi_Q)^i 
    \label{eqn:nonmomenta}
\end{equation}
\noindent where a summation over $i$ is assumed. Note that every vector field $\xi_Q$ on $TQ$ is associated with a component of the momentum map. Given the above definition, the evolution equations corresponding to the nonholonomic momenta are computed as
\begin{equation}
    \dot{J} = \frac{\partial L}{\partial \dot{q}^i} \bigg[\frac{d\xi}{dt}\bigg]_Q^i.
    \label{eqn:evoeqnsbackground}
\end{equation}
We will utilize Eq. \eqref{eqn:evoeqnsbackground} in Section \ref{sec:chap}  in order to provide proof of particular stable trajectories in this reduced space. It is worth noting that the symmetries by which we reduce the dynamics of each system we study correspond to group actions of $G = G^\text{int} \times G^\text{ext}$ on the configuration manifold of the system.

The Lie algebra elements $\xi$ associated with each vector field $\xi_Q$ are needed to determine $\dot J$ in Eq. \eqref{eqn:evoeqnsbackground}. To compute each $\xi$, it is necessary to define the subspace on which the constrained dynamics evolve. Let $\mathcal{D}$ be the distribution spanned by all tangent vectors that annihilate the constraints, and let the group orbit of the Lie group $G$ be
\begin{equation}
\text{Orb}(q) := \{gq \, | \, g \in G\}. 
\end{equation}

Associated with every vector field $\xi_Q$ referenced in Eq. \eqref{eqn:nonmomenta} is a Lie algebra element $\xi \in \mathfrak{g}$ where $\mathfrak{g}$ is the Lie algebra of the Lie group $G$. We define the tangent space to the group orbit through a point $q\in Q$ as
\begin{equation}
    T_q(\text{Orb}(q)) = \{\xi_Q(q)\, | \, \xi \in \mathfrak{g}\},
\end{equation}
where $\xi_Q(q)$ is the vector field $\xi_Q$ evaluated at $q$. We define the intersection of $\mathcal{D}_q$ and $T_q(\text{Orb}(q))$ to be the reduced space in which the constrained dynamics evolve, which we write as
\begin{equation}
    S_q = \mathcal{D}_q \cap T_q(\text{Orb}(q)),
    \label{eqn:intersection}
\end{equation}
where $\mathcal{D}_q$ denotes the constraint distribution at point $q$. We have the freedom to choose the set of vector fields comprising $T_q(\text{Orb}(q))$, and our choice of vector fields determines the reduced space (Eq. \eqref{eqn:intersection}) and thus the resulting evolution equations in Eq. \eqref{eqn:evoeqnsbackground}.

%%%%%%%%%%%%%%%%%%%%%%%%%%%%%%%%%%%%%%%%%%%%%%%%%%%%%%%%%%%%%%%%%%%%%%
\section{Role of External Actuation via Fiber Bundles}
\label{sec:roleofext}
The fiber bundle description of a system assumes that \emph{all} of its configuration variables can be neatly categorized into either the base or a single fiber space. However, this is no longer sufficient when there are other configuration variables, such as those describing the evolution of a non-static ambient medium. In such a scenario, these additional variables will form a third subspace, whose role depends on its interaction with the original fiber bundle and whether system symmetries are still preserved.

\subsection{Stratified Fiber Bundle}
If we consider the example of the three-link wheeled robot of Fig.~\ref{fig:snakeRobot} on a moving platform, then the configuration of the platform would live in a new fiber space. Furthermore, this new fiber space corresponds to another set of symmetries in the system. That is, the system's dynamics and energetics are invariant with respect to the \emph{robot's} $(x,y)$ position (two of its fibers) as well as the \emph{platform's} position (external fibers). A subtlety does arise regarding the robot's orientation, $\theta$, relative to the platform; the dynamics of the platform depend on how the robot is oriented in the environment. A helpful simplification would be to continuously orient the platform's frame of reference to the robot's body frame, eliminating the dependency of the dynamics on $\theta$. In order for this arrangement to be practical, the platform would also ideally be able to move along any direction in the plane.

%%%%%%%%%%%%%%%%%%%%%%%%%%%%%%%%%%%%%%%%%%%%%%%%%%%%%%%%%%%%%%%%%%%%%%
%%%%%%%%%%%%%%%% begin figure %%%%%%%%%%%%%%%%%%%
\begin{figure}[t]
	\begin{center}
		\includegraphics[width=.4\textwidth]{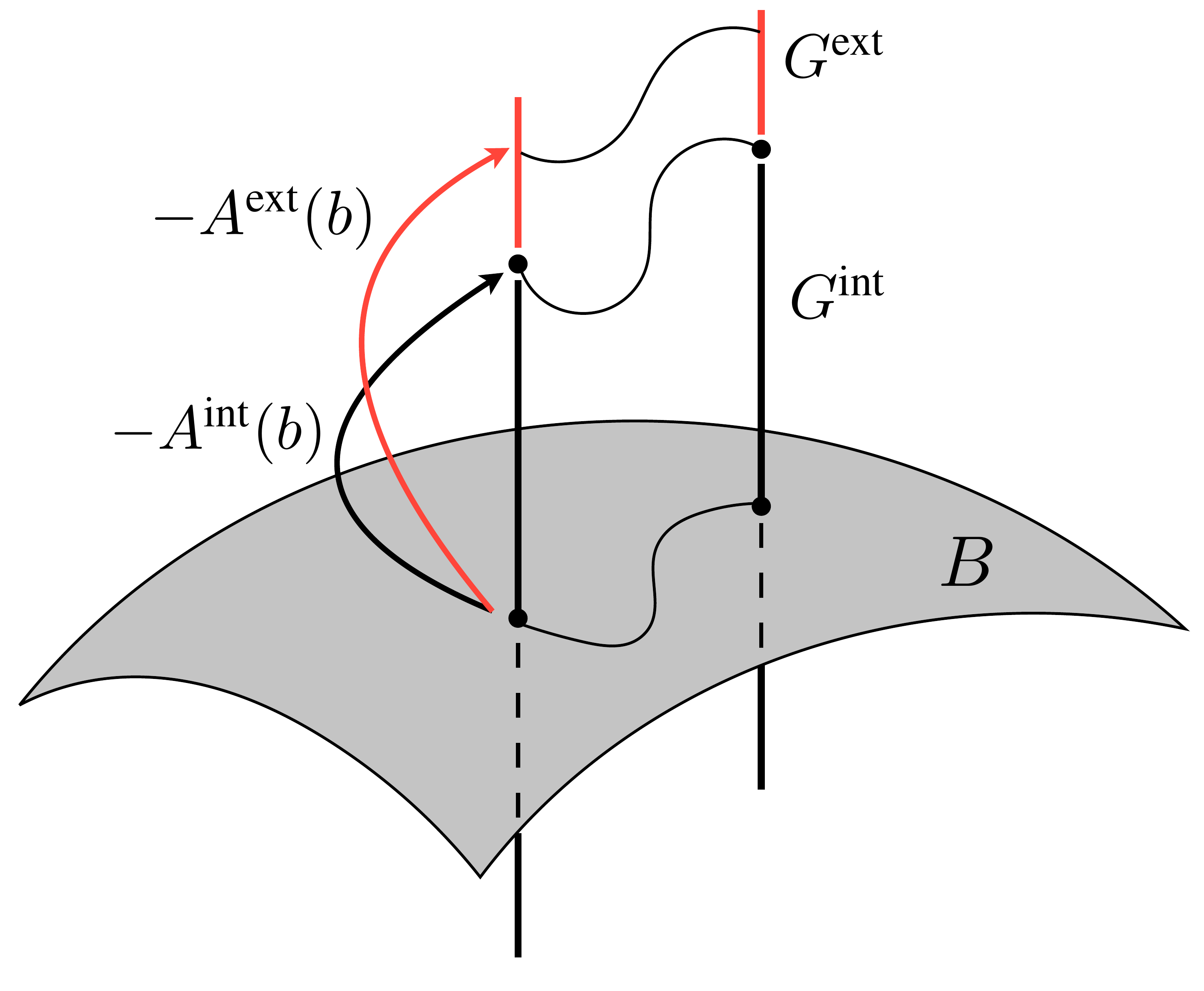}
	\end{center}
	\caption{A stratified fiber bundle with an \textit{internal} fiber $G^\text{int}$ and \textit{external} fiber $G^\text{ext}$, from which controls determine base trajectories through the inverse mapping of $-A^\text{ext}$, which are then lifted back to $G^\text{int}$ via the connection $-A^\text{int}$.}
	\label{fig:bundle-stratified}
\end{figure}
%%%%%%%%%%%%%%%% end figure %%%%%%%%%%%%%%%%%%% 
%%%%%%%%%%%%%%%%%%%%%%%%%%%%%%%%%%%%%%%%%%%%%%%%%%%%%%%%%%%%%%%%%%%%%%

For this type of system, the fiber space takes on a stratified structure $G = G^\text{int} \times G^\text{ext}$. We separately identify fiber variables associated with the configuration of the system $g^\text{int} \in G^\text{int}$ making up an \emph{internal fiber}, describing the configuration of the robot, from those describing the configuration of the platform $g^\text{ext} \in G^\text{ext}$, the \emph{external fiber}. This is visualized in Fig.~\ref{fig:bundle-stratified} as two differently colored fiber components. 

Note that for systems for which locomotion is governed by a principal connection, the original fiber bundle is unchanged and we retain our original connection $-A^\text{int}(b)$. Furthermore, a second connection $-A^\text{ext}(b)$ from the robot's base configuration to the medium's configuration can be derived in a manner analogous to $-A^\text{int}(b)$, and takes the form
\begin{align}
\begin{bmatrix} \dot g^\text{int} \\ \dot g^\text{ext} \end{bmatrix} = -T_e L_g \left(
\begin{bmatrix} A^\text{int}(b) \\ A^\text{ext}(b) \end{bmatrix} \dot b \right).
\end{align}

The structures of the two connections $-A^\text{int}(b)$ and $-A^\text{ext}(b)$ can both be analyzed in the same visual manner as shown in Fig.~\ref{fig:TL-ED}. For example, if we care only about the robot's fiber motion in response to certain joint trajectories without regard to how the platform moves, then we need only look at $-A^\text{int}(b)$, and the problem is unchanged from before. On the other hand, looking at $-A^\text{ext}(b)$ tells us how the robot's joint trajectories lead to motion of the \emph{platform}. 
% An interesting optimization problem here is that of trying to effect certain fiber motion in both $G^\text{int}$ and $G^\text{ext}$, which would involve simultaneous search of base trajectories on both exterior derivative plots. This may be considered in future work.

In this paper, and in particular for the three-link robot on a platform system, we will consider both the traditional problem of mapping base trajectories to the external fiber, as well as the problem of having the controls in the external fiber $G^\text{ext}$, with the base variables remaining passive. 
% As shown in Fig.~\ref{fig:bundle-stratified}, we can generally find a relationship between trajectories in $G^\text{ext}$ and trajectories in the base space $B$. 
Once we know our trajectories in $B$, the original connection $-A^\text{int}(b)$ ultimately lifts them to the internal fiber $G^\text{int}$. The problem of finding platform inputs in $G^\text{ext}$ to produce a set of desired base trajectories in $B$ is more difficult, as the ``inverse'' mapping of $-A^\text{ext}(b)$ may not be well defined. In this paper we consider only cases for which the dimension of the external actuation is equivalent to the dimension of the base manifold, ensuring $A^\text{ext}(b)$ is invertible.

\subsection{Fibers Without Symmetries}
In the three-link robot system, the preservation of symmetries holds as long as all quantities are prescribed or expressed relative to the body-fixed frame of the robot. If this is not the case, \textit{e.g.}, the platform velocity inputs are prescribed relative to an inertial frame, then one of the fiber components, the robot's orientation $\theta$, breaks this symmetry. The decision to keep all quantities relative to an inertial frame may be one of practicality, especially when considering multiple distinct locomoting robots in an ambient medium.

%%%%%%%%%%%%%%%%%%%%%%%%%%%%%%%%%%%%%%%%%%%%%%%%%%%%%%%%%%%%%%%%%%%%%%
%%%%%%%%%%%%%%%% begin figure %%%%%%%%%%%%%%%%%%%
\begin{figure}[t]
	\begin{center}
		\includegraphics[width=.42\textwidth]{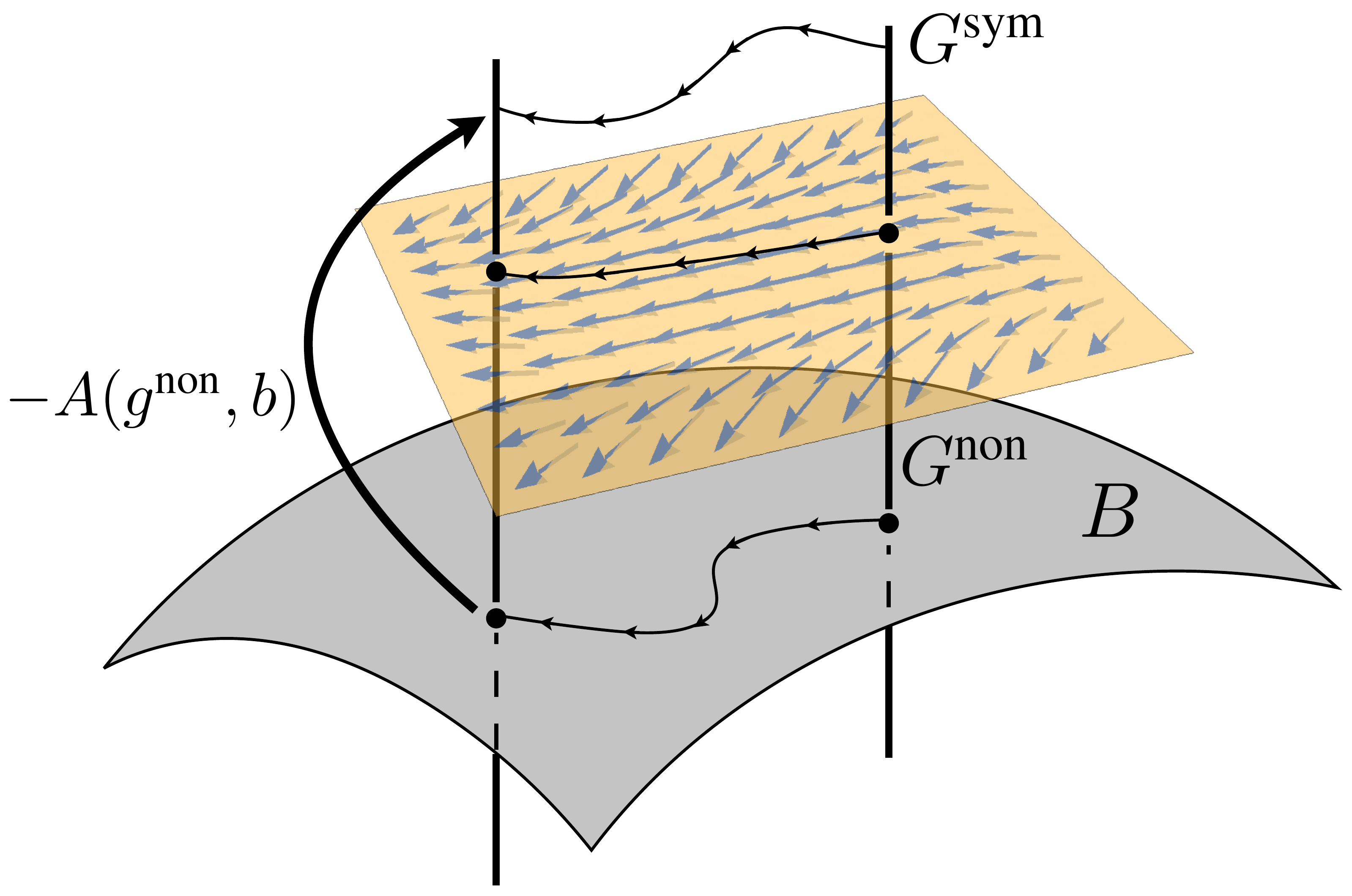}
	\end{center}
	\caption{Trajectories in the base space $B$ determine the dynamics through which the ``non-symmetric'' fiber components in $G^\text{non}$ evolve. Both $B$ and $G^\text{sym}$ trajectories are then mapped together through the connection $-A$ to trajectories in $G^\text{sym}$.}
	\label{fig:no-symmetry} 
\end{figure}
%%%%%%%%%%%%%%%% end figure %%%%%%%%%%%%%%%%%%% 
%%%%%%%%%%%%%%%%%%%%%%%%%%%%%%%%%%%%%%%%%%%%%%%%%%%%%%%%%%%%%%%%%%%%%%

% In general, this situation may also arise even if the ambient medium remains static but contains pose-dependent disturbances that act on the system. Alternatively, the medium may facilitate interactions among multiple systems, but these interactions take place in a local manner, for example depending on the relative configurations of two neighboring swimmers. 

In all of these cases, the system's internal dynamics will no longer be symmetric with respect to a subset of the fiber degrees of freedom (\textit{e.g.}, $(x,y)$ in the case of a three-link robot on a platform). However, we may still be able to define a principal bundle on the base variables and the fibers that are symmetric. For example, a falling, reorienting robot is affected by gravity in one fiber direction (or two, if orientation is also considered). A visual representation is shown in Fig.~\ref{fig:no-symmetry}; like Fig.~\ref{fig:bundle-stratified} we see fibers atop a base space, the former of which is subdivided into two distinct components. Trajectories in a non-symmetric fiber component evolve dynamically depending on trajectories in the base. This evolution is depicted as flow along a vector field in the semi-transparent yellow plane in Fig. ~\ref{fig:no-symmetry}. Both trajectories are then mapped through a connection together to obtain trajectories in the symmetric fiber component.

Mathematically, a new connection mapping $-A^\text{sym}$ from the base variables to all of the configuration variables would depend explicitly on the fiber variables that lack symmetries. That is, if $g^\text{sym}$ are the fiber variables of which the dynamics are independent, and $g^\text{non}$ are the external variables that are not (previously we had $g^\text{int}$ and $g^\text{ext}$ for internal and external fiber variables, respectively), then 
\begin{equation}
\begin{bmatrix} \dot g^\text{sym} \\ \dot g^\text{non} \end{bmatrix} = -T_e L_g \left(
\begin{bmatrix} A^\text{sym}(g^\text{non}, b) \\ A^\text{non}(g^\text{non}, b) \end{bmatrix} \dot b \right).
\label{eq:asymmetric}
\end{equation}

This splitting between $g^\text{sym}$ and $g^\text{non}$ is conducive to studying motion planning for the three-link snake robot on a platform when considering the platform fibers to be relative to an inertial reference frame, and not coinciding with the body frame of the robot. The first involves being able to \emph{decouple} the two lines of Eq.~\eqref{eq:asymmetric}. If it is the case that $\dot g^\text{sym}$ depends very weakly on $g^\text{non}$ or if $g^\text{non}$ evolves on a significantly slower time scale relative to $b$ and $g^\text{sym}$, then one can approximate the first line of Eq.~\eqref{eq:asymmetric} as a form of Eq.~\eqref{eq:fiber-reconstruction-eq}, where any $g^\text{non}$ dependencies are made constant and inputs are prescribed in $b$. The second line of Eq.~\eqref{eq:asymmetric}, which deals with the evolution of $g^\text{non}$, can then be treated separately from the first as it has no dependencies on $g^\text{sym}$.

If such a decoupling cannot be done, we would have to work with the full form of Eq.~\eqref{eq:asymmetric}. In this case, we can define $B^\text{non} = G^\text{non} \times B$ as an \emph{extended base space} in which we have partial control over the base variables. We treat $g^\text{non}$ as if they are passive base variables, and we can first analyze their kinematics from the second line of Eq.~\eqref{eq:asymmetric} using techniques that we have already described. Once we have solved for the motion involving base and non-symmetric fibers only, it is straightforward to push through the original connection in the first line of Eq.~\eqref{eq:asymmetric} to find the rest of the system's fiber motion.

In summary, we have identified two different approaches for accommodating external actuation into the principal fiber bundle picture. In the first, all external degrees of freedom are themselves symmetries of the entire system, so that we can consider them to be a separate fiber with their own connection mapping. In the second, these new degrees of freedom actually break symmetries, in which case we consider a splitting of the configuration space to differentiate between symmetric and non-symmetric variables, dealing with them separately. It is also possible to have systems that exhibit both types of splittings simultaneously.

%%%%%%%%%%%%%%%%%%%%%%%%%%%%%%%%%%%%%%%%%%%%%%%%%%%%%%%%%%%%%%%%%%%%%%
\section{Three-link Wheeled Snake Robot on a Platform}
\label{sec:snake}
We now describe the three-link snake robot of Fig.~\ref{fig:snakeRobot} in more detail, deriving the mechanical models corresponding to its locomotion in isolation as well as on a movable platform. This example will illustrate the stratified fiber bundle structure introduced in the previous section, as well as our approach to the problem of locomotion with controls applied externally (platform) rather than in the base (robot joints). Note that in our analysis of the three-link wheeled snake robot and the Chaplygin beanie systems on a platform, we consider the platform to not possess a rotational degree of freedom.

\subsection{Robot Description}
Recall that the configuration space of the robot can be written as $Q_w = G^\text{int}_w \times B_w$, where $g^\text{int}_w = (x,y,\theta)^T \in G^\text{int}_w$ are the fiber variables describing the robot's position and orientation, and $b_w = (\alpha_1, \alpha_2)^T \in \mathbb T^2$ are the base variables describing the robot's joint angles. We designate the constant parameter $R$ for the link lengths of the robot, which are identical across all links.

The kinematics of the system are described by the set of nonholonomic constraints on the wheels, which prohibit motion perpendicular to each of the links' longitudinal directions. They can be written as three equations of the form 
\begin{equation}
-\dot x_i \sin \theta_i + \dot y_i \cos \theta_i = 0,
\label{eq:gen-constraint}
\end{equation}
where $(\dot x_i, \dot y_i)$ is the velocity and $\theta_i$ is the inertial orientation of the $i$th link. These quantities can be computed recursively to express them as functions of $q_w$. Starting with the proximal link, we have $(x_1, y_1, \theta_1) = (x, y, \theta)$; for $i = 2, 3$: 
\begin{align}
\theta_i &= \theta_{i-1} + \alpha_{i-1}, \nonumber \\
x_i &= x_{i-1} + \frac{R}{2} (\cos \theta_{i-1} + \cos\theta_i), \nonumber \\
y_i &= y_{i-1} + \frac{R}{2} (\sin \theta_{i-1} + \sin\theta_i). 
\label{eq:kinematics}
\end{align}

The constraint equations are symmetric with respect to the fiber part $G_w$ of the configuration, since the kinematics do not explicitly depend on where the system is positioned or how it is oriented in space. We can thus rewrite the constraints in a reduced Pfaffian form as 
\begin{equation}
\omega_\xi(b_w) \xi + \omega_b(b_w) \dot b_w = 0,
\label{eq:pfaffian}
\end{equation}
where $\omega_\xi \in \mathbb R^{3 \times 3}$, $\omega_b \in \mathbb R^{3 \times 2}$, and $\xi = (\xi_x, \xi_y, \xi_\theta)^T \in \frak{se}(2)$ are the body velocities of the system expressed in a frame attached to the proximal link, as shown in Fig.\ \ref{fig:snakeRobot}. Recall that the velocities $\dot g_w$ and $\xi$ are related through $\dot g_w = T_e L_g \xi$, where $T_e L_g$ is the \emph{lifted left action} given by Eq.~\eqref{eq:TeLg}.

The reduced constraints of Eq.~\eqref{eq:pfaffian} can then be rearranged to derive the connection relationship of Eq.~\eqref{eq:reduced-reconstruction-eq}. For the three-link robot, this can be written as 
\begin{align}
\xi &= \!\frac{1}{D} \!\begin{bmatrix}
\cos\alpha_1 \! + \! \cos(\alpha_1-\alpha_2) \!\! & \! \! 1+\cos\alpha_1 \\ 0 & 0 \\
\frac{2}{R}(\sin\alpha_1 \!+\! \sin(\alpha_1\!-\!\alpha_2)) \!\! & \!\! \frac{2}{R}\sin\alpha_1
\end{bmatrix}\!\!\! 
\begin{bmatrix} \dot \alpha_1 \\ \dot \alpha_2 \end{bmatrix} \nonumber \\
& \triangleq -A^\text{int}_w(b_w) \dot b_w,
\label{eq:TL-recon}
\end{align}
where $D = \frac{2}{R} (-\sin \alpha_1 - \sin(\alpha_1 - \alpha_2) + \sin \alpha_2)$. The values of the second row, corresponding to $\xi_y$, are zero since this corresponds to the direction prohibited by the wheel of the proximal link. The mapping $-A^\text{int}_w(b_w)$ corresponds to the connection form between the base (robot joints) and the internal fiber (robot position and orientation). Visualizations of the exterior derivative of this connection form were shown in Fig.~\ref{fig:TL-ED}.

\subsection{Adding the Platform}

We now add in the dynamics of an underlying movable platform to the kinematics of the three-link wheeled snake robot. The position of the platform, which we denote by $(x_p, y_p)$, itself constitutes a symmetry group $G^\text{ext}_p$ since the modified system's properties do not depend on where the platform is located in space. The configuration manifold is now rewritten as $Q_p = G^\text{int}_w \times G^\text{ext}_p \times B_w = \mathrm{SE(2)} \times \mathbb R^2 \times \mathbb T^2$, although it is important to note that the connection relationship of Eq.~\eqref{eq:TL-recon} still holds, as the nonholonomic constraints Eq. \eqref{eq:pfaffian} depend only on the robot's pose and velocities relative to the platform, rather than an inertial frame.

Suppose that each of the links has identical mass $M^l$ and a moment of inertia $J$, while the platform has a mass $M^p$. Recalling that the inertial link positions and orientation are given by $(x_i,y_i,\theta_i)$, the system's Lagrangian is 
\begin{equation*}
L = \frac{1}{2} \sum_{i=1}^3 \left( M^l (\dot {x}_i^2 + \dot {y}_i^2) + J \dot \theta_i^2 \right) +  \frac{1}{2} M^p (\dot x_p^2 + \dot y_p^2).
\label{eq:lagrangian-simple}
\end{equation*}

Since the nonholonomic constraints are independent of $\dot x_p$ and $\dot y_p$, governing the trajectories of $(\dot x, \dot y, \dot \theta)$, the free directions of motion are simply the degrees of freedom of the platform. The conserved momenta are given by
\begin{align}
p = \begin{bmatrix} p_1 \\ p_2 \end{bmatrix} = 
\begin{bmatrix} \frac{\partial L}{\partial \dot x_p} \\ \frac{\partial L}{\partial \dot y_p}  \end{bmatrix} = \rho_g(b_w) \dot g_w + \rho_b(b_w) \dot b_w.
\label{eq:momenta-simple}
\end{align}
Noting that the form of Eq.~\eqref{eq:momenta-simple} is the same as the non-reduced form of Eq.~\eqref{eq:pfaffian}, we can rearrange to obtain 
\begin{equation}
\begin{bmatrix} \dot x_p \\ \dot y_p \end{bmatrix} =
-\begin{bmatrix} \cos\theta \! & \! -\sin\theta \\ \sin\theta \! & \! \cos\theta \end{bmatrix}
A^\text{ext}_p(b_w) \dot{b}_w + 
\frac{1}{M} \begin{bmatrix} p_1 \\ p_2 \end{bmatrix} .
\label{eq:simple-reconstruction}
\end{equation}
where $M = 3M^l + M^p$. 

Here, the matrix $-A^\text{ext}_p(b_w)$ is the external connection of our now stratified bundle structure. It is also the local form of the \emph{mechanical connection}, so named because it is derived from the conservation of momentum for the combined robot-platform system. In practice, we can drop the momentum drift terms if the platform starts at rest. The main complexity comes from the rotation matrix in front of the mechanical connection---the range of the connection is still the robot's Lie algebra, or velocities expressed in the robot-fixed frame. In the next subsection we first bypass this problem by assuming that we can prescribe platform velocities in the robot's body frame.

\subsection{Platform Actuation in Body Frame}

%%%%%%%%%%%%%%%%%%%%%%%%%%%%%%%%%%%%%%%%%%%%%%%%%%%%%%%%%%%%%%%%%%%%%%
%%%%%%%%%%%%%%%% begin figure %%%%%%%%%%%%%%%%%%%
\begin{figure}[t]
	\begin{center}
			\includegraphics[width=.33\textwidth]{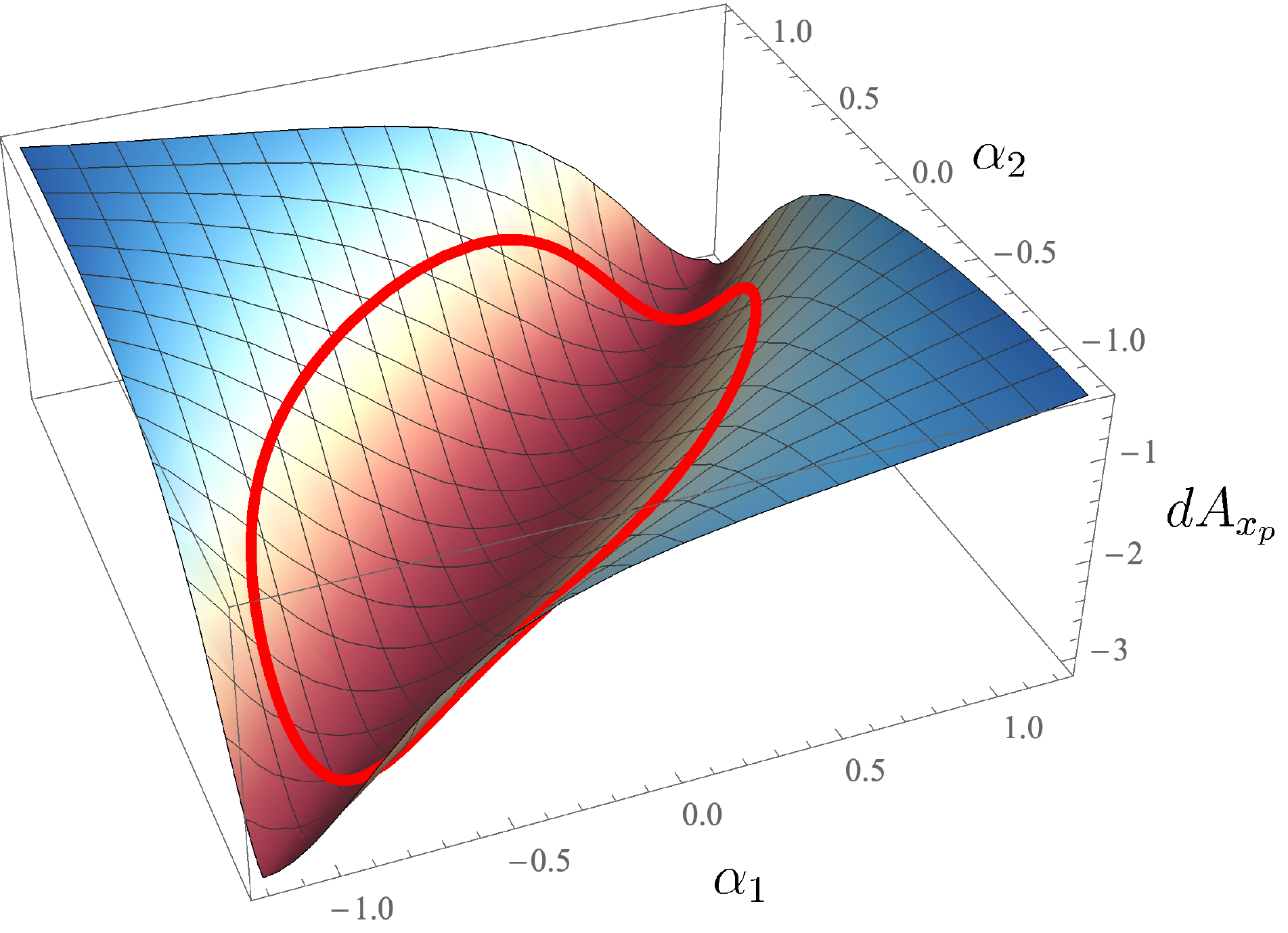}
			\includegraphics[width=.33\textwidth]{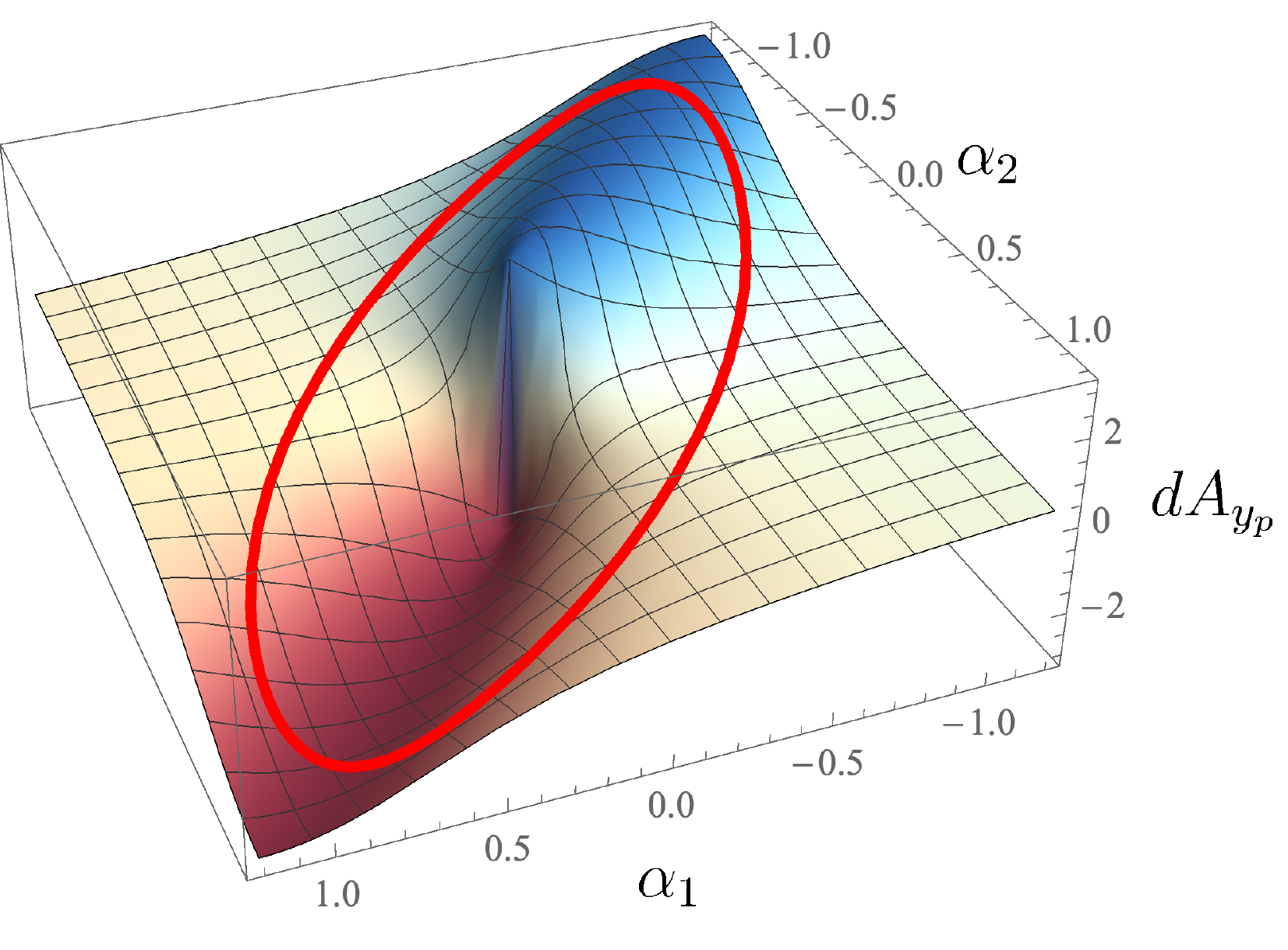}
	\end{center}
	\caption{The $u_p$ (top) and $v_p$ (bottom) components of the connection exterior derivative for the platform; that is, the exterior derivative of the connection that relates the robot's joint velocities to the platform's velocities in the robot's body frame.}
	\label{fig:platform-curv}
\end{figure}
%%%%%%%%%%%%%%%% end figure %%%%%%%%%%%%%%%%%%% 
%%%%%%%%%%%%%%%%%%%%%%%%%%%%%%%%%%%%%%%%%%%%%%%%%%%%%%%%%%%%%%%%%%%%%%

In deriving Eq.~\eqref{eq:simple-reconstruction} we noted that the main difficulty in using it, even after dropping the drift terms using a start-from-rest assumption, is the presence of the rotation matrix. This relationship signifies a dependence on $\theta$, the orientation of the robot relative to the platform (and inertial frame). However, if we know or can prescribe the platform's position relative to the robot's body frame\footnote{Of course, this assumes that we have free control over the platform's motion in all directions, rather than along a set of fixed axes.}, we can directly use the external, mechanical connection $-A^\text{ext}_p(b_w)$ without accounting for the effect of $\theta$ on the robot-platform interaction. In other words, we can choose to specify the platform's velocity using the variables 
\begin{equation}
	\begin{bmatrix} \dot u_p \\ \dot v_p \end{bmatrix} = 
	\begin{bmatrix} \cos\theta & \sin\theta \\ -\sin\theta & \cos\theta \end{bmatrix}
	\begin{bmatrix} \dot x_p \\ \dot y_p \end{bmatrix}.
	\label{eq:platform-vels-body}
\end{equation}
Again assuming that the system starts from rest, Eq.~\eqref{eq:simple-reconstruction} then reduces to 
\begin{equation}
\begin{bmatrix} \dot u_p \\ \dot v_p \end{bmatrix} = -A^\text{ext}_p(b_w) \dot b_w .
\label{eq:external-conn}
\end{equation}
Since $-A^\text{ext}_p(b_w)$ plays a role identical to that of the kinematic connection for the robot, we can also plot its exterior derivative in the same way that we have done for the robot. Fig.~\ref{fig:platform-curv} shows their shape for a set of chosen system parameters. 

The main observation one finds from these plots is that the $u_p$ plot is in general a flipped version of the robot's $x$ exterior derivative plot (Fig.~\ref{fig:TL-ED}). This is reasonable; as the robot moves in the forward or backward direction, we would expect the platform to move in the opposite direction, preserving net initial momentum of zero. The exterior derivative for $v_p$ (motion of the platform in a direction lateral to the robot's heading) resembles the robot's $\theta$ exterior derivative (second plot of Fig.~\ref{fig:TL-ED}), but reflected about the $\alpha_1 = -\alpha_2$ axis. This can be physically interpreted as a counteraction of the platform due to rotation of the robot; as the platform does not have a rotation degree of freedom, the counteraction is projected onto the $v_p$ component.

With the derivation of Eq.~\eqref{eq:external-conn} we are now also able to consider the problem of actuating the platform to induce a completely passive robot to move in a desired way. In other words, we can prescribe platform velocities $(\dot u_p, \dot v_p)$, which would produce trajectories $b_w(t)$ in the robot's joint variables. These base trajectories (and the associated velocities) then result in the robot moving along its fiber degrees of freedom according to Eq.~\eqref{eq:TL-recon} and Fig.~\ref{fig:TL-ED}.

%%%%%%%%%%%%%%%%%%%%%%%%%%%%%%%%%%%%%%%%%%%%%%%%%%%%%%%%%%%%%%%%%%%%%%
%%%%%%%%%%%%%%%% begin figure %%%%%%%%%%%%%%%%%%%
\begin{figure}[t]
	\begin{center}		
			\includegraphics[width=.4\textwidth]{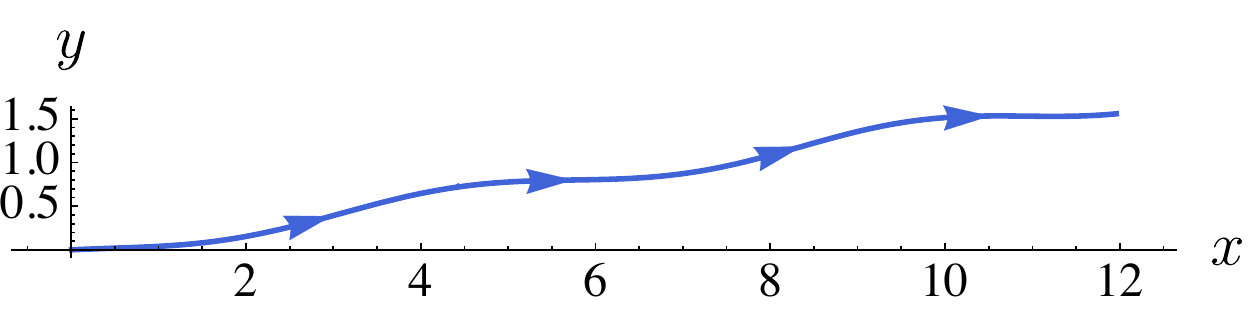}
			\includegraphics[width=.4\textwidth]{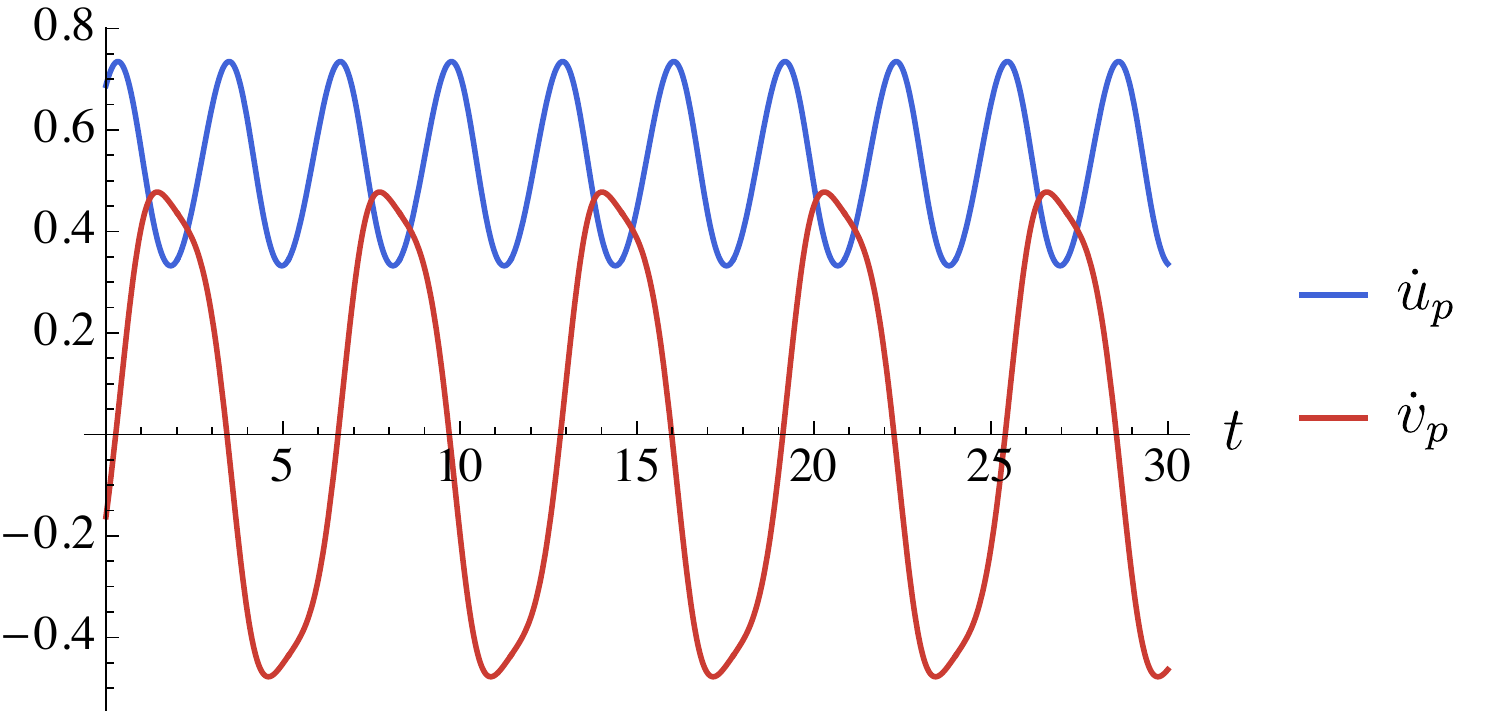}	
	\end{center}
	\caption{Top: The robot's desired fiber trajectory, which moves forward without reorientation. Bottom: The platform velocity inputs, relative to the robot's body frame, required to obtain the base gait in Fig.~\ref{fig:platform-curv} and ultimately the fiber trajectory above.}
	\label{fig:bodyPlatformControl}
\end{figure}
%%%%%%%%%%%%%%%% end figure %%%%%%%%%%%%%%%%%%% 
%%%%%%%%%%%%%%%%%%%%%%%%%%%%%%%%%%%%%%%%%%%%%%%%%%%%%%%%%%%%%%%%%%%%%%

This problem can be approached in two ways. The first is to assume that we have access to a particular robot fiber motion, either desired or measured, for which we can find a corresponding gait in the base space through analysis of the robot's original connection mapping in Eq.~\eqref{eq:TL-recon}. The same gait can also then be pushed through the platform connection relationship shown in Eq.~\eqref{eq:external-conn}. This would yield the required platform velocity inputs $\dot u_p(t)$ and $\dot v_p(t)$ that would generate the robot's overall desired motion. As an example, suppose we want to induce the robot to locomote as shown in the top figure of Fig.~\ref{fig:bodyPlatformControl}, in which the robot advances forward without reorientation. The gait that would achieve this is the gait shown as a closed loop in Fig.~\ref{fig:platform-curv}. Mapping this gait through the platform connection produces the required velocities as shown in the bottom figure of Fig.~\ref{fig:bodyPlatformControl}.

One drawback with the above approach is that there is no guarantee on the feasibility of platform input. This is reflected in Fig.~\ref{fig:bodyPlatformControl} in that the required input requires a constant nonzero offset in $\dot u_p(t)$, meaning that the platform would essentially have to move on the plane with the robot. The second, more control-oriented, way to approach this problem is to invert the mechanical connection in Eq.~\eqref{eq:external-conn} to find feasible $(\dot u_p, \dot v_p)$ trajectories to obtain a desired $\dot b_w$.

That Eq.~\eqref{eq:external-conn} is reduced, \textit{i.e.}, we have eliminated all robot fiber dependencies from the equation, allows us to consider only the interaction between the robot's joints and the platform fiber variables. Furthermore, the equation bears a resemblance to the dynamics between a set of actuated and a set of unactuated joints on the same robot, as detailed in \cite{dear2020}; the only difference is that the actuated variables in this case are external to the robot. 
%%%%%%%%%%%%%%%% end figure %%%%%%%%%%%%%%%%%%% 
%%%%%%%%%%%%%%%%%%%%%%%%%%%%%%%%%%%%%%%%%%%%%%%%%%%%%%%%%%%%%%%%%%%%%%
\begin{figure*}[t]
\centering
\subfloat{%
       \includegraphics[width=.33\textwidth]{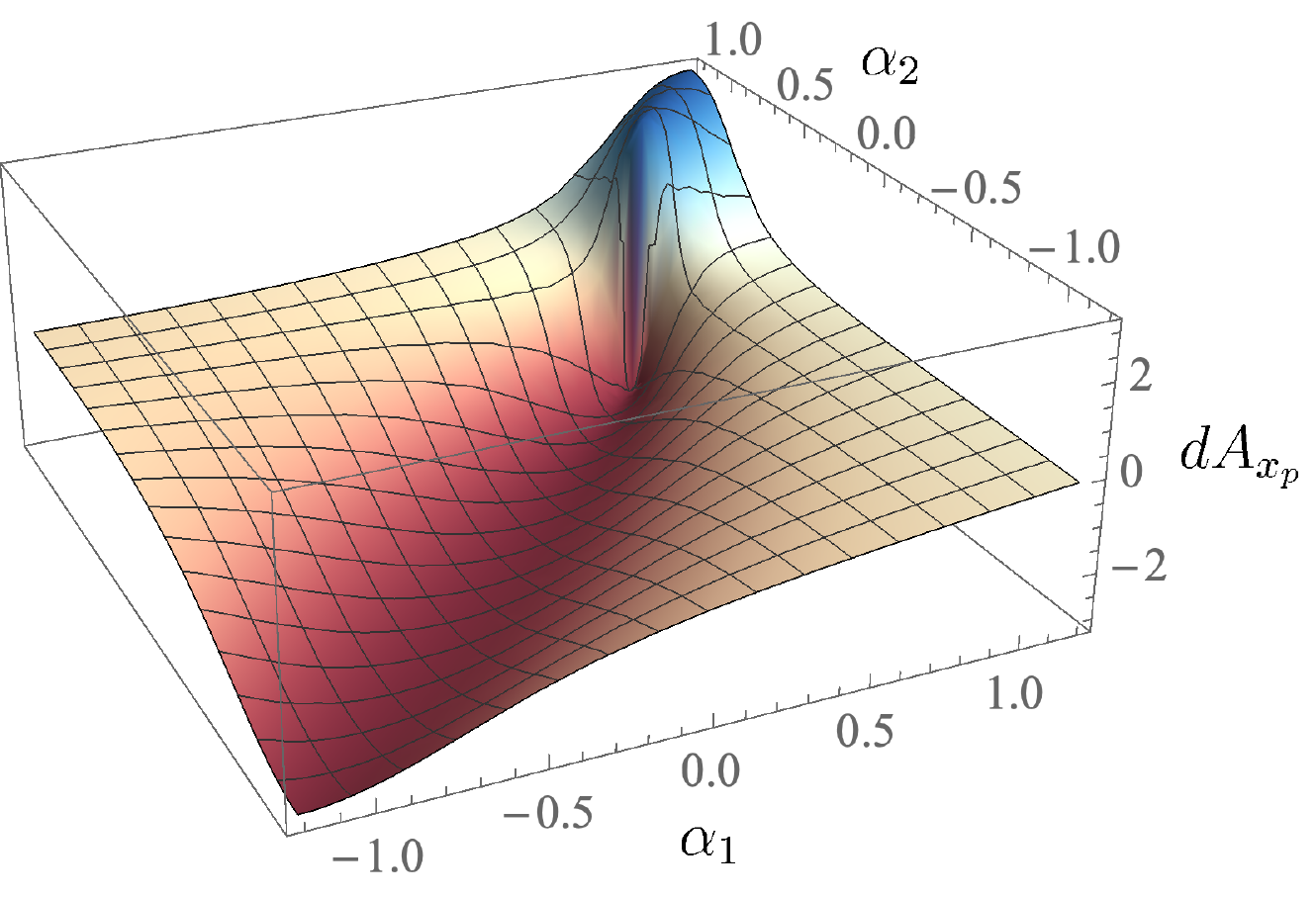}}
    \hfil
  \subfloat{%
        \includegraphics[width=.33\textwidth]{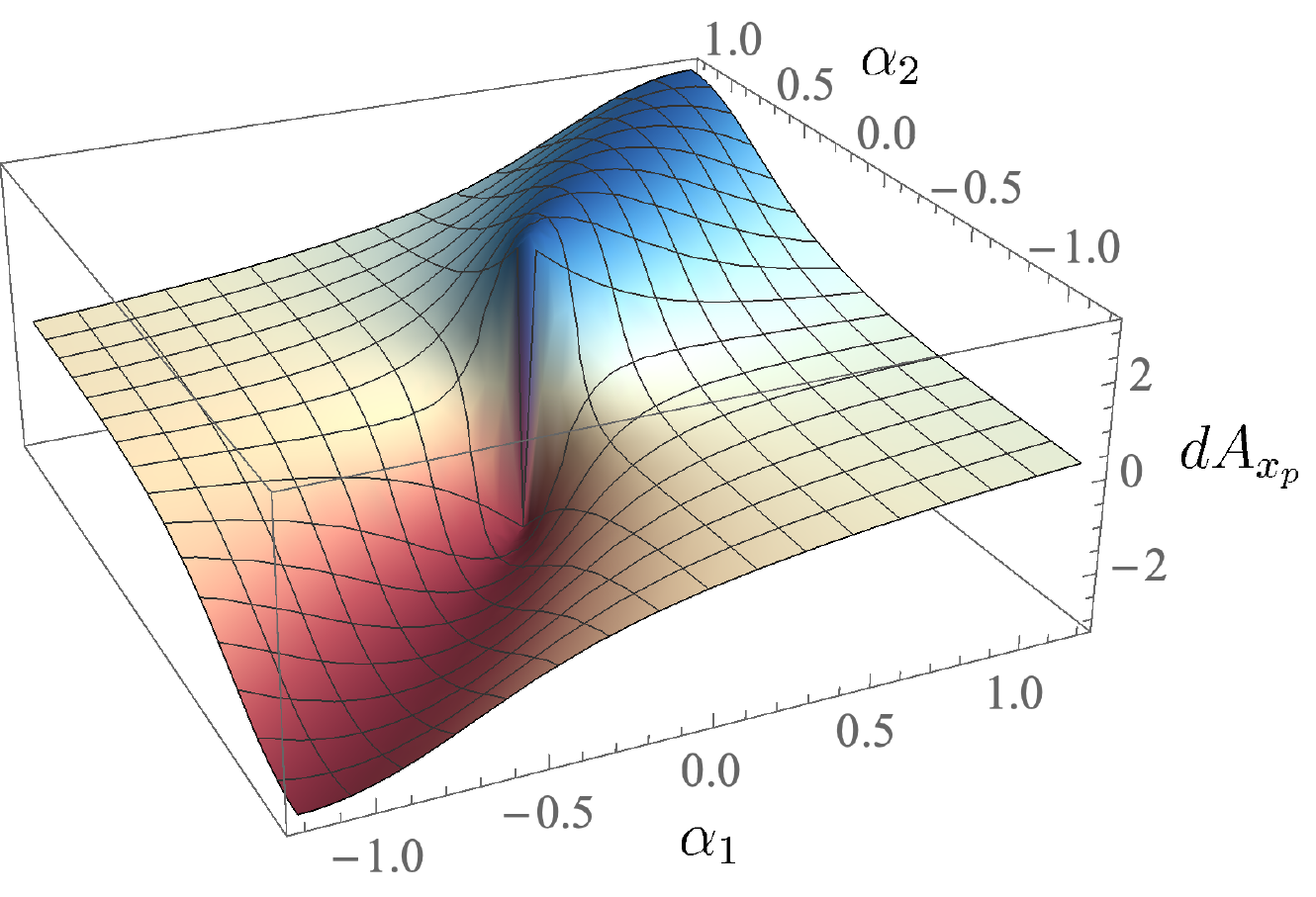}}
    \hfil
  \subfloat{%
        \includegraphics[width=.33\textwidth]{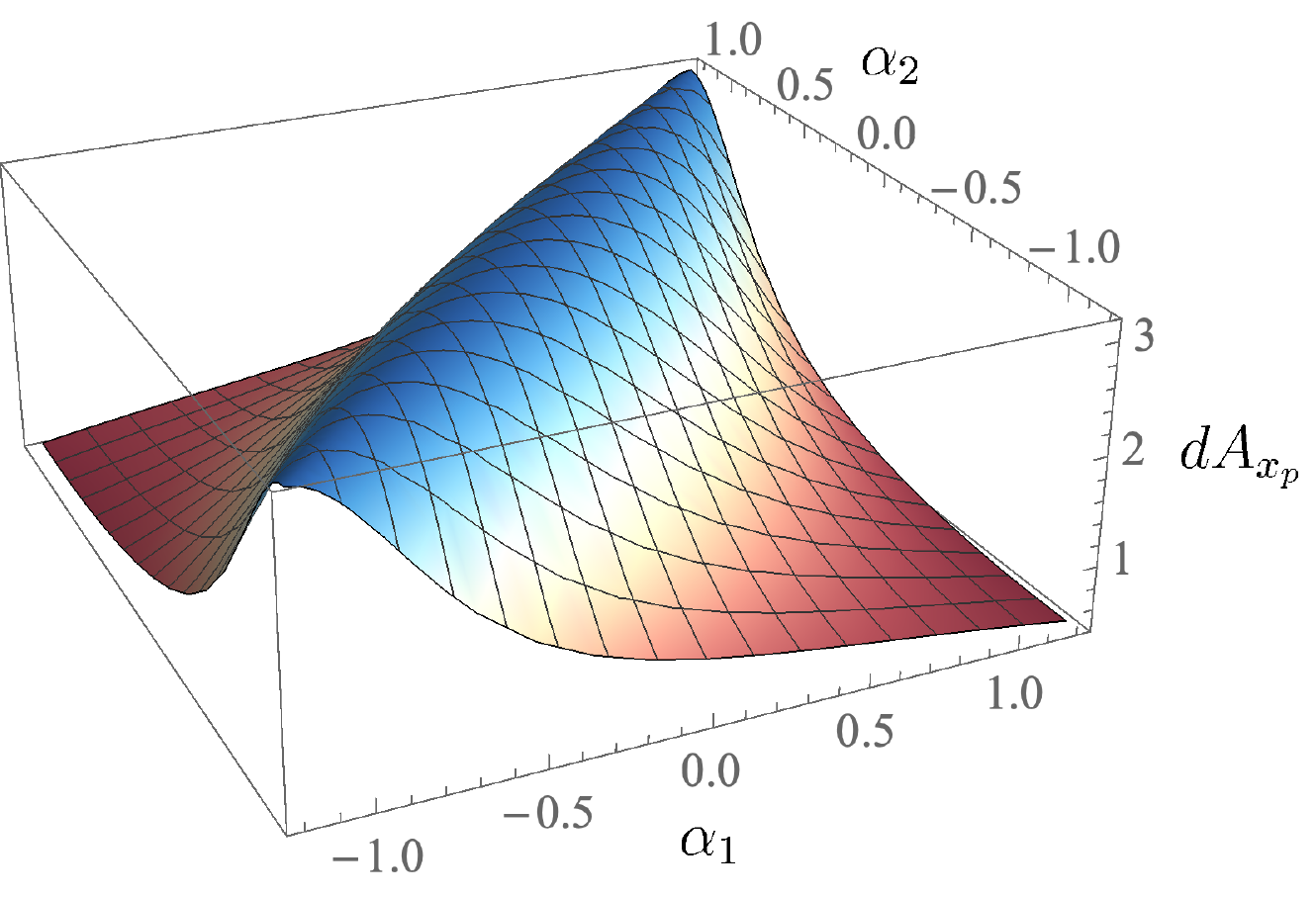}}
    \\
  \subfloat{%
       \includegraphics[width=.31\textwidth]{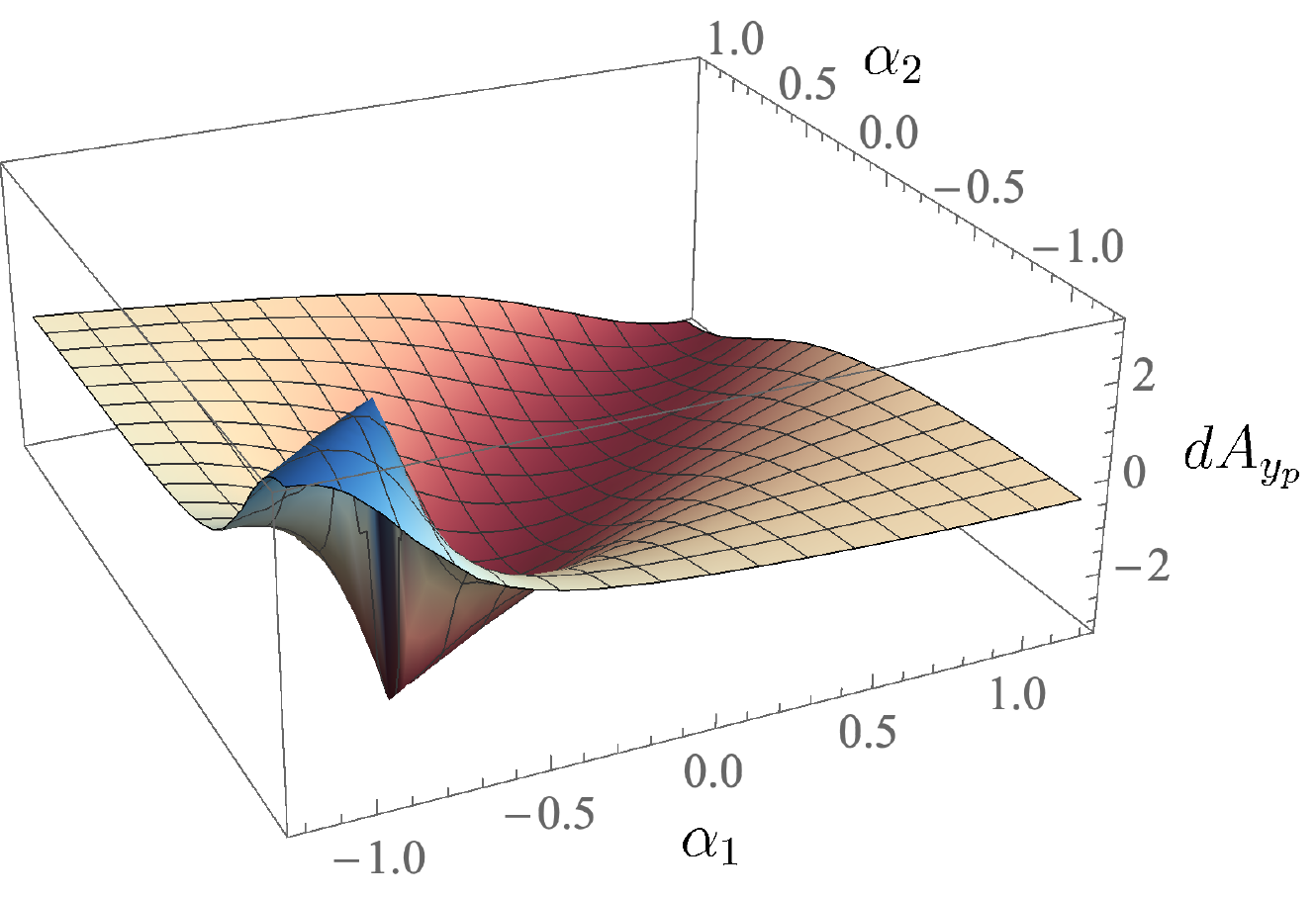}}
    \hfil
  \subfloat{%
        \includegraphics[width=.31\textwidth]{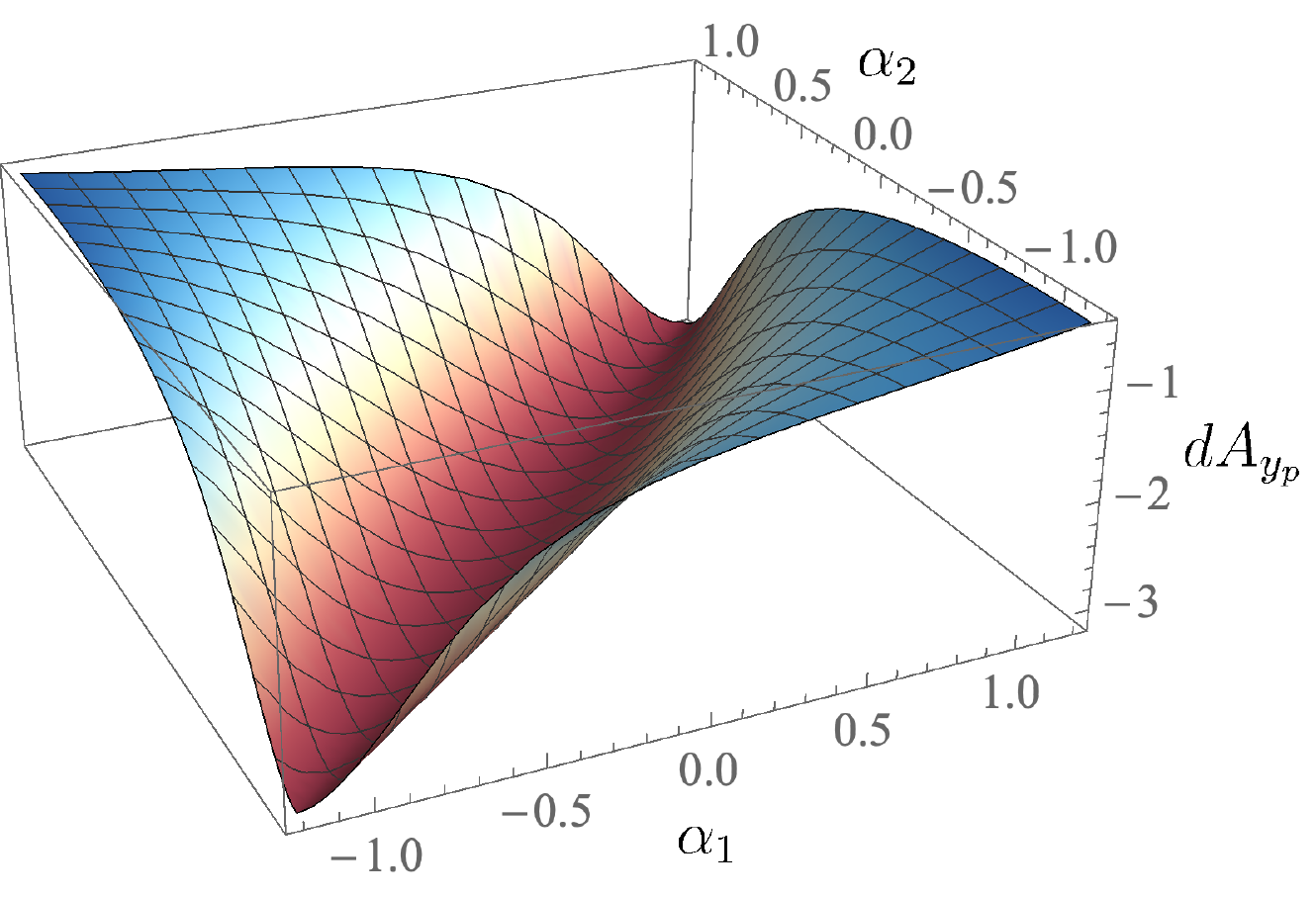}}
    \hfil
  \subfloat{%
        \includegraphics[width=.31\textwidth]{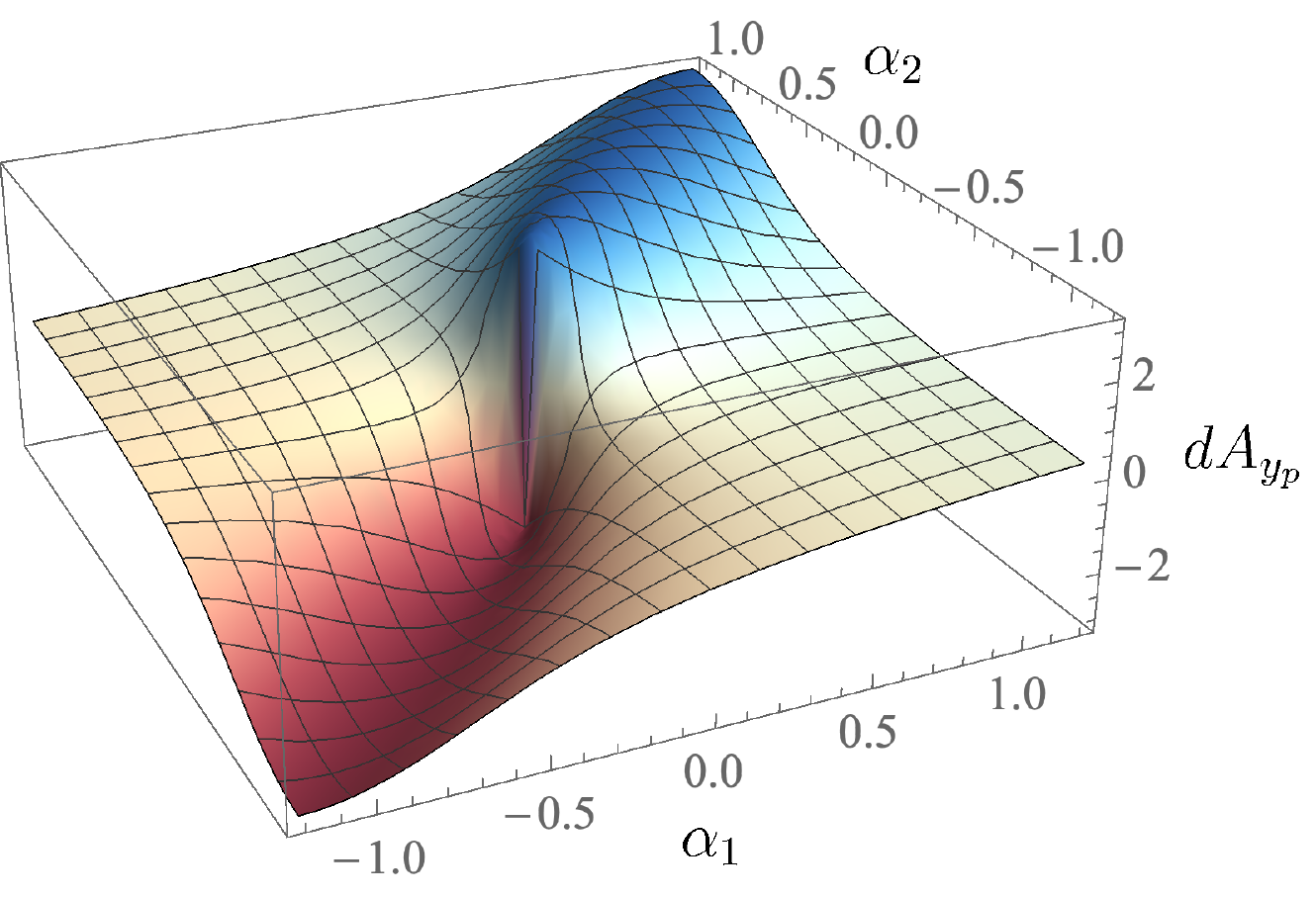}}
\caption{The $x_p$ (top) and $y_p$ (bottom) components of the connection exterior derivative for the platform, corresponding to the robot's orientation $\theta$ at 45 (left), 90 (middle), and 180 degrees (right). The case in which the robot is aligned with the platform ($\theta=0$) is the same as the plots in Fig.~\ref{fig:platform-curv}. Note that the plots undergo an inversion as $\theta$ increase from 0 to 180.}
\label{fig:rotationPlatformCurv}
\end{figure*}
%%%%%%%%%%%%%%%% end figure %%%%%%%%%%%%%%%%%%% 
%%%%%%%%%%%%%%%%%%%%%%%%%%%%%%%%%%%%%%%%%%%%%%%%%%%%%%%%%%%%%%%%%%%%%%
%%%%%%%%%%%%%%%%%%%%%%%%%%%%%%%%%%%%%%%%%%%%%%%%%%%%%%%%%%%%%%%%%%%%%%
%%%%%%%%%%%%%%%% begin figure %%%%%%%%%%%%%%%%%%%
\begin{figure}[t]
	\begin{center}
		\includegraphics[width=.44\textwidth]{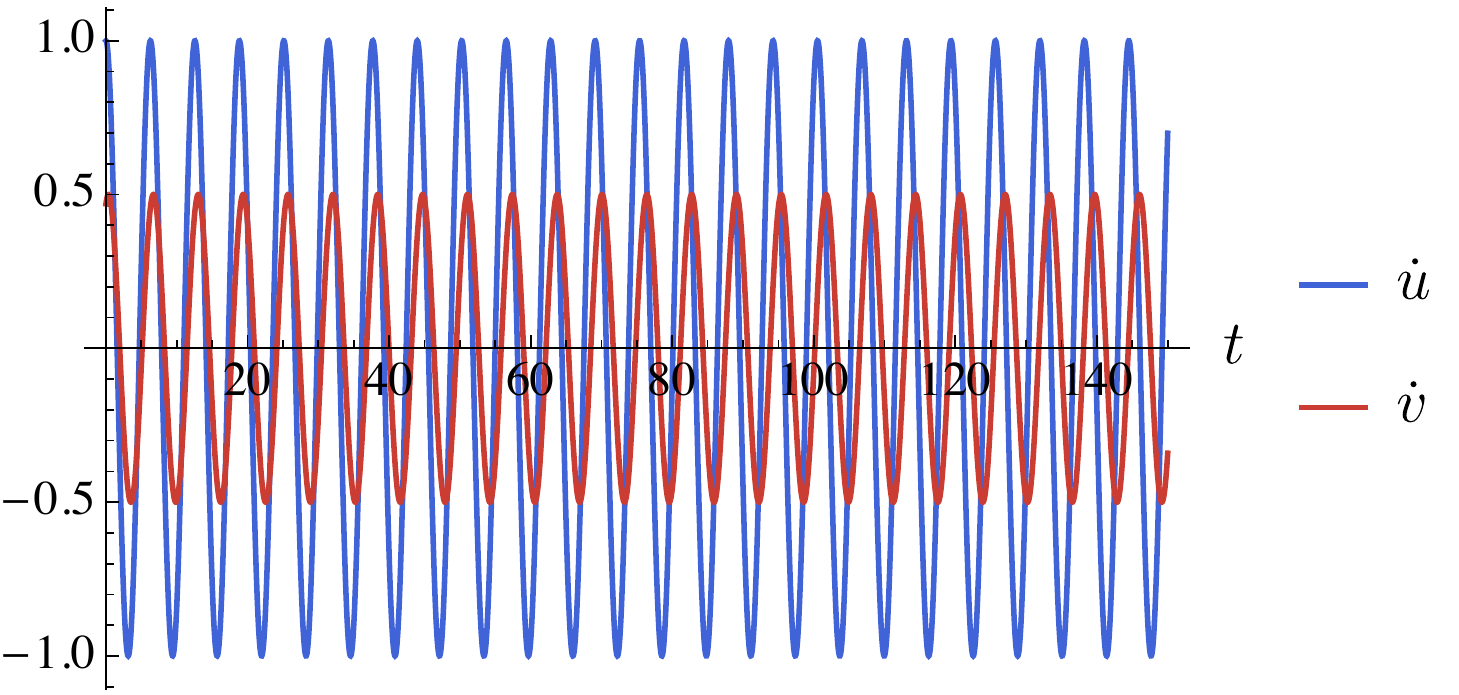}
		\includegraphics[width=.22\textwidth]{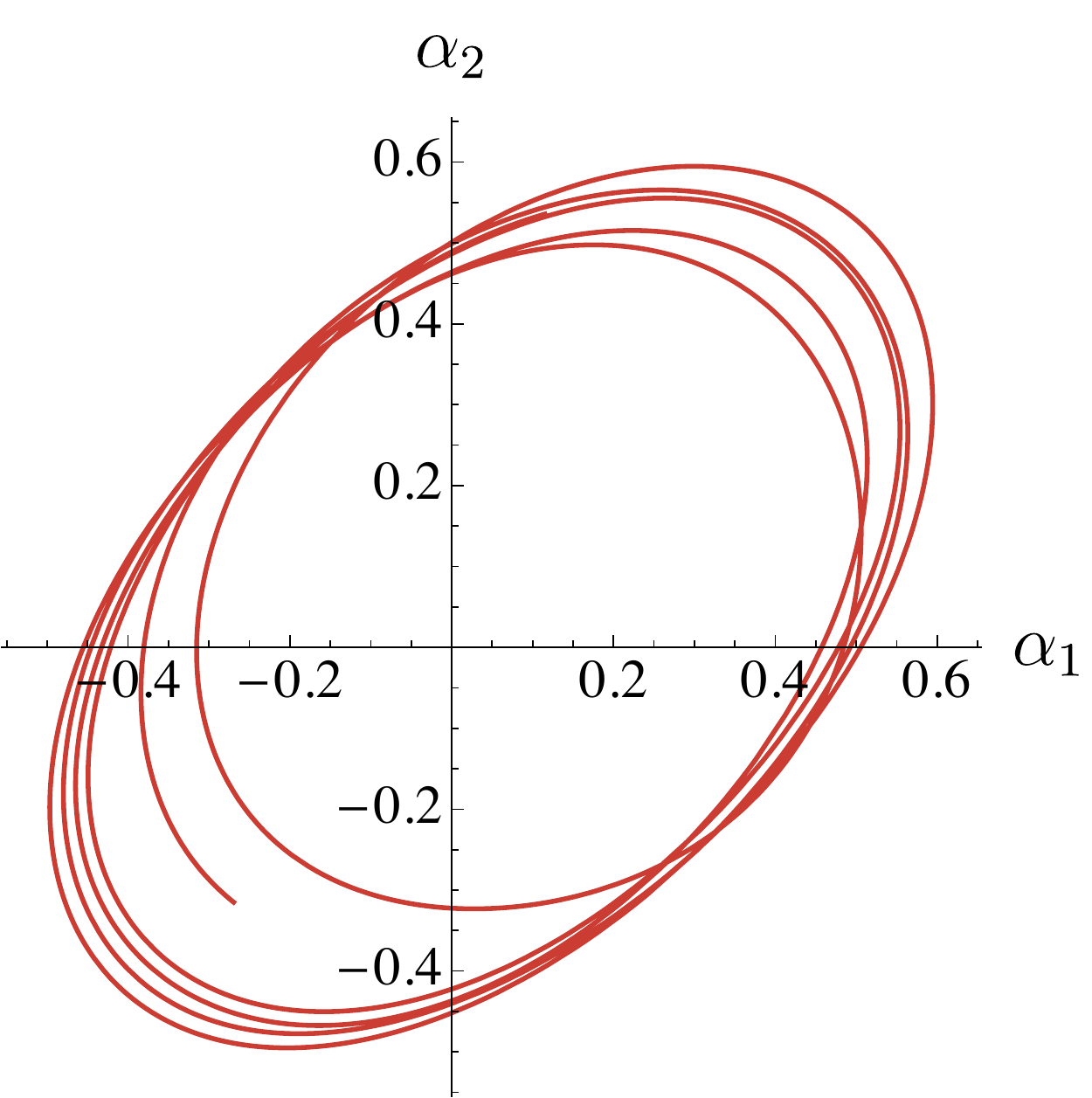}	
		\includegraphics[width=.22\textwidth]{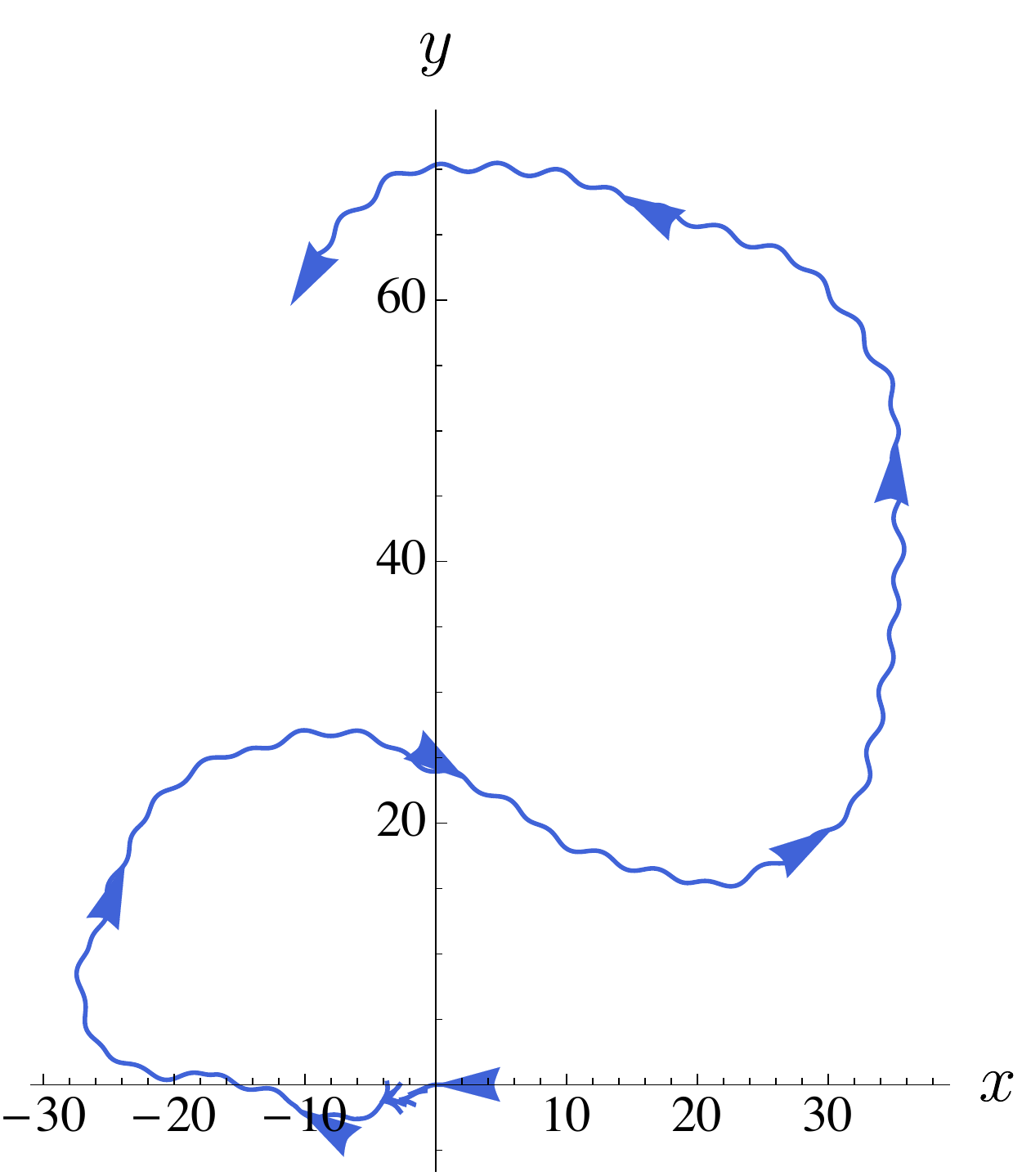}	
	\end{center}
	\caption{Top: A set of platform inputs that does not lead to net motion of the platform over time. Bottom left: The offset of the gait in shape space shifts from the third quadrant to the first. Bottom right: The resulting fiber motion of the robot, a trajectory with an increasing curvature.}
	\label{fig:UVPlatformFeedback}
	\vspace{-0.5cm}
\end{figure}

For example, it can be numerically shown that increasing the relative phase between the two input directions will increase the offset of the resultant joint trajectories. Such a gait is shown in Fig.~\ref{fig:UVPlatformFeedback} (top); note that $\dot v_p$ lags $\dot u_p$ more and more over time. This has the effect of shifting the center of the gait in shape space (Fig.~\ref{fig:UVPlatformFeedback}, bottom left) from the third quadrant to the first quadrant. As we know from the $\theta$ plot of Fig.~\ref{fig:TL-ED}, gaits with negative offsets in $\alpha_1$ and $\alpha_2$ acquire negative net orientation; gaits with positive offsets acquire positive net orientation. In addition, the robot continues to move forward in its body direction while rotating, since all gaits enclose a positive net area in the $x$ plot of Fig.~\ref{fig:TL-ED}. All of these observations are apparent in the robot's fiber trajectory for this simulation (Fig.~\ref{fig:UVPlatformFeedback}, bottom right). It starts at the origin moving forward on a trajectory rotating clockwise (negative curvature) and at around $t=60$ begins rotating counterclockwise (positive curvature), as the relative phase between $\dot u_p$ and $\dot v_p$ becomes large and the center of the resulting $\alpha_1$-$\alpha_2$ gait becomes positive.

\subsection{Platform Actuation in Inertial Frame}
In the previous subsection, the assumption that we can actuate the platform relative to the robot body's longitudinal and lateral directions requires that the platform can freely move in \emph{all} directions. An alternative scenario is that it may be limited to move only in two fixed directions, corresponding to two independent degrees of freedom. Assuming that the system starts from rest, this model is described exactly by Eq.~\eqref{eq:platform-vels-body}.

Even if the system again starts from rest, dropping the momenta terms from the equation, the dependency of Eq.~\eqref{eq:platform-vels-body} on the robot's orientation $\theta$ introduces an additional complexity. We assume that we know the robot's orientation throughout the system's operation, for example via an overhead camera, so that $\theta$ is not an unknown quantity. However, it is a symmetry-breaking fiber variable that changes the form of the connection $-A_p(b_w)$ (shown in the general case by Eq.~\eqref{eq:asymmetric}) and its corresponding exterior derivative plots in Fig.~\ref{fig:platform-curv}.

Fig.~\ref{fig:rotationPlatformCurv} shows the effect of $\theta$ on $\text{d}A_p(b_w)$; in other words, the plots show the exterior derivative, for different values of $\theta$, of a $\theta$-dependent connection
\begin{equation}
A^\text{non}_\theta(\theta, b_w) = \begin{bmatrix} \cos\theta & -\sin\theta \\ \sin\theta & \cos\theta \end{bmatrix}
A^\text{ext}_p(b_w).
\label{eq:theta-conn}
\end{equation}
The plots shown correspond to $\theta$ at $45$, $90$, and $180$ degrees (of course, the nominal plots of Fig.~\ref{fig:platform-curv} correspond to $\theta=0$). As we would expect, the nature of the interaction between the robot and platform changes with the orientation of the robot. As the robot rotates to $90$ degrees, its body $x$ axis points along the inertial $y$ axis, while its body $y$ axis points along the inertial $x$ axis. The $x_p$ plot is thus identical to $-v_p$, or an inverted version of the $v_p$ plot in Fig.~\ref{fig:platform-curv}, while $y_p$ is identical to $u_p$ (recall that the corresponding robot and platform components are inverted). At $180$ degrees, both $x_p$ and $y_p$ are completely inverted from $u_p$ and $v_p$, the latter again corresponding to $x_p$ and $y_p$ at $0$ degrees. Finally, as the robot reorients back to $0$ degrees, both plots smoothly deform back to their nominal forms (Fig.~\ref{fig:platform-curv}).

%%%%%%%%%%%%%%%%%%%%%%%%%%%%%%%%%%%%%%%%%%%%%%%%%%%%%%%%%%%%%%%%%%%%%%
%%%%%%%%%%%%%%%% begin figure %%%%%%%%%%%%%%%%%%%
\begin{figure*}[t]
	\begin{center}
		\includegraphics[width=.28\textwidth]{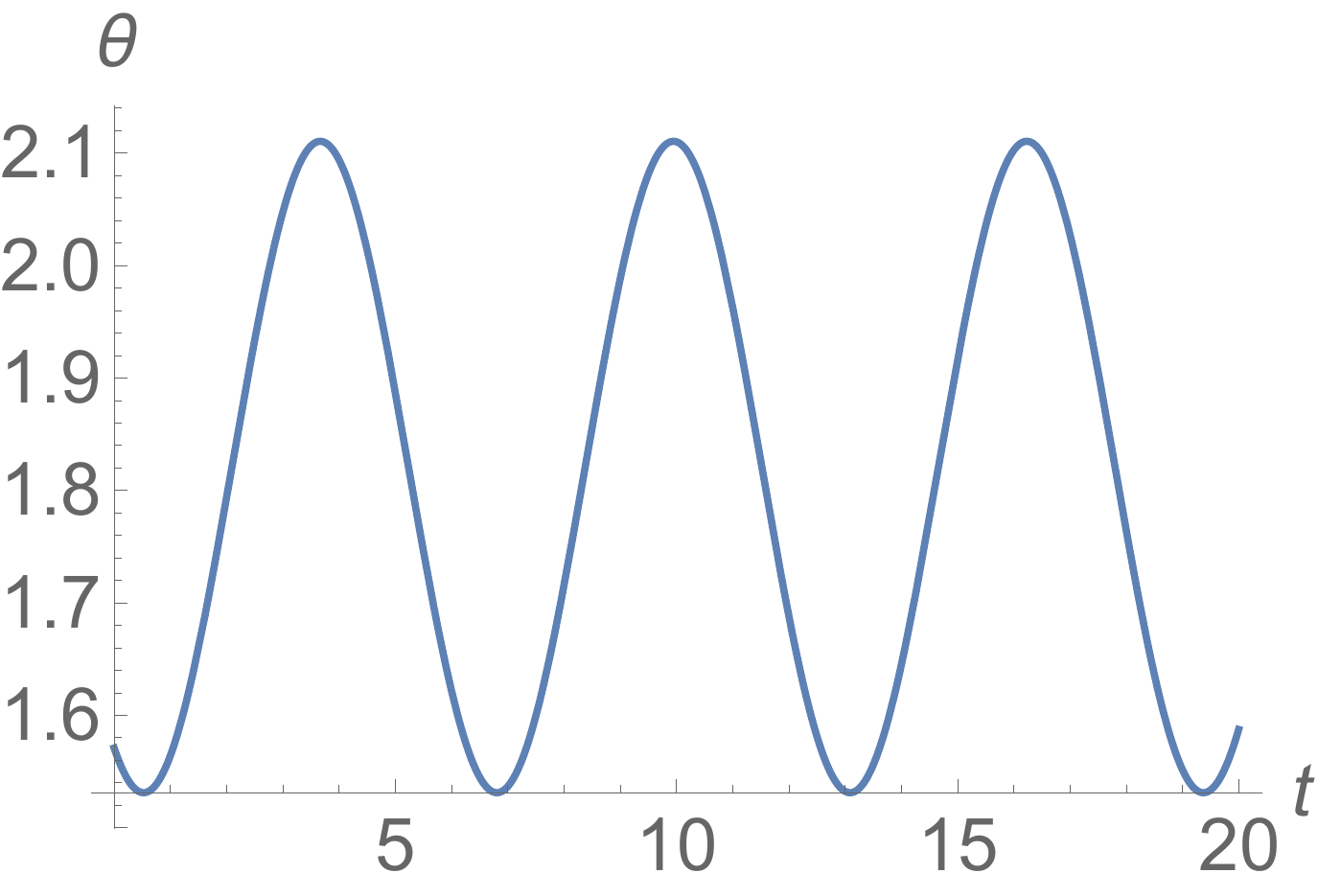}			
		\includegraphics[width=.35\textwidth]{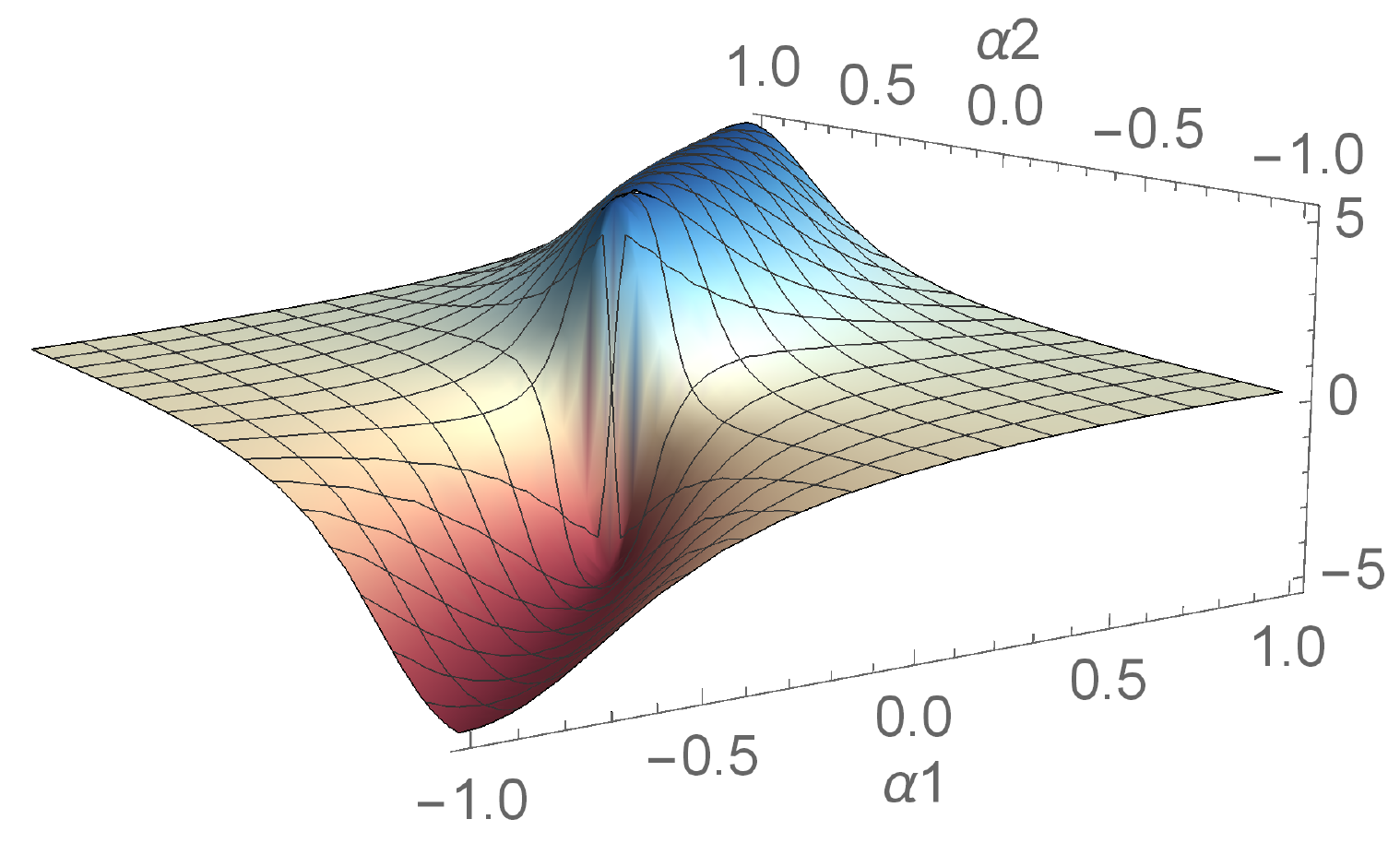}
		\includegraphics[width=.35\textwidth]{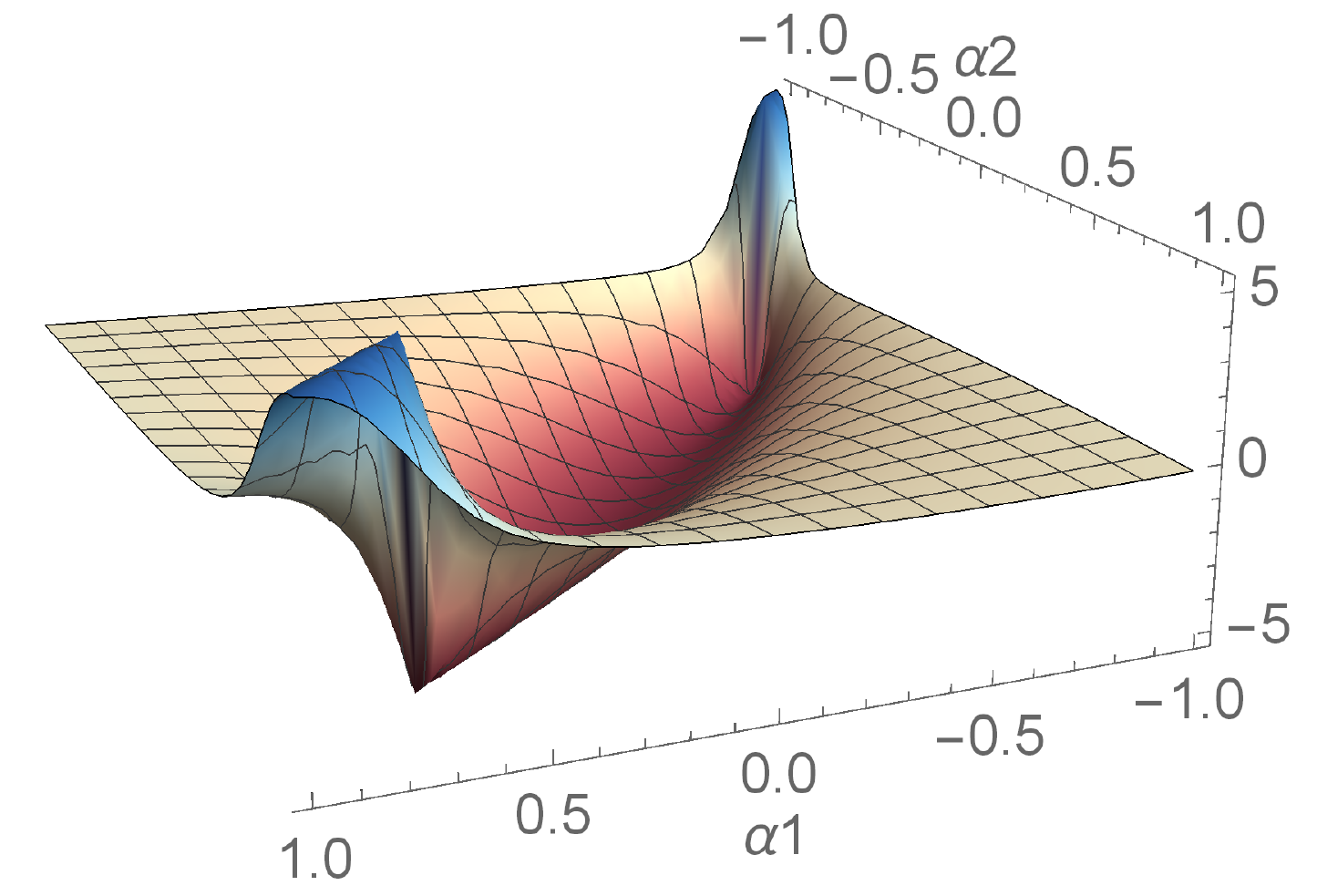}
		\includegraphics[width=.28\textwidth]{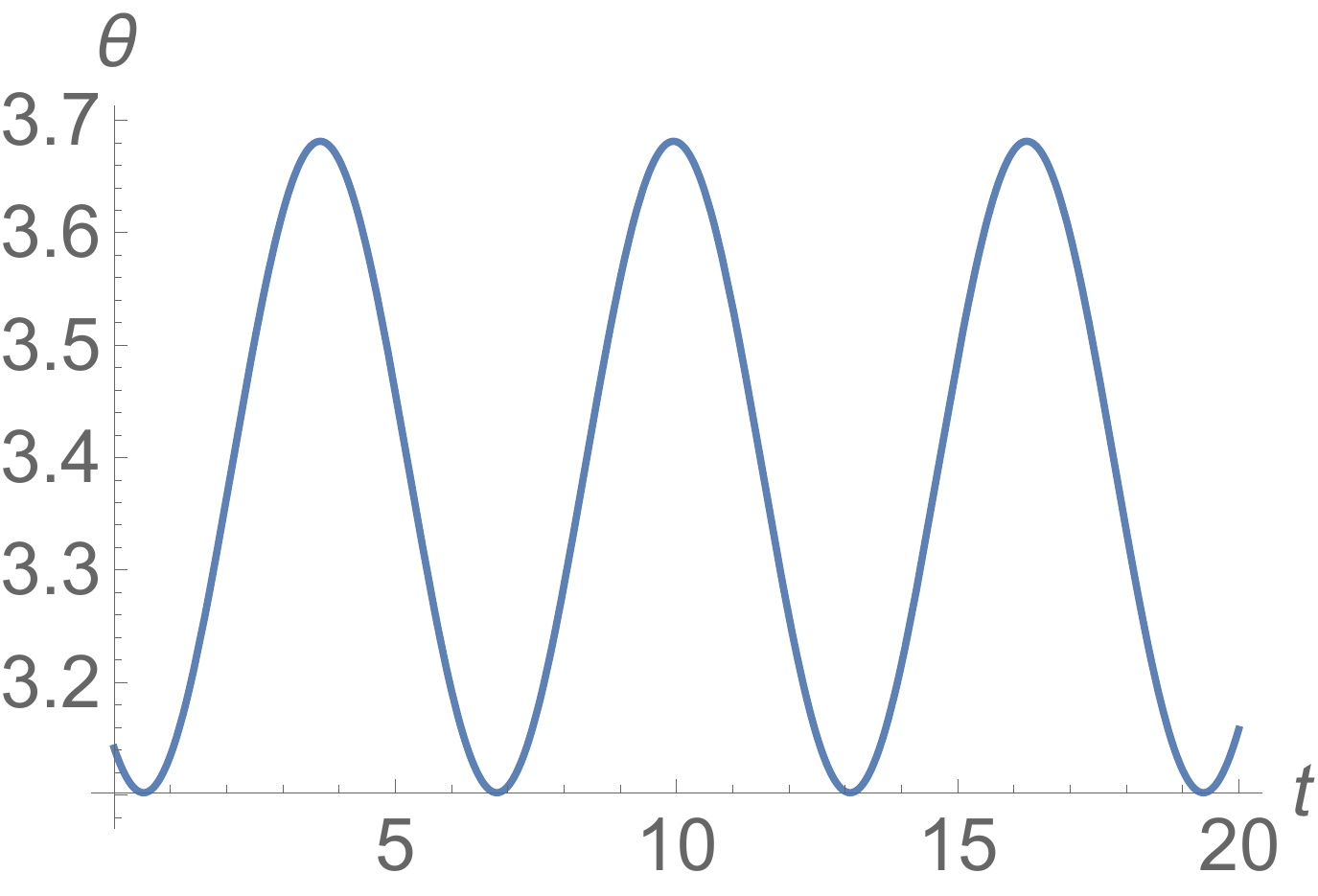}			
		\includegraphics[width=.35\textwidth]{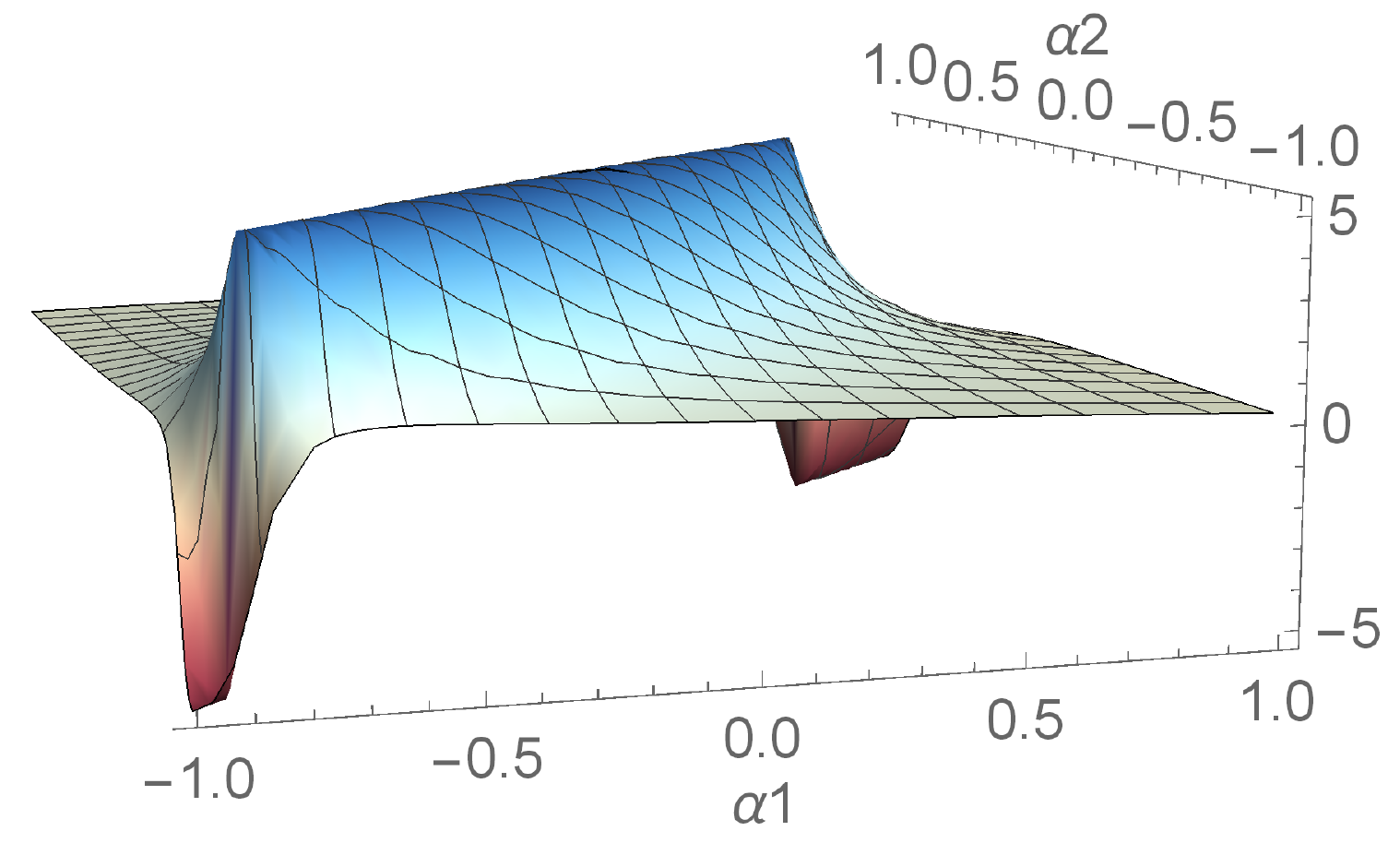}
		\includegraphics[width=.35\textwidth]{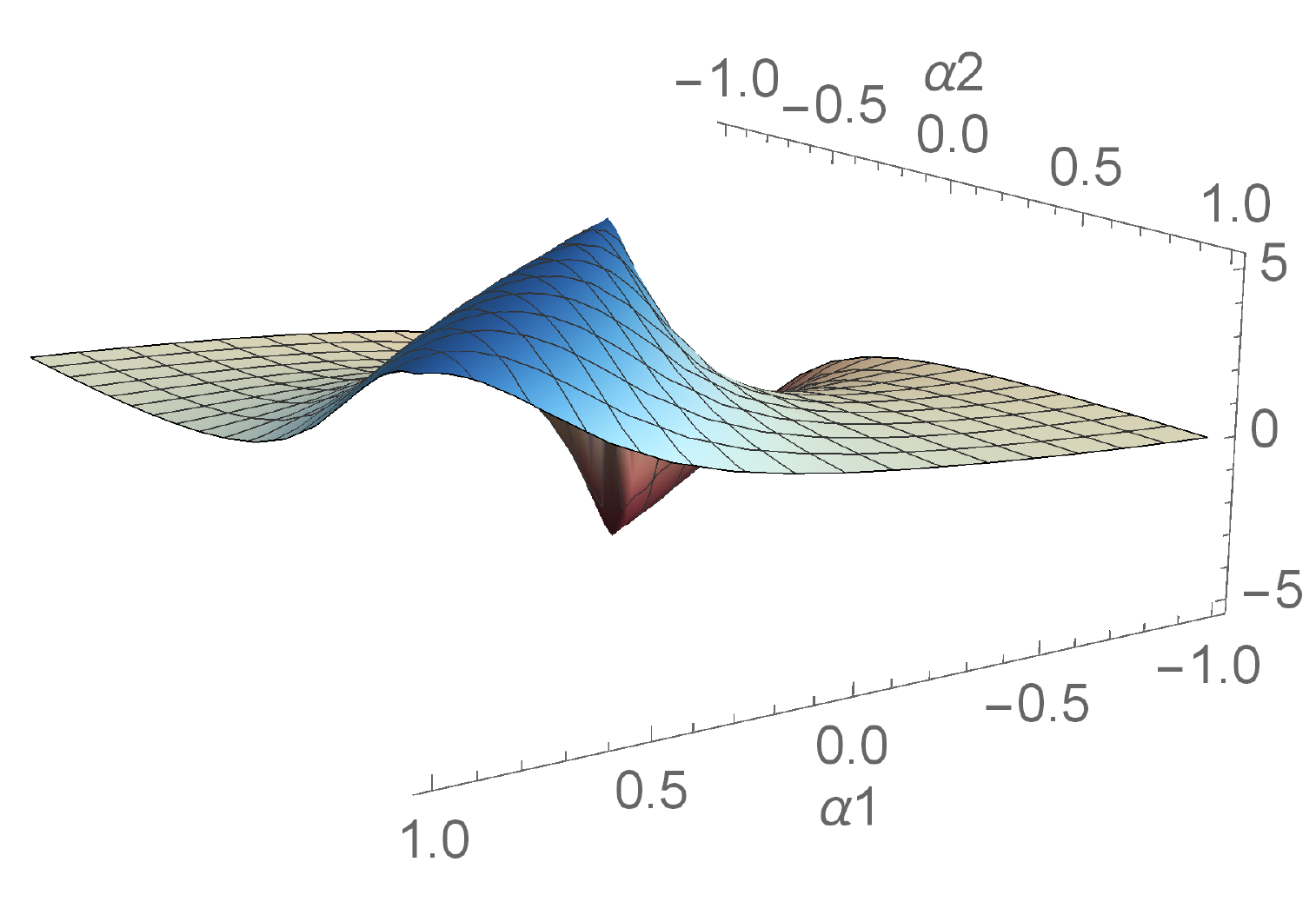}
	\end{center}
	\caption{Top: The exterior derivatives of an $A_\theta$ for $\theta$ between $\frac{\pi}{2}$ and $\frac{2\pi}{3}$ (middle plot corresponding to $x_p$, right to $y_p$). Bottom: Same functions but for $\theta$ between $\pi$ and $\frac{7\pi}{6}$. Note that some of the plots appear relatively unchanged from those of Fig.~\ref{fig:rotationPlatformCurv}; the others acquire deformations starting from the edges.}
	\label{fig:averagePlatformCurv}
\end{figure*}
%%%%%%%%%%%%%%%% end figure %%%%%%%%%%%%%%%%%%% 
%%%%%%%%%%%%%%%%%%%%%%%%%%%%%%%%%%%%%%%%%%%%%%%%%%%%%%%%%%%%%%%%%%%%%%

With a varying $\theta$, robot base trajectories no longer solely lie on the $\alpha_1$-$\alpha_2$ plane in interacting with the inertial platform velocities $\dot x_p$ and $\dot y_p$. However, we are able to simplify this problem and reduce motion planning to analysis of a single exterior derivative function as before if $\theta$ is known to be periodic. In other words, the robot may start out at some arbitrary orientation, but its net motion only moves it forward in the body direction and does not turn it away from its initial orientation. From the connection exterior derivative plot that depicts the mapping between internal joint changes and robot position changes in Fig.~\ref{fig:TL-ED}, we know that such gaits are centered about the origin. 

The following strategy is one which we referred to in Section \ref{sec:roleofext}, that of \emph{decoupling} the dependency of the fiber trajectories on the evolution of the non-symmetric fiber, which in the present case is $g^\text{non} = \theta$. Assuming that the joint trajectories $\alpha_1$ and $\alpha_2$ are also periodic, we can apply \textit{harmonic balance} on the third line of Eq.~\eqref{eq:TL-recon}, a first-order differential equation $\dot \theta = f(b_w, \dot b_w)$, to obtain a solution for $\theta(t)$ in terms of the parameters of $\alpha_1(t)$ and $\alpha_2(t)$ \cite{Hayashi2014}. In other words, if 
\begin{align}
\alpha_1(t) &= B_1 \cos(\omega t), \\
\alpha_2(t) &= B_2 \cos(\omega t - \phi),
\end{align}
then we can find
\begin{align}
\theta(t) &= \Theta \cos(\omega t - \psi) + C,
\end{align}
where $\Theta$, $\psi$, and $C$ are functions of $B_1$, $B_2$, and $\phi$. Taking this procedure one step further, we can apply trigonometric identities to write
\begin{equation}
\theta = a_1 \alpha_1 + a_2 \alpha_2 + C,
\label{eq:theta}
\end{equation}
where $a_1$ and $a_2$ are the solutions of the nonlinear equations
\begin{align*}
\Theta &= \sqrt{(a_1 B_1)^2 + (a_2 B_2)^2 + 2 a_1 a_2 B_1 B_2 \cos \phi}, \\
\psi &= -\text{atan}2(-a_2 B_2 \sin \phi, a_1 B_1 + a_2 B_2 \cos \phi).
\end{align*}
The unknowns can be analytically solved as
\begin{align}
(a_1, a_2) = \pm \left(\frac{\Theta \sin(\phi-\psi)}{B_1 \sin\phi}, \frac{\Theta \sin\psi}{B_2 \sin\phi} \right).
\label{eq:coeff}
\end{align}

With the numerical values of $a_1$ and $a_2$ in hand, we can substitute Eq.~\eqref{eq:theta} into Eq.~\eqref{eq:theta-conn} and obtain a reduced connection mapping solely between the joint variables $b_w$ and the platform fiber variables $x_p$ and $y_p$, without regard to $\theta$. Again, this new connection assumes that $\theta$ is periodic and is only valid for the given or known functional form of $\theta$, whether through analysis or visual tracking. With this connection form, we are able to reduce the previous exterior derivative plots from three dimensions ($\alpha_1$, $\alpha_2$, and $\theta$) back down to two ($\alpha_1$ and $\alpha_2$ only).

Two effective connection exterior derivatives over different ranges of $\theta$ are shown in Fig.~\ref{fig:averagePlatformCurv} (one may be able to characterize such approximations if the robot has limited reorientation over its trajectory, for example). In the first row, $\theta$ has a range from about $\frac{\pi}{2}$ to about $\frac{2\pi}{3}$; in the second, $\theta$ ranges from about $\pi$ to about $\frac{7\pi}{6}$. In both rows, the representative exterior derivative functions corresponding to $x_p$ and $y_p$ are shown. Interestingly, the $x_p$ plot in the first row and $y_p$ plot in the second row are not very different from their constant $\theta$ counterparts at $90$ and $180$ degrees, respectively. This indicates that those particular surfaces are more stable and hold over large ranges of $\theta$. On the other hand, the other two plots are noticeably different, particularly on the edges. In both the $y_p$ plot in the first row and $x_p$ plot in the second, the edges are the first regions to deform as $\theta$ changes.

Given that we are able to find single representative connection derivative plots over a given range of $\theta$, we can use this visual tool for locomotion analysis and planning for the three-link wheeled snake robot on a platform system, whether the inputs are applied to the robot on a passive platform or to an actuated platform underneath a passive robot. A weakness of this approach is that the computed connection only holds for the specific $\theta$ range. If this range changes, whether by shifting or scaling, a new connection must be found.

\section{Chaplygin Beanie on a Platform}
\label{sec:chap}
We now shift our focus from a system for which external actuation strategies can be guided by a principal connection, to one for which no such connection exists, requiring that we develop additional tools or intuition for motion planning. We first investigate behaviors for a completely passive system with the robot being driven by elastic elements in its body, provide a formal proof for these behaviors, and then illustrate the degree to which this formal understanding can inform investigations into an externally actuated version of the system. We then compute motion primitives for multiple passively compliant Chaplygin beanies on an actuated platform. The material in this section appeared previously in \cite{Buchanan_2020} without the broader context of the present paper.
%%%%%%%%%%%%%%%%%%%%%%%%%%%%%%%%%%%%%%%%%%%%%%%%%%%%%%%%%%%%%%%%%%%%%%
%%%%%%%%%%%%%%%% begin figure %%%%%%%%%%%%%%%%%%%
\begin{figure}[t]
	\begin{center}
		\includegraphics[width=.28\textwidth,trim={0cm 0cm 0cm 0cm},clip]{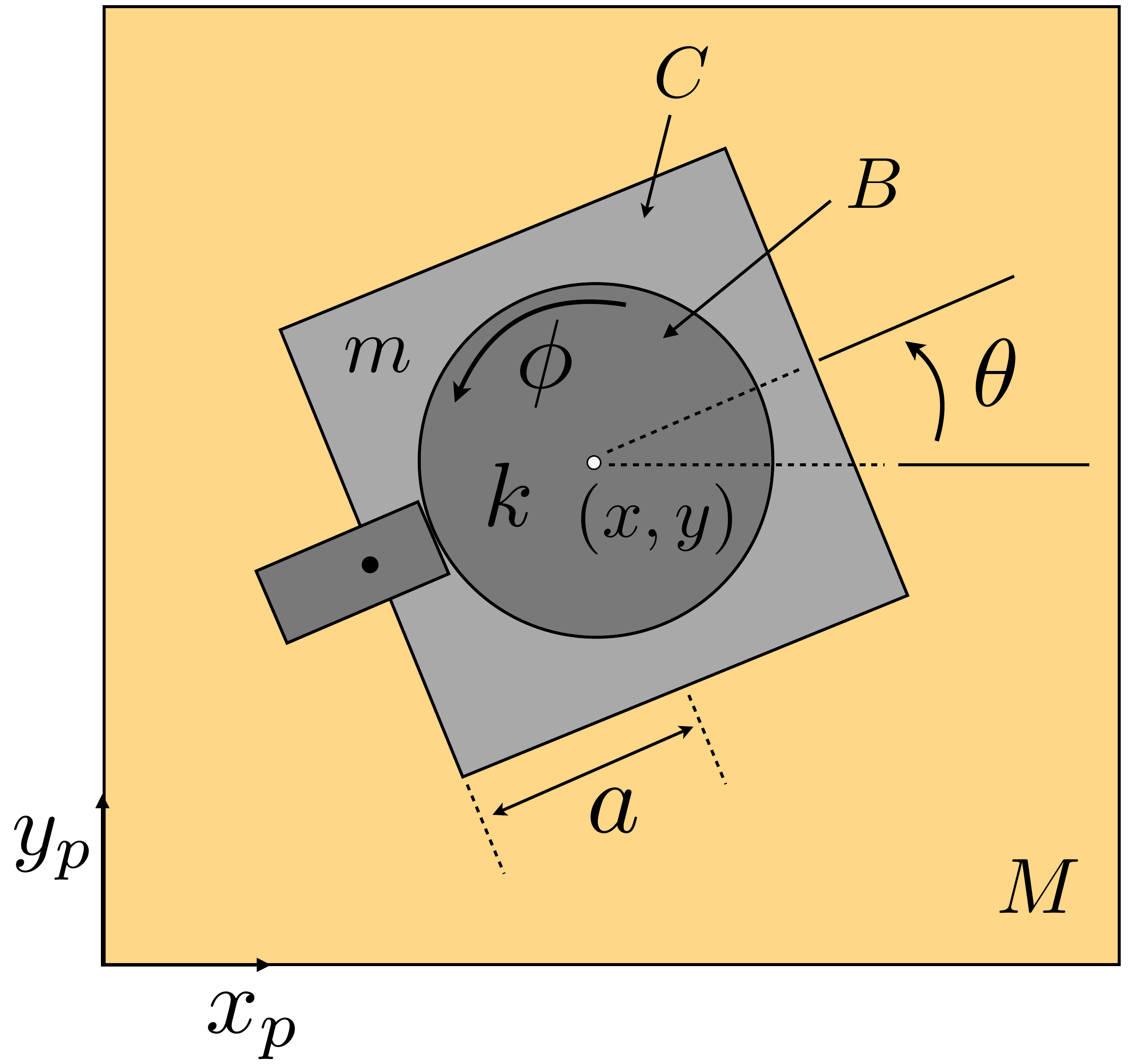}
	\end{center}
	\caption{Coordinate assignments and inertial parameters of a Chaplygin beanie on a platform. We designate $m$, $B$, $C$ the mass of the Chaplygin beanie, rotational inertia of the rotor, and rotational inertia of the cart, respectively. We let $k$ be the stiffness of the linear torsional spring coupling the cart to the rotor, and $M$ the total mass of the platform.}
	\label{fig:beanieparams}
\end{figure}
%%%%%%%%%%%%%%%% end figure %%%%%%%%%%%%%%%%%%%
%%%%%%%%%%%%%%%%%%%%%%%%%%%%%%%%%%%%%%%%%%%%%%%%%%%%%%%%%%%%%%%%%%%%%%
\subsection{Robot Description} 
Constrained to the platform via a wheel located at its rear, the Chaplygin beanie locomotes using a rotor sitting atop  its body. The rotor can be driven actively, for example by a motor, or passively, via an elastic element coupling the cart to the rotor. In this subsection, we study the latter of these two configurations, and assume the elastic element to be a linear torsional spring. The total mass of the vehicle is represented by $m$, its rotational inertia about the center of mass as $C$, rotational inertia of the rotor about the center of mass as $B$, and the mass of the platform as $M$. The distance between the center of mass and the contact point at the wheel is denoted by $a$, and the stiffness of the spring coupling the rotor to the body denoted by $k$. The position of the vehicle relative to the platform is given coordinates $(x, y)$, its orientation $\theta$, the rotor angle relative to the vehicle heading by $\phi$, and the position of the platform in a laboratory frame by $(x_p, y_p)$.

The configuration manifold for this system is $Q_c = \text{SE}(2) \times \mathbb{S}^1 \times \mathbb{R}^2$, with a subscripted $()_c$ to denote \textit{Chaplygin beanie}. Adhering to our earlier notation that delineates between internal and external configurations, we let $G^\text{int}_c = \text{SE}(2)$, $B_c = \mathbb{S}^1$, and $G^\text{ext}_c = \mathbb{R}^2$. Furthermore, we let $g^\text{int}_c = (x,y,\theta) \in G^\text{int}_c$, $b_c = \phi \in B_c$, and $g^\text{ext}_c = (x_p,y_p) \in G^\text{ext}_c$. Note that the use of this notation does not assume the existence of a connection relating trajectories in $B_c$ to trajectories in $G^\text{int}_c$ or $G^\text{ext}_c$. Like the three-link snake robot, the Chaplygin beanie is constrained via a no-slip condition on its wheel. The nonholonomic constraint at the wheel is given by
\begin{equation}
    -\dot{x}\sin\theta + \dot{y}\cos\theta - a\dot{\theta} = 0.
\end{equation}
Constraints of this kind can also be thought of as one forms lying in the codistribution on the configuration manifold $Q_c$, written equivalently as
\begin{equation}
    \omega = -\sin\theta dx + \cos\theta dy - a d\theta.
\end{equation}

\begin{figure}[t]
	\begin{center}
		\includegraphics[width=.48\textwidth]{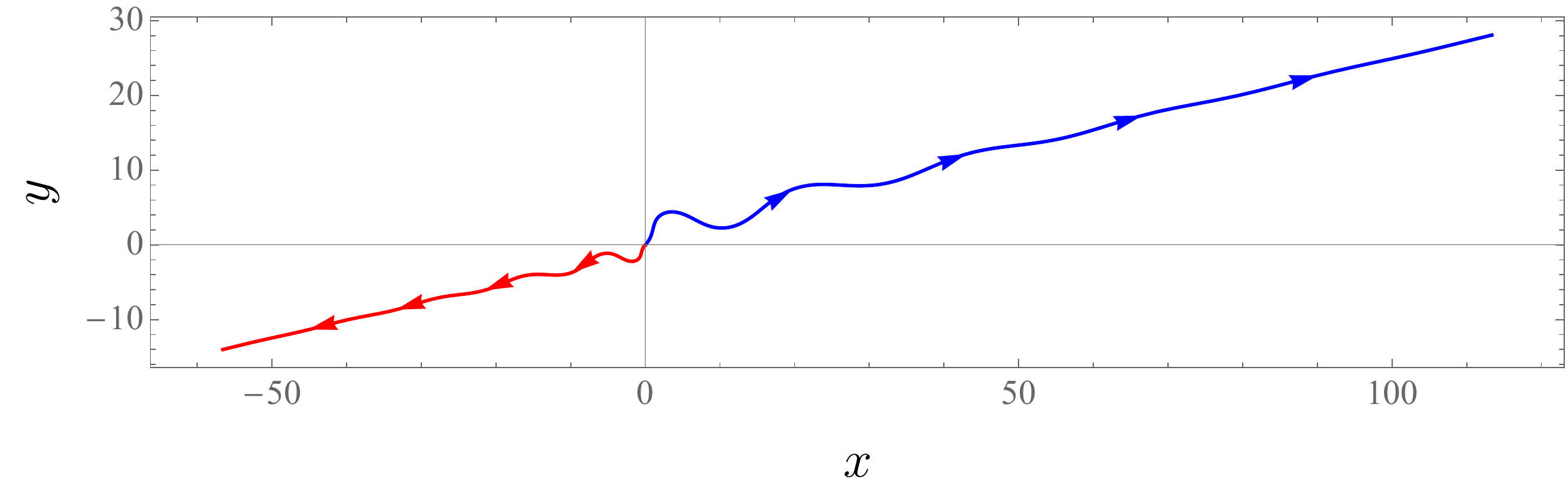}
	\end{center}
	\caption{Trajectory of a Chaplygin beanie (in blue) as it locomotes atop a platform. The platform trajectory is shown in red.}
	\label{fig:beanieplattraj} 
\end{figure}

Given an initial condition $g^\text{int}_c(0) = (0,0,-\pi/4)$, $b_c = \pi$, $g^\text{ext}_c(0) = (0,0)$, and setting the inertial parameters and $k$ equal to unity, we observe the trajectory in Fig \ref{fig:beanieplattraj}.

% It is generally true that initial conditions corresponding to the system starting at rest with arbitrary nonzero $b_c(0) = \phi_0$ will result in such stable trajectories \cite{Buchanan_2020}.

\subsection{Nonholonomic Reduction} 
Stable trajectories like the one in Fig. \ref{fig:beanieplattraj} illustrate fixed points in a reduced space. To analyze these fixed points, we employ \textit{nonholonomic reduction}, and leverage the geometric structure of the \textit{passive} system to guide our development of a formal understanding of the types of stable locomotive trajectories that can arise from external actuation.

The evolution equations arising from the reduction are the changes in linear and angular momentum permitted by the no-slip constraint at the wheel and will replace the equations describing the evolution of $\dot{x}$, $\dot{y}$, and $\dot{\theta}$. The presence of a platform gives rise to two additional evolution equations, one of which is the time evolution of forward translational momentum of both the Chaplygin beanie and the platform, the second of which is the time evolution of the momentum of both the Chaplygin beanie and the platform in the direction orthogonal to that allowed by the no-slip consraint at the wheel. We also refer to this momentum term as momentum \textit{lateral} to the forward motion of the vehicle. The Lagrangian for the system is given by
\begin{equation}
    \begin{aligned}
    L_c & = \frac{1}{2}m((\dot{x} + \dot{x}_p)^2 + (\dot{y} + \dot{y}_p)^2) + \frac{1}{2}C \dot{\theta}^2 \\ & + \frac{1}{2} B(\dot{\theta} + \dot{\phi})^2
    + \frac{1}{2}M(\dot{x}_p^2 + \dot{y}_p^2) - \frac{1}{2}k \phi^2.
    \end{aligned}
    \label{eqn:beanielagrangian}
\end{equation}

We require that the one form describing the no-slip constraint be annihilated by the system's generalized velocity, having coordinates $(x,y,\theta,\phi,x_p,y_p)$, at every point $q_c \in Q_c$. For the Chaplygin beanie on a platform with finite inertia, the manifold on which the dynamics evolve is described by the configuration manifold $Q_c = \text{SE}(2) \times \mathbb{S}^1 \times \mathbb{R}^2$.

% Note that both $\text{SE}(2)$ and $\mathbb{R}^2$ together with matrix multiplication as the group operation are both Lie groups, and that their Cartesian product will also yield a Lie group with matrix multiplication as the group operation. 
We let $G = \text{SE}(2) \times \mathbb{R}^2 = G^\text{int}_c \times G^\text{ext}_c$ be the Lie group representing the group of rigid translations and rotations of the Chaplygin beanie and rigid translations of the platform. Let an arbitrary element of the Lie group $G$ be described by $g = (x,y,\theta,x_p,y_p)$. Consider now a different element of $G$ with assignment $g = (a,b,\alpha,c,d)$ and an element $q_c \in Q_c$ given by $q_c = (x,y,\theta,x_p,y_p,\phi)$. The left action of $G$ on $Q$ is given by
\begin{equation}
    \begin{aligned}
        \Phi: & G \times Q_c \mapsto Q_c:(g,q_c) \mapsto (a + x\cos\alpha - y\sin\alpha, \\ & b+y\cos\alpha+x\sin\alpha,  c+x_p\cos\alpha-y_p\sin\alpha, \\ & d+y_p\cos\alpha + x_p\sin\alpha,  \theta+\alpha, \phi).
    \end{aligned}
\end{equation}
The above defines a tangent lifted action, $T\Phi$, of $G$ on the tangent bundle $TQ_c$, given by
\begin{equation}
    \begin{aligned}
        T\Phi & : G \times TQ_c \mapsto TQ_c \\ & : \big ( g, (q_c,\dot{q}_c) \big ) \mapsto (\Phi(g,q_c),T_q\Phi(g,q_c)\dot{q}_c)
    \end{aligned}
    \label{eqn:tangentliftedbeanie}
\end{equation}
\noindent where $T_{q_c}\Phi(g,q_c)$ is given by
\begin{equation}
\begingroup
\setlength\arraycolsep{2pt}
    T_{q_c}\Phi(g,q_c) \! = \! \begin{bmatrix} \cos\alpha & -\sin\alpha & 0 & 0 & 0 & 0 \\ \sin\alpha & \cos\alpha & 0 & 0 & 0 & 0 \\ 0 & 0 & 1 & 0 & 0 & 0 \\ 0 & 0 & 0 & \cos\alpha & -\sin\alpha & 0 \\ 0 & 0 & 0 & \sin\alpha & \cos\alpha & 0 \\ 0 & 0 & 0 & 0 & 0 & 1\end{bmatrix}.
    \endgroup
\end{equation}

We can use the mapping of tangent vectors according to the above Jacobian to show that the Lagrangian, Eq. \ref{eqn:beanielagrangian}, is invariant under the tangent lifted action, Eq. \ref{eqn:tangentliftedbeanie}. We must now verify the constraint one form, $\omega$, is invariant under the group action, $\Phi$. Given a point $q_c \in Q_c$ and a tangent vector $\dot{q}_c \in T_{q_c}Q_c$, we check for left-invariance by computing
\begin{equation}
    \langle \omega_{q_c}, \dot{q}_c\rangle = \langle\omega_{\Phi_g q_c},T_{q_c}\Phi_g \dot{q}_c\rangle,
    \label{eqn:oneforminvariance}
\end{equation}

\noindent where $\omega_{q_c}$ is the constraint one form evaluated at point $q_c$ and $\omega_{\Phi_g q_c}$ is the constraint one form evaluated at point $q_c$ after being mapped through the group action. It can be shown that the Eq. \eqref{eqn:oneforminvariance} holds. Thus, the system's Lagrangian is invariant under the tangent lifted action and the constraint one form is invariant under the cotangent lifted action, making $G$ a symmetry group. The Lie group $G$ then acts on the $G$ part of $Q_c$ via left translation, leaving the $B_c = \mathbb{S}^1$ part unchanged.

% The natural pairing above is computed as
% \begin{equation}
%     \begin{bmatrix} -\sin\theta & \cos\theta & -a & 0 & 0 & 0 \end{bmatrix} \begin{bmatrix} \dot{x} \\ \dot{y} \\ \dot{\theta} \\ \dot{x}_p \\ \dot{y}_p \\ \dot{\phi} \end{bmatrix} = \begin{bmatrix} -\sin(\theta+\alpha) & \cos(\theta+\alpha) & -a & 0 & 0 & 0 \end{bmatrix} T_q\Phi_g \begin{bmatrix} \dot{x} \\ \dot{y} \\ \dot{\theta} \\ \dot{x}_p \\ \dot{y}_p \\ \dot{\phi}. \end{bmatrix}
% \end{equation}

We take an approach presented in \cite{bloch1996nonholonomic}, involving the choosing of appropriate left-invariant vector fields spanning the intersection of the constraint distribution and the space tangent to the orbit of the group action, and leverage \cite{Kelly2012} in computing the components of the nonholonomic momentum. We designate the distribution $\mathcal{D}_{q_c}$ as the space of all tangent vectors which annihilate the constraint one form, $\omega$, and is given by
\begin{equation}
    \begin{aligned}
        \mathcal{D}_{q_c} & = \text{span}\{-a \sin\theta \frac{\partial}{\partial x} + a \cos\theta \frac{\partial}{\partial y} + \frac{\partial}{\partial \theta}, \\ & \cos\theta \frac{\partial}{\partial x} + \sin\theta\frac{\partial}{\partial y},\frac{\partial}{\partial \phi} \}.
    \end{aligned}
\end{equation}

Furthermore, we designate $T_{q_c}\text{Orb}(q_c)$ as the space tangent to the orbit of the group action, given by
\begin{equation}
    T_{q_c}\text{Orb}(q_c) = \text{span}\{ \frac{\partial}{\partial x}, \frac{\partial}{\partial y}, \frac{\partial}{\partial \theta}, \frac{\partial}{\partial x_p}, \frac{\partial}{\partial y_p}\}.
\end{equation} 
\noindent We then choose appropriate vector fields on the configuration space, $Q_c$, to span
\begin{equation}
    S_{q_c} = \mathcal{D}_{q_c} \cap T_{q_c}\text{Orb}(q_c), \quad \forall q_c \in Q_c.
    \label{eqn:span}
\end{equation}
\noindent The intersection of $\mathcal{D}_{q_c}$ and $T_{q_c}\text{Orb}(q_c)$ constitutes the space in which a reduced representation of the dynamics evolve. Its dimension corresponds to the number of evolution equations obtained from the reduction. The following choice of vector fields is made to define this intersection.
\begin{equation}
    \begin{aligned}
    S_{q_c} & = \text{span}\{-a \sin \theta \frac{\partial}{\partial x} + a\cos\theta\frac{\partial}{\partial y} + \frac{\partial}{\partial \theta}, \\
    &
    \cos\theta \frac{\partial}{\partial x} + \sin\theta\frac{\partial}{\partial y},  \cos\theta \frac{\partial}{\partial x_p} + \sin\theta \frac{\partial}{\partial y_p}, \\
    &
    -\sin\theta \frac{\partial}{\partial x_p} + \cos\theta\frac{\partial}{\partial y_p} \}.
    \end{aligned}
\end{equation}
The first two vector fields correspond to rotation about the contact point at the wheel and longitudinal translation of the vehicle, respectively \cite{Kelly2012}. Flow along the third corresponds to forward translation of the entire system, including the platform, along the forward direction of the Chaplygin beanie. The fourth of these vector fields represents motions of the entire system lateral to the forward direction of the Chaplygin beanie. 

We invoke the Einstein summation convention in the following definition of the momentum map and designate $q_c^i$ as the $i$th coordinate on the configuration manifold, $Q_c$. The nonholonomic momenta are computed following 
\begin{equation}
    J^{nhc} = \frac{\partial L_c}{\partial \dot{q}_c^i} (\xi_{Q_c})^i.
    \label{eqn:nonholomom}
\end{equation}

The resulting momenta are given directly by Eq. \eqref{eqn:nonholomom2}. It is clear by inspection that $J_{LT}$ and $J_{RW}$ correspond to forward translational momentum and angular momentum about the contact point of the wheel, respectively. The quantities $J_X$ and $J_Y$ correspond to forward translational momentum and momentum lateral to the direction allowed by the nonholonomic constraint for both the Chaplygin beanie and the platform.
\twocolumn[
\begin{@twocolumnfalse}
\begin{equation}
    \begin{gathered}
        \dot{J}_{LT} = -\frac{m((B+C)J_Y + Ma(J_{RW} - B\alpha))(maJ_Y - (m+M)(J_{RW} - B\alpha))}{(M(ma^2 + B +C) + m(B+C))^2}, \\
        \dot{J}_{RW} = \frac{aJ_{LT}(maJ_Y - (m+M)(J_{RW}-B\alpha))}{M(ma^2 + B +C) + m(B+C)},  \\
        \dot{J}_X = -\frac{J_Y(maJ_Y +(m+M)(J_{RW}-B\alpha))}{M(ma^2 + B +C) + m(B+C)}, \\
        \dot{J}_Y = -\frac{J_X(-maJ_Y+(m+M)(J_{RW}-B\alpha))}{M(ma^2 + B +C) + m(B+C)} 
    \end{gathered} \label{eqn:evoeqns}
\end{equation}
\end{@twocolumnfalse}]
\begin{equation}
    \begin{split}
      J_{LT} & = m(\dot{x} + \dot{x}_p)\cos\theta + m(\dot{y} + \dot{y}_p)\sin\theta, \\
        J_{RW} & = -ma(\dot{x} + \dot{x}_p)\sin\theta + ma(\dot{y} + \dot{y}_p)\cos\theta \\
        & + (B+C)\dot{\theta} + B\dot\phi,  \\
        J_{X}  = & m (\dot{x}+\dot{x}_p)\cos\theta + m(\dot{y}+\dot{y}_p)\sin\theta + \\
        & M\dot{x}_p\cos\theta + M\dot{y}_p\sin\theta,\\
      J_{Y}  = & -m (\dot{x}+\dot{x}_p)\sin\theta + m(\dot{y}+\dot{y}_p)\cos\theta - \\
        & M\dot{x}_p\sin\theta + M\dot{y}_p\cos\theta.
    \end{split} \label{eqn:nonholomom2}
\end{equation}

\noindent Similarly, the evolution equations are computed following 
\begin{equation}
    \dot{J}^{nhc} = \frac{\partial L_c}{\partial \dot{q}_c^i} \bigg[\frac{d\xi_c}{dt}\bigg]_{Q_c}^i,
    \label{eqn:nonholonomicevo}
\end{equation}
and are given by Eq. \eqref{eqn:evoeqns}. 

\subsection{Stable Trajectories of Passive System}
A formal understanding of the types of stable locomotive trajectories of a completely passive system for which $b_c(0) = \phi_0$ is nonzero can give us insight into how to actuate the platform to induce locomotion in a passive Chaplygin beanie. In this section, we provide proof for the following proposition.

\begin{prop}
For all initial conditions corresponding to the system starting at rest and for which $b_c(0) = \phi_0 \neq 0$, $J_{LT}$ will asymptotically approach a positive constant, with $J_{RW}$, $\phi$, and $\dot \phi$ exponentially decreasing to zero as $t\rightarrow \infty$.
\end{prop}

This proposition effectively says that all of the rotational momentum in the system will be transformed into forward translational momentum. We now present a formal argument for Proposition 1.

\begin{figure}[t]
    \centering
    \includegraphics[width=0.45\textwidth,trim={0 0cm 0 0cm},clip]{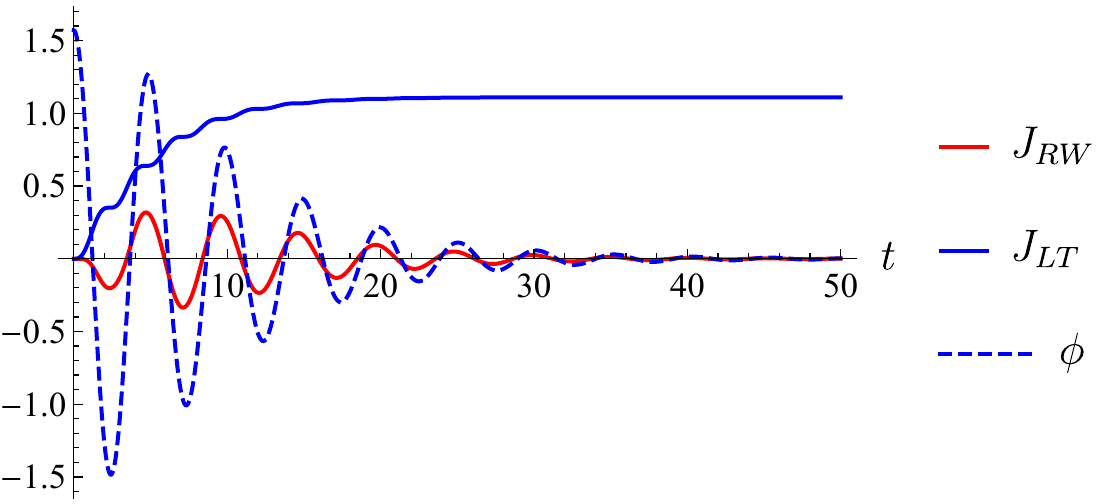}
    \caption{Rotational momentum about the rear wheel, longitudinal translational momentum, and the rotor angle.}
    \label{fig:momentumplot}
\end{figure}

\begin{proof} Defining the following variables, the nonholonomic momenta, $\phi$, and $\dot{\phi}$ can be expressed as 
\begin{gather}
    % \label{eqn:evoeqns}
    r = \frac{J_{LT}}{d}, \quad w = \frac{J_{RW} - B\alpha}{d}, \quad p_x = \frac{J_X}{d},\label{eqn:neatermom} \\
    p_y = \frac{J_Y}{d}, \quad \alpha = \dot{\phi}\nonumber.
    \label{eqn:evoeqnsmod}
\end{gather}
The constants $d = m(B+C)+M(ma^2 + B + C)$, $\gamma_1 = -m^2a(B+C)/d$, $\gamma_2 = (m(m+M)(B+C)-m^2Ma^2)/d$, $\gamma_3 = mMa(m+M)/d$, $\lambda_1 = ma^2$, $\lambda_2 = -a(m+M)$, $\mu_1 = -ma$, $\mu_2 = m+M$, $\nu_0 = B/d$, $D = B(mMa^2 + C(m+M))$, $\nu_1 = -dk/D$, $\nu_2 = -Bma^2(m+M)/(dD)$, $\nu_3 = aB(m+M)^2/(dD)$, $\nu_4 = Bm^2a^2/(dD)$, and $\nu_5 = -mBa(m+M)/(dD)$ fully encapsulate the presence of system parameters in a more convenient form for analysis. The reduced dynamics are then easier to analyze for stability. Taking the time derivatives of \eqref{eqn:neatermom} and using \eqref{eqn:evoeqns} to make the necessary substitutions, the evolution equations become
\begin{equation}
    \begin{aligned}
        \dot{r}  &= \gamma_1 p_y^2 + \gamma_2p_yw + \gamma_3 w^2, \\
        \dot{w}  &=  \lambda_1 r p_y + \lambda_2 r w  - \nu_0 (\nu_1 \phi + \nu_2 r p_y \\ &+ \nu_3 r w + \nu_4 p_x p_y + \nu_5 p_x w),\\
        \dot{p}_x &= \mu_1p_y^2 + \mu_2p_yw, \label{eqn:neaterdyn}\\
        \dot{p}_y &= -\mu_1p_xp_y - \mu_2p_xw, \\
        \alpha &= \dot{\phi},\\
        \dot{\alpha} &= \nu_1 \phi + \nu_2 r p_y + \nu_3 r w + \nu_4 p_x p_y + \nu_5 p_x w.
    \end{aligned}
\end{equation}
The dynamics given by Eq. \eqref{eqn:neaterdyn} can be further simplified under assumptions of momentum conservation. The quantity $p_x^2 + p_y^2$ is conserved, with all its level sets invariant under Eq. \eqref{eqn:neaterdyn}. We wish to prove that all trajectories corresponding to $p_x^2 + p_y^2 = 0$ approach the $r$ axis asymptotically. Restricting the dynamics to this level set, Eq. \eqref{eqn:neaterdyn} can be written as
\begin{equation}
    \begin{gathered}
        \dot{r} = \gamma_3 w^2, \quad \dot{w} = \lambda_2 r w  - \nu_0 (\nu_1 \phi + \nu_3 r w),\\
        \dot{p}_x = 0, \quad \dot{p}_y = 0, \quad \alpha = \dot{\phi},\label{eqn:neaterdyn2}\\
        \dot{\alpha} = \nu_1 \phi + \nu_3 r w.
    \end{gathered}
\end{equation}

The time evolution of $r$, $w$, $\phi$, and $\alpha$ then fully describe the behavior of the system. By inspection it is clear that $\dot{r}$ is nonnegative, and is positive where $w$ is nonzero. For $w=0$, $\dot{w}$ is nonzero for $\phi \neq 0$. It follows that $r$ will increase for all time given that $w \neq 0$ and $\phi \neq 0$. Thus, $r$ will increase for all time unless $w$, $\phi$, and $\alpha$, are zero for all time, requiring $r$ to increase unless the flow of the vector field corresponding to Eq. \eqref{eqn:neaterdyn2} is always on the $r$ axis. All fixed positive values of $r$ correspond to a linear dynamical system described by $\dot{w}$, $\alpha$, and $\dot{\alpha}$. For every such $r$, denoted by $r_c$, the dynamics are then
\begin{equation}
    \begin{bmatrix} \dot{w} \\ \alpha \\ \dot{\alpha} \end{bmatrix} = \begin{bmatrix} \lambda_2 r_c w  - \nu_0 (\nu_1 \phi + \nu_3 r_c w) \\ \dot{\phi} \\ \nu_1 \phi + \nu_3 r_c w \end{bmatrix}.
    \label{eqn:awayfromr}
\end{equation}
The Jacobian of Eq. \eqref{eqn:awayfromr} is
\begin{equation}
    \Lambda = \begin{bmatrix} \lambda_2 \nu_0 \nu_3 r_c & -\nu_0 \nu_1 & 0 \\
    0 & 0 & 1 \\
    \nu_3 r_c & \nu_1 & 0 \end{bmatrix}.
    \label{eqn:jacobianawayfromr}
\end{equation}
The eigenvalues of Eq. \eqref{eqn:jacobianawayfromr} at $(w,\phi,\alpha) = (0,0,0)$ correspond to the roots of a third order polynomial in $p$ with parameter-dependent coefficients, written as
\begin{equation}
    p^3 + (\nu_0\nu_3 r_c - \lambda_2 r_c)p^2 - \nu_1p + \lambda_2\nu_1r_c = 0.
\end{equation}

With $r_c > 0$, Decartes' rule of signs tells us that the polynomial above has roots with all negative real part, showing that $w$, $\phi$, and $\alpha$ exponentially decrease to zero as $r$ increases. Given knowledge of the asymptotic values of $w$, $\phi$, and $\alpha$, this result suggests a stable fixed point of Eq. \eqref{eqn:neaterdyn2} at $(r,w,\phi,\alpha) = (r_\infty,0,0,0)$.
\end{proof}
Our proof is supported by numerical simulations, an example of which is shown in Fig. \ref{fig:momentumplot}. Note that this aligns with the behavior exhibited by the Chaplygin beanie in Fig. \ref{fig:beanieplattraj}.

\subsection{Platform Actuation in Body Frame} 
We have established that for every $b_c(0) = \phi_0 \neq 0$ corresponding to the zero level set of momentum, all of the rotational momentum decreases exponentially to zero as time approaches infinity. For the Chaplygin beanie equipped with a torsional spring, we see from numerical simulations that the rotational dynamics are dampened and oscillatory given such initial conditions. This perceived damping is a result of the conversion of rotational momentum into translational momentum. Using our knowledge of the steady-state behaviors of the completely passive system, we can therefore assume that there might exist periodic control inputs to the platform that will induce limit cycle behavior in the dynamics of the Chaplygin beanie. That is, there might exist platform control inputs that result in steady periodic trajectories in $J_{LT}$, $J_{RW}$, $\phi$, and $\dot \phi$. Limit cycles for an internally-actuated Chaplygin beanie were shown to exist in \cite{pollard2019swimming}.

\begin{figure}[t]
    \centering
    \includegraphics[width=0.3\textwidth,trim={0 0cm 0 0cm},clip]{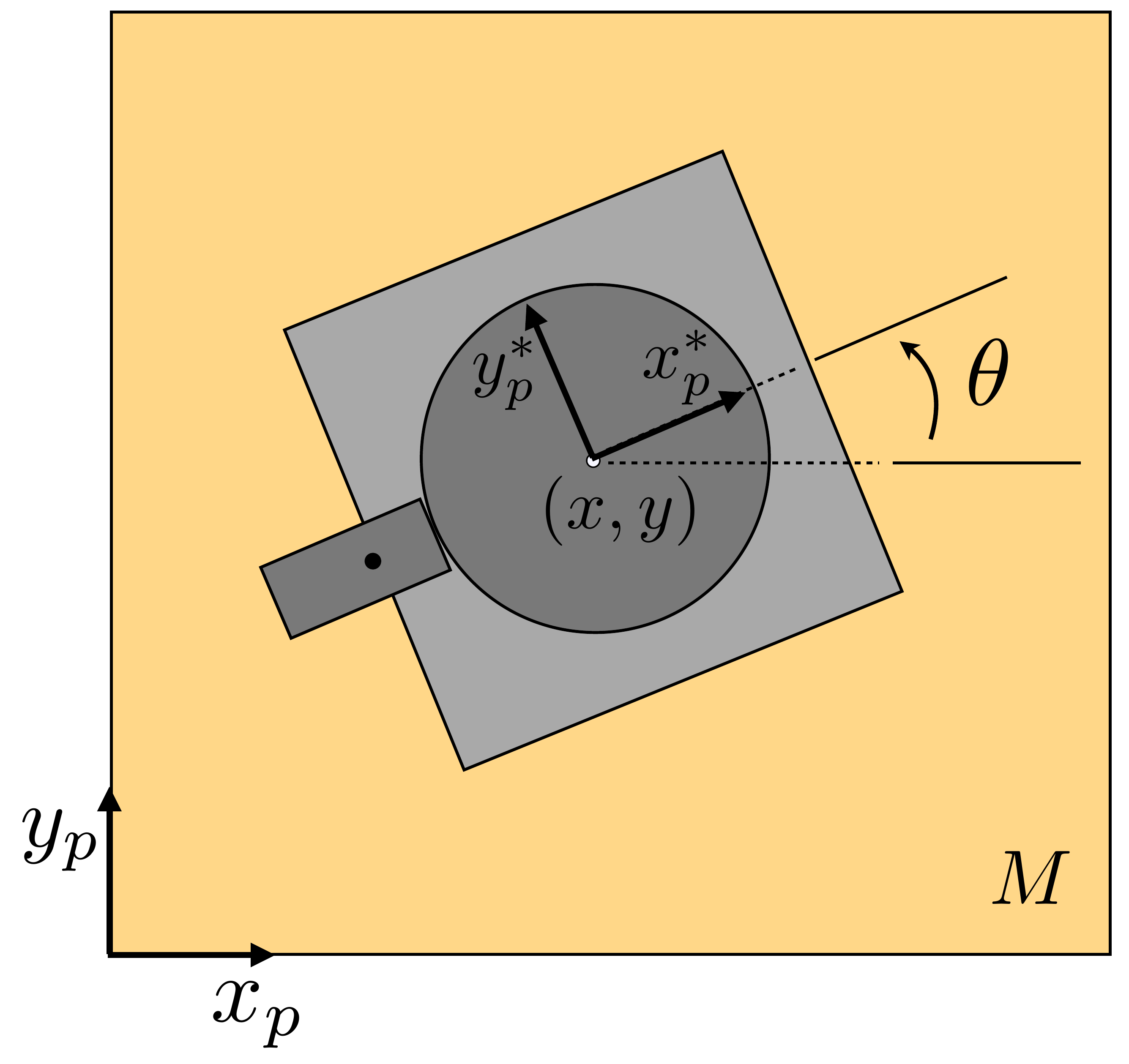}
    \caption{Platform actuation rotated so as to exert control in the direction orthogonal to direction of motion allowed by the no-slip constraint at the wheel}
    \label{fig:beaniePlatformActuation}
\end{figure}

\subsubsection{Frequency Response Analysis}
We seek to characterize locomotive behaviors when the frequency at which the environment is stimulated varies over a range of values containing the natural frequency of the rotor and the modal frequency of the body-rotor couple when not constrained to the platform, both of which are dependent on the stiffness of the spring. These two frequencies provide a natural starting point for our analysis. Consider a single passively compliant Chaplygin beanie atop an actuated platform and let its parameters, $m$, $B$, $C$, $a$, and $k$, be equal to unity. Note that the platform can only be actuated in the directions $(x_p,y_p)$ as shown in Fig. \ref{fig:beaniePlatformActuation}. There's no reason to assume a relationship exists between forward translational speed, heading, or even stability, when actuating purely in the $(x_p,y_p)$ directions. In fact, such a relationship is obfuscated by dependence on the initial heading of the robot. However, such a relationship could exist when considering actuation in a rotated frame of reference, orthogonal to the allowable direction of motion required by the nonholonomic constraint at the robot's wheel, much like the actuation strategy considered for the three-link wheeled snake robot in the preceding section. 

Fig. \ref{fig:beaniePlatformActuation} shows the rotated reference frame of the platform. Actuation along $y_p^*$ is not only independent of the heading of the vehicle, but also the direction for which its passive dynamics are most responsive. Actuation along $x_p^*$, for example, causes no deformations of the rotor relative to the body and therefore no passive response. Consider the natural frequency of the rotor and the modal frequency of the vehicle when not constrained to the platform, given by
\begin{equation}
    \begin{gathered}
        \omega_{nat} = \sqrt{\frac{k}{B}}, \qquad \omega_{mod}  = \sqrt{\frac{k(B+C)}{BC}}. \label{eqn:frequencies}\\
    \end{gathered} 
\end{equation}
With a spring coupling the cart to the rotor, we sweep through a range of frequencies for a particular set of parameters to analyze the response of the system to external actuation in the $y_p^*$ direction and use asymptotic mean forward translational momentum of the system ($J_{LT}$) as a performance metric. Since we expect that under certain periodic actuation the system dynamics will approach a limit cycle behavior, we will compute the mean forward translational momentum for an integer number of periods of oscillation. Forward translational momentum of the Chaplygin beanie is given by $J_{LT}$ in Eq. \ref{eqn:nonholomom}.

\begin{figure}[t]
    \centering
    \includegraphics[width=0.5\textwidth,trim={0 0cm 0 0cm},clip]{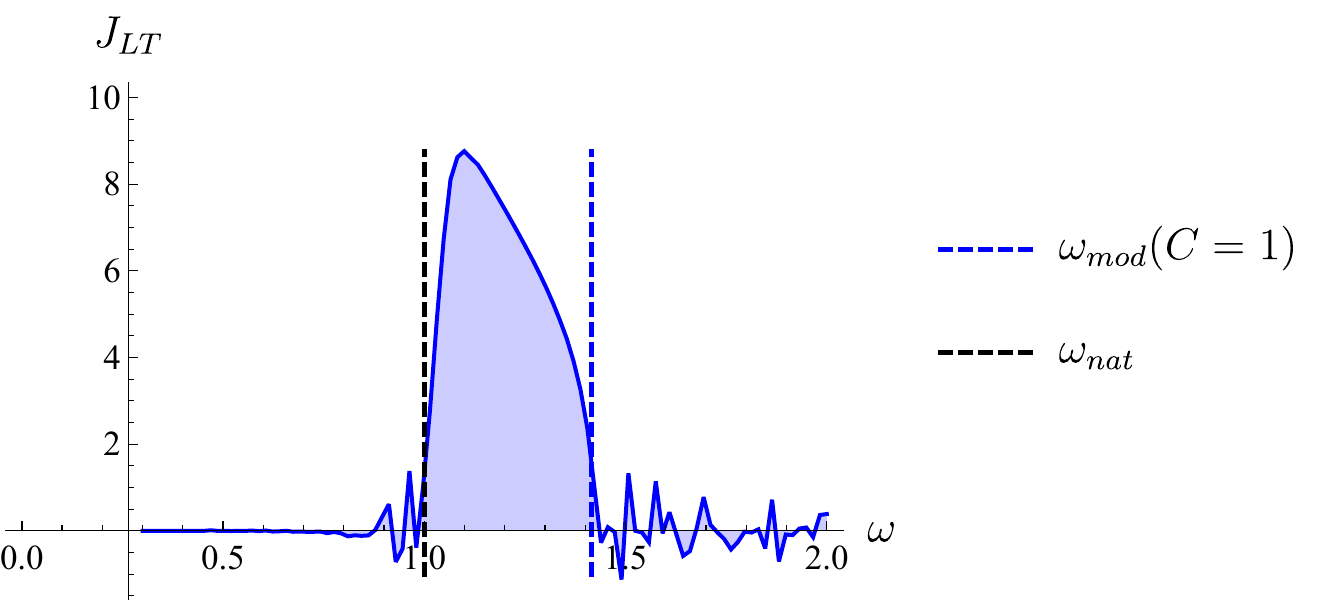}
    \caption{Frequency response of a passive Chaplygin beanie under external actuation in the body frame. The parameters were set to unity to obtain this response.}
    \label{fig:frequencyresponse}
\end{figure}

Consider a situation that allows the orientation of the robot to be tracked in the environment. We actuate the platform according to
\begin{equation}
    \begin{aligned}
       \begin{bmatrix} x_p^* \\ y_p^* \end{bmatrix} = \begin{bmatrix}\cos\theta & -\sin\theta \\ \sin\theta & \cos\theta \end{bmatrix}^{-1} \begin{bmatrix} x_p \\ y_p \end{bmatrix}
       \label{eqn:controlrotated}
    \end{aligned}
\end{equation}
\noindent and let $x_p^* = 0$ and $y_p^* = A\sin(\omega t)$. Note that this is effectively a feedback-like controller in that it requires tracking the heading of the robot, which is then used to compute the control $x_p$ and $y_p$. Setting system parameters and the amplitude $A$ to unity, we discretize the range of frequencies between 0.3 and 2.0 into $N=100$ equally-spaced intervals. Using the final three periods of oscillation to compute $J_{LT}$ for a given actuation frequency $\omega$, we obtain the frequency response plot shown in Fig.  \ref{fig:frequencyresponse}.

\begin{figure}[t]
    \centering
    \includegraphics[width=0.5\textwidth,trim={0 0cm 0 0cm},clip]{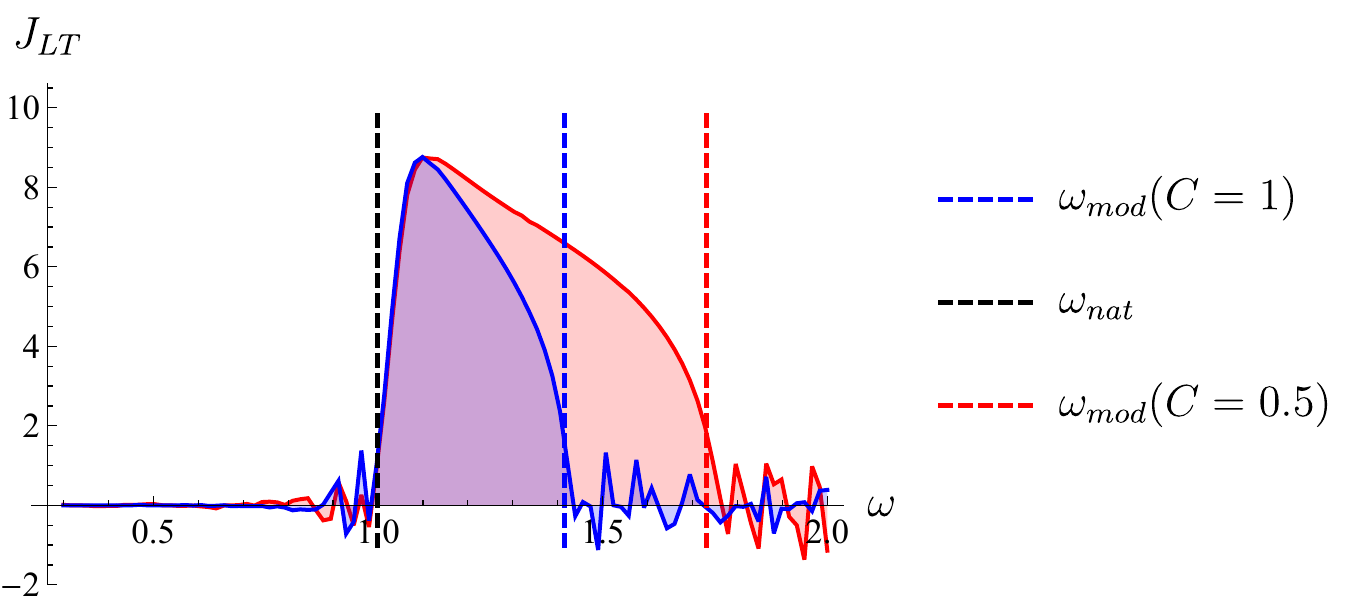}
    \caption{Frequency responses of a Chaplygin beanie for two different parameter combinations}
    \label{fig:frequencyresponse2}
\end{figure}

In carrying out this experiment, it is clear that the natural frequency of the rotor and modal frequency of the system in free space bound a region of high performance when considering mean forward translational momentum as a metric. Additional questions concerning generalizability and associated behaviors arise from this particular result. To address the first of these questions, we consider a variation in the parameters of the robot, compute the corresponding frequencies given in Eq. \eqref{eqn:frequencies} and generate similar results to Fig. \ref{fig:frequencyresponse}. The results of these experiments are shown in Fig. \ref{fig:frequencyresponse2}. The natural frequency of the rotor and modal frequency of the robot in free space again yield lower and upper bounds on regions of high performance.

Of further interest are behaviors emerging from actuating within, and outside of, the frequency bounds set by Eq. \eqref{eqn:frequencies}. In particular, we wish to characterize the frequencies that result in stable dynamics and from that characterization deduce motion primitives for controlling multiple passive vehicles.
\begin{figure}[t]
    \centering
    \includegraphics[width=0.5\textwidth,trim={0 0cm 0 0cm},clip]{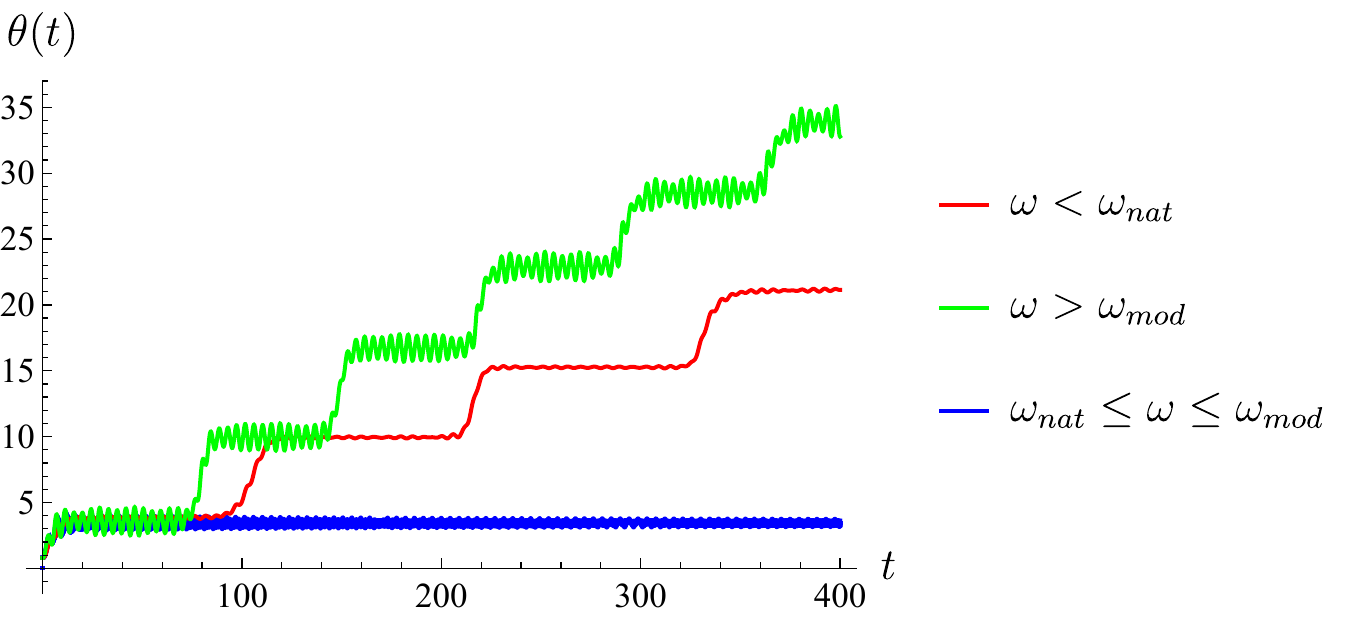}
    \caption{Analysis of the asymptotic heading of a Chaplygin beanie over the actuation bounds described in Fig. \ref{fig:frequencyresponse}}
    \label{fig:headinganalysis}
\end{figure}
An analysis of the time evolution of $\theta$ when actuating the platform at frequencies inside and oustide of the bands given above is shown in Fig. \ref{fig:headinganalysis}, experimentally clarifying the existence of distinct dynamics in the robot's heading. The blue curve is actually a family of curves all resulting from frequencies lying within the bounds of the natural and modal frequency of the robot, all resembling stable oscillatory behavior. The red and green curves each correspond to $\theta$ dynamics of a single frequency chosen that satisfies the inequalities shown in the legend.

\subsubsection{Manipulation}
\begin{figure}[t]
    \centering
    \includegraphics[width=0.45\textwidth,trim={0 0cm 0 0cm},clip]{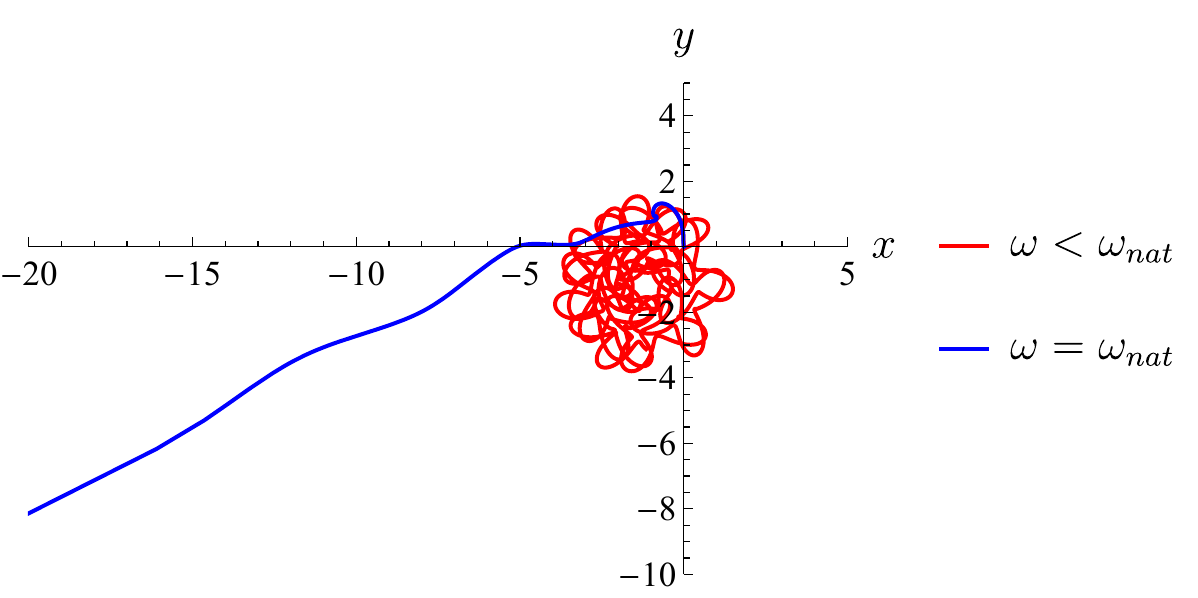}
    \caption{Trajectories of two individual simulations for actuation of the platform inside of the bounds (blue) and outside of bounds (red) defined by $\omega_{nat}$ and $\omega_{mod}$}
    \label{fig:trajectories}
\end{figure}

The frequency characterization carried out above provides clear rules by which we can exert control over the platform to manipulate Chaplygin beanies. Actuating the platform at frequencies within the bounds set by the natural frequency of the rotor and the modal frequency of the robot in free space allow for control primitives which induce robots to achieve stable undulatory locomotion along a particular heading. We term such behavior in the context of manipulating multiple robots as \textit{dispersion}. Actuation outside of these boundaries yield trajectories corresponding to much more complex dynamics, not as easily classified as those of stable undulatory behavior. We discuss some of these behaviors in Section \ref{sec:conc}. Two such trajectories are shown in Fig. \ref{fig:trajectories}.

The Chaplygin beanie under external actuation with $\omega = \omega_{nat}$ will disperse from its initial position and undulate stably at a particular heading for all time. The degree to which it stably oscillates in $\theta$ increases with increasing $\omega$, as long as actuation stays within the bounds of the natural frequency of the rotor and the modal frequency of the robot. Though no formal guarantee is given in this work, the authors assert that trajectories corresponding to those of platform actuation at frequencies of $\omega < \omega_{nat}$ or $\omega > \omega_{mod}$ will stay within some neighborhood of its initial position, much like that of the red trajectory in Fig. \ref{fig:trajectories}. Such trajectories are also prone to exhibit dynamics that reveal the presence of multi-scale time dynamics, discussed below.

\begin{figure}[t]
    \centering
    \includegraphics[width=0.45\textwidth,trim={0 0cm 0 0cm},clip]{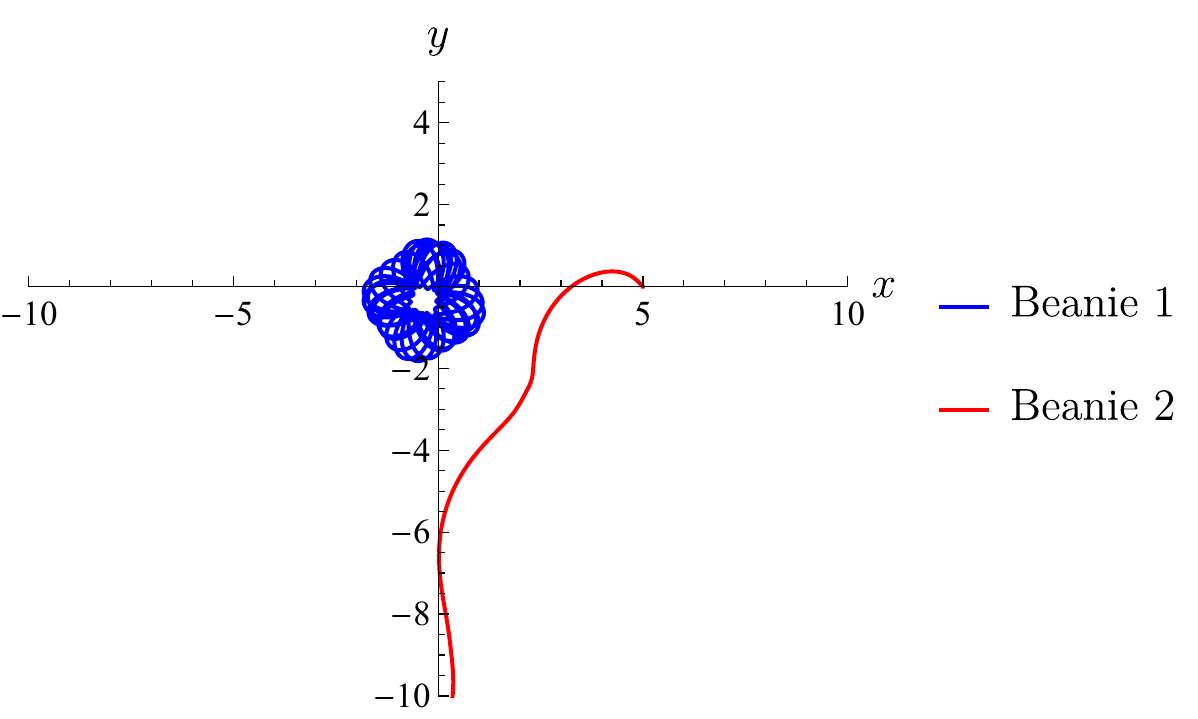}
    \caption{Resulting trajectories for two Chaplygin beanie agents atop an actuated platform. The blue trajectory corresponds to actuation of beanie 1 at a frequency lower than the natural frequency of its rotor. The red trajectory corresponds to the dynamics induced by actuating the platform so as to induce the behavior seen in beanie 1.}
    \label{fig:twoBeaniesOnAPlatform}
\end{figure}

This result clarifies the ability to control passive robots using the kind of actuation given by Eq. \eqref{eqn:controlrotated}. Consider a case with two identical passive Chaplygin beanies at rest atop an actuated platform with different initial headings. Naively assuming control over the platform to manipulate one robot in this sense does not guarantee a certain behavior for the other. In the presence of other passive robots, however, to achieve a desired behavior, the platform can be actuated corresponding to the desired control for that particular robot and its resulting behavior remains independent of the others. Fig. \ref{fig:twoBeaniesOnAPlatform} shows such a case if the desired behavior for Chaplygin beanie 1 is to stay within some neighborhood of its initial position.

Though the second Chaplygin beanie appears to stably locomote away in this example, there is no basis in assuming it does so. Similarly, actuation within the frequency bands discussed above would result in the vehicle approaching a stable oscillatory trajectory rather than oscillating about the initial position. In this result, we emphasize the importance of exerting control over a particular robot in a multi-robot setting given that we actuate the platform according to Eq. \eqref{eqn:controlrotated}. The ability to actuate the platform in this way leads to questions concerning the control of multiple robot. Naively, one may conclude this control methodology can be targeted to one robot and switched at any time to target another to disperse or station-keep robots as needed, but this is nullified by each robot having attained nonzero momentum.

%%%%%%%%%%%%%%%%%%%%%%%%%%%%%%%%%%%%%%%%%%%%%%%%%%%%%%%%%%%%%%%%%%%%%%
% \input{sections/swimmers}

%%%%%%%%%%%%%%%%%%%%%%%%%%%%%%%%%%%%%%%%%%%%%%%%%%%%%%%%%%%%%%%%%%%%%%
\section{Conclusions and Future Work}
\label{sec:conc}
We have presented an investigation into two distinct examples of externally actuated planar locomoting systems in ambient media, namely systems \textit{without} and \textit{with} drift, that rely on external interactions to achieve locomotion. In contrast to more traditional systems that locomote via commands to internal joints, the types of systems that we consider may require that we analyze and exploit the resulting external dynamics. These present difficulties in more complex mappings between different configuration components, as well as extra structure that cannot be easily reduced as before. Nevertheless, we have also presented a framework that incorporates these changes and is still amenable to established methods for analysis and motion planning.

The three-link wheeled robot on a platform is an example in which a \emph{stratified fiber bundle} structure, with the external platform's configuration forming a separate fiber space of its own. By manipulating both of the resultant connection mappings, we can find base trajectories that lead to desired fiber trajectories and vice versa, addressing the problem of actuating a passive robot using the platform only. We also discussed the problem of the broken symmetry in the relative orientation, and we proposed a method by which to eliminate this dependency for different ranges by using its periodicity.

\begin{figure}
    \centering
    \includegraphics[width=0.5\textwidth,trim={0 0cm 0 0cm},clip]{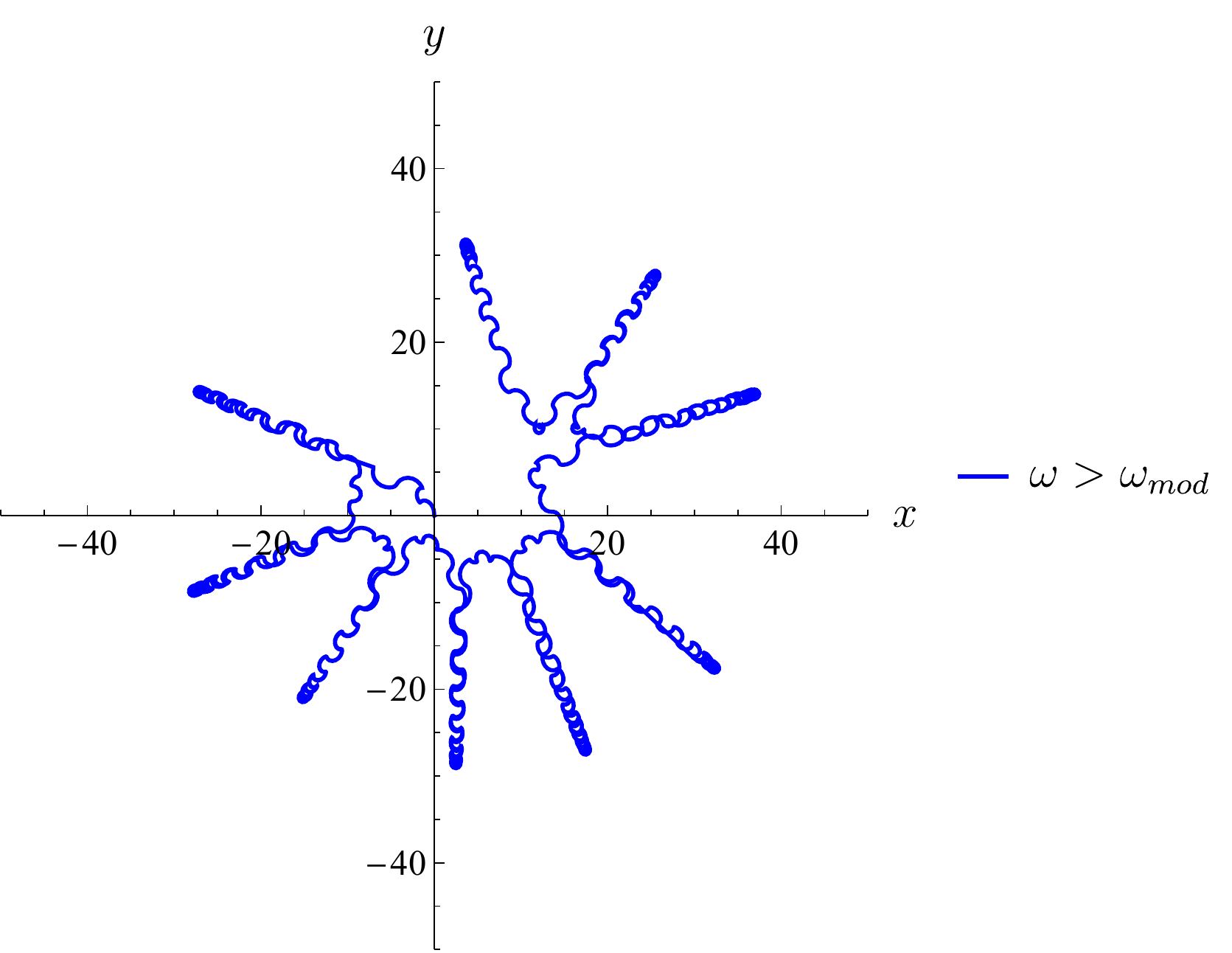}
    \caption{Trajectory resulting from actuating the platform at a frequency of $\omega < \omega_{mod}$ for a Chaplygin beanie with parameters $C = 0.5$, $m = B = k = 1$ for a duration of 500 simulation seconds.}
    \label{fig:trajectoryabovemodal}
\end{figure}

The Chaplygin beanie on a platform exemplifies the challenges introduced by drift in an externally actuated system's dynamics. Though it is clear based on the present work that there exist parameter-invariant bounds on platform actuation frequencies for achieving stable undulatory behavior of a passive Chaplygin beanie under external control, formal proof for this result is sought. Other interesting phenomena are exhibited by the nonlinear dynamics that warrant further exploration. Namely, certain platform actuation frequencies yield behaviors which indicate the presence of multi-scale time dynamics, demonstrated in Fig. \ref{fig:trajectoryabovemodal}. 

This behavior relates qualitatively to that exhibited by an externally actuated three-link snake-like robot in \cite{Kelly2016}.
The system was given an initial position at the origin and controlled corresponding to Eq. \eqref{eqn:controlrotated}. The robot locomotes away along some heading for some time, reverses direction, locomotes for some time, switches its heading, and repeats this behavior. The dynamics of moving along in some heading occur at a fast time scale, while the dynamics for switching direction occur at a much slower time scale. Changes in the direction taken by the robot likely correspond to bifurcations in the dynamics at one of these time scales, the analysis of which is a topic of future work. The actuated system has also been shown to display stable oscillatory behavior for frequencies within the frequency bands discussed. As such, proving stabilizability for the externally-actuated system will also be included in future publications.

% The problem of two interacting swimmers is different from the previous in that the medium itself does not have a configuration; instead of providing actuation to the fluid, we provide commands to physically separated bodies and allow their dynamic coupling to determine resultant motions. Such dynamics naturally depended on relative displacement, again yielding a symmetry-breaking configuration variable. However, by assuming small, oscillatory motions, we were able to consider the slow evolution of the bodies' trajectories as functions of the oscillation parameters only. These dynamics were further simplified by using certain motion primitives that either leave relative displacement unchanged or change it in a known way, which can then be combined to execute any desired locomotion in the plane. 

There are several avenues along which to move forward with this work. One is to produce a more general description of the ``inversion'' of the connection mapping when pushing trajectories from one fiber to another in a stratified structure. Here we were able to reuse some of the results of our companion work in analyzing the harmonics of the passive joint variables; we envision a more complete framework describing the usage of such methods in all these situations.

More generally, we showed that leveraging the tools of connection exterior derivative plots allowed us to achieve motion planning for a fully dynamic externally actuated three-link snake robot by varying the phasing between the two platform inputs. These tools gave us a structured way to accomplish motion planning, but do not consider the drift inherent in externally actuated systems. That is, any actuation of the platform beneath any robot will introduce second-order dynamics in the robot, like those seen in the Chaplygin beanie. Moving forward, considering, and even leveraging, these second-order dynamics will be a priority in controlling externally actuated systems. Of particular interest to the authors is the introduction of geometric optimal control in the sense of \cite{bloch2015geometric, colombo2017variationalgeometric} to externally actuated locomoting systems. Moreover, applications of this work in a multi-robot setting are also of interest to the authors, particularly in the context of \textit{mechanical communication} \footnote{Mechanical communication is a form of stigmergy \cite{holland1999stigmergy,tang2017stigmergy} and has been studied in the context of cell coordination in biology and engineering \cite{reinhartking2008,Nitsan2016,schwager2019cell}.}, where robots use a shared ambient medium to transmit \textit{information} as well as propulsive energy for locomotion.

% Secondly, we have dealt with the symmetry-breaking fiber variables in the two examples very differently in order to obtain reduced systems. In the robot-platform example, we exploited the periodicity of $\theta$ to obtain ``representative'' connections, on which the usual motion planning methods may be applied. In the interacting swimmers example, we placed restrictions on the types of commanded inputs in order to decouple kinematic and dynamic components of motion, and we also identified motion primitives that simplify the role of $x_{2,d}$. While both strategies were well suited to their respective examples, it would be desirable to again find a common solution framework. In this paper we systematically described the problem and presented solutions specific to certain problems; in future work we would expect to find a systematic way of solving different problems.

\bibliographystyle{ieeetr}
\bibliography{tro}

% \begin{IEEEbiographynophoto}{Blake Buchanan}
% Blake Buchanan is currently part of the research staff in the Biorobotics lab at Carnegie Mellon University. 


% % insert where needed to balance the two columns on the last page with
% %\newpage



%\vfill


\end{document}